**Title Page:**

Fine-tuning Large Language Model (LLM) Artificial Intelligence Chatbots in Ophthalmology and LLM-based evaluation using GPT-4

**Authors:**

Ting Fang Tan[1*], Kabilan Elangovan[1,2*], Liyuan Jin[2,3], Yao Jie[1,3], Li Yong[1,3], Joshua Lim[1], Stanley Poh[1], Wei Yan Ng[1], Daniel Lim[5,6], Yuhe Ke[4,6], Nan Liu[2,3], Daniel Shu Wei Ting[1,2,3] +

1. Singapore National Eye Centre, Singapore Eye Research Institute, Singapore, Singapore

2. Singapore Health Services, Artificial Intelligence Office, Singapore

3. Duke-NUS Medical School, Singapore, Singapore

4. Department of Anesthesiology, Singapore General Hospital, Singapore, Singapore

5. Department of Gastroenterology, Singapore General Hospital, Singapore

6. Data Science and Artificial Intelligence Lab, Singapore General Hospital, Singapore

*Contributed Equally

+ Corresponding Author

**Corresponding Author:**

Name: Daniel Shu Wei Ting

Email: daniel.ting45@gmail.com

Address: 31 Third Hospital Ave. Singapore 168753

Institution: Singapore Health Services, Artificial Intelligence Office, Singapore

**Abstract**

**Purpose:** To assess the alignment of GPT-4-based evaluation in relation to human clinician experts, for the evaluation of responses to ophthalmology-related patient queries generated by fine-tuned large language model (LLM) chatbots.

**Methods:** A dataset of 400 general ophthalmology questions and 400 paired answers were created by our ophthalmologists to represent commonly asked questions in real-world, across the spectrum of cataracts, myopia, and retinal diseases. This dataset was divided into fine-tuning (368 QnA pairs; 92%), and testing (40 QnA pairs; 8%). We find-tuned 5 different LLMs, including LLAMA2-7b, LLAMA2-7b-Chat, LLAMA2-13b, and LLAMA2-13b-Chat, based on the fine-tuning dataset as domain-specific knowledge. For the independent testing dataset, an additional 8 glaucoma QnA pairs were included. 200 responses to the testing dataset were generated by 5 fine-tuned LLMs for evaluation. A customized clinical evaluation rubric was used to guide GPT-4 evaluation of these LLM-generated responses, grounded on clinical accuracy, relevance, patient safety, and ease of understanding. GPT-4 evaluation was then compared against human ranking by 5 clinicians for clinical alignment.

**Main outcome measures:** To evaluate the clinical congruence of GPT-4 versus human clinician rankings of LLM-generated responses, quantitative—statistical metrics such as Pearson and Kendall Tau correlation coefficients, complemented by Cohen's Kappa— and qualitative assessment of GPT-4 evaluation responses were evaluated for further insights.

**Results:** Among all fine-tuned LLMs, GPT-3.5 scored the highest (87.1%), followed by LLAMA2-13b (80.9%) and LLAMA2-13b-chat (75.5%), with LLAMA2-7b-Chat and LLAMA2-7b scoring lower at 70.0% and 68.8%, respectively, based on the GPT-4 evaluation. GPT-4 evaluation demonstrated significant agreement with human clinician rankings, with Spearman's and Kendall Tau correlation coefficients of 0.90 and 0.80 respectively; while correlation based on Cohen's Kappa was more modest at 0.50. Notably, qualitative analysis and the glaucoma sub-analysis revealed clinical inaccuracies in the LLM-generated responses, which were appropriately identified by the GPT-4 evaluation.

**Conclusions:** The notable clinical alignment of GPT-4 evaluation highlighted its potential to streamline the clinical evaluation of LLM chatbot responses to healthcare-related queries. By complementing the existing clinician-dependent manual grading, this efficient and automated evaluation could value add to the validation of future developments in LLM applications for healthcare.

# 1. Introduction

The unprecedented language comprehension and generative capabilities of large language models (LLMs) have gained attention for their improved natural language processing capabilities and in particular their ability to generate relevant responses to unstructured free-text inputs in a conversational manner[1-3]. This is driven by transformer-based neural networks[4] that consist of an encoder-decoder architecture with self-attention capabilities to extract the contextual semantics of a sequence of words to process entire sentences in parallel, enabling pretraining on a massive corpus of text data on the internet.

Its early excellence in medical licensing exams[5-7] demonstrated its alignment and capability in medical knowledge retrieval and reasoning, catalyzing the exploration of LLM-driven applications in healthcare to potentially revolutionize existing clinical and administrative workflows[8]. To further align with healthcare-domain use cases, strategies developed included the use of large-scale medical corpus to pretrain or fine-tune general-purpose LLMs[9, 10], prompt engineering, and retrieval augmented generation. One of the key examples is MedPaLM2 (Google, 2023)[11], which surpassed its predecessor with over 85% accuracy on the MultiMedQA and several other benchmarking medical question-answering datasets. This was achieved by an improved backend PaLM2 LLM[12], medical domain-specific fine-tuning, and prompting strategies with a proposed ensemble refinement approach. Specific to ophthalmology, Ophtha-LLaMA2, fine-tuned on a limited dataset of reports of three ophthalmic investigation modalities (Color fundus photographs, optical coherence tomography, ocular surface analyzer), achieved comparable performance to other LLMs in diagnosing eye conditions [13]. While there is rapid development and their techniques, there are few study investigating at LLM medical evaluation phase and safety other than simple MCQ evaluations, readiness.

Beyond medical examination question-answering, the generative capability of LLMs has shown potential to transform clinical practice in delivering efficient and personalized patient care. The objective of this study is to evaluate 5 fine-tuned LLMs: GPT-3.5, LLAMA2-7b, LLAMA2-7b-Chat, LLAMA2-13b, and LLAMA2-13b-Chat in responding to patients' general questions related to ophthalmology conditions [8]. As a secondary aim, we also evaluated the use of customized Generative Pretrained Transformer 4 (GPT-4) for the automatic evaluation as a "clinical judge" for the responses generated by the fine-tuned LLMs (**Figure 1**).

# 2. Methodology

## 2.1. Clinical Tasks and Datasets: Fine-tuning and Testing

A total of 400 question and answer (QnA) pairs were manually crafted by a team of ophthalmologists to ensure clinical accuracy and relevance, consisting of a broad spectrum of common ophthalmic subspecialty conditions, namely cataract, cornea, retina, neuro-ophthalmology, oculoplastics, refractive error and pediatrics; as well as common eye complaints. The clinical categories of QnA pairs included risk factors, signs and symptoms, investigations, and management for each condition. Subsequently, a curated selection of 368 QnA pairs was extracted as the domain-specific knowledge used for fine-tuning both GPT-3.5 and LLAMA2 series (as detailed in Section 2.3). The distribution of QnA pairs for the fine-tuning dataset across the eight ophthalmic domains is detailed in **Table 1.**

The remaining 32 questions were randomly selected from the initial pool to be kept as the independent testing dataset, with 4 QnA pairs from each ophthalmic domain (**Table 1**). The testing dataset served as a representative sample for the objective of evaluation of model proficiency and applicability, providing a glimpse into the nature and scope of queries addressed in the study, offering insights into the practical utility and relevance of the LLMs in interpreting and responding to domain-specific inquiries. Additionally, 8 glaucoma-related QnA pairs were specifically crafted and included in the test dataset for further subanalysis (**Table 1**). This choice was deliberate: glaucoma-related QnA pairs were intentionally omitted from the fine-tuning datasets. The inclusion of these pairs in the testing phase allows for a focused assessment of the LLMs' performance on non-fine-tuned data, providing valuable insights into their ability to generalize knowledge in areas not explicitly covered during training. This sub-analysis serves as a crucial component of our study, highlighting the adaptive capabilities of the LLMs in handling diverse and unforeseen medical scenarios within the field of ophthalmology.

## 2.2 Selection of Large Language Models (LLMs)

We integrate two prominent language models, namely GPT-3.5 and the 4 LLAMA-2 series, into our study framework. The predilection for GPT-3.5 and LLAMA2 emanates from their demonstrated superior performance, adaptability, and distinct operational efficiencies. These attributes collectively furnish a solid and versatile basis for comprehensive fine-tuning and empirical evaluation. Such a strategic selection aims to augment the scope of machine learning proficiency in the domain of linguistics, facilitating an in-depth exploration of the capabilities inherent in these sophisticated computational architectures.

Specifically, GPT-3.5, an iteration in the GPT lineage, provides an unrivaled depth in language comprehension and generation capabilities, rendering it an exemplary candidate for intricate and nuanced linguistic tasks. The model's continuous enhancements are directed towards not only improving performance metrics but also ensuring ethical applications, positioning it as a vanguard in the arena of generative language technologies. Conversely, the LLAMA2 series is recognized for its efficiency and scalability, serving an extensive spectrum of linguistic applications while accommodating diverse computational demands. The adoption of these models permits an expansive examination of the capabilities and constraints of contemporary LLMs, thereby laying a foundational framework for investigating the effects of fine-tuning on various aspects and practical applications.

## 2.3 Fine-Tuning Experiments

In our research, we fine-tuned 5 different LLMs, including GPT-3.5 and 4 variations within the LLAMA2 series: LLAMA2-7b, LLAMA2-7b-Chat, LLAMA2-13b, and LLAMA2-13b-Chat. The primary objective was to augment the capability of these models to yield pertinent and accurate responses to ophthalmology patient queries. While existing research demonstrates the proficiency of GPT and LLAMA2 models in addressing clinical inquiries, our aim was to objectively tune the LLM models to reduce possible chances of hallucinations, thereby enhancing their reliability and accuracy in medical contexts. We sought to anchor the models' knowledge base in curated data, thereby refining their responses with a heightened degree of reliability and specificity tailored to the domain of ophthalmology. For this purpose, we utilized GPT-3.5-turbo-0613 via the OpenAI fine-tuning API, and the LLAMA2 series models[14] were fine-tuned using the H2O.ai library[15].

The LLAMA2 series offers two core models: LLAMA2-7b and LLAMA2-13b, which are general-purpose language models with 7 billion and 13 billion parameters respectively. These foundational models are designed for a wide range of machine learning tasks. However, to better suit conversational applications, we also included their specialized counterparts: LLAMA2-7b-Chat and LLAMA2-13b-Chat. These "Chat" variants are fine-tuned versions of their respective base models, optimized specifically for generating and understanding the nuances of conversational dialogue. They are particularly adept at maintaining context and producing more coherent and contextually appropriate responses in conversational settings[14].

To ensure experimental fidelity, a consistent system prompt was adopted across all models, and the number of training epochs was standardized at 3 for each fine-tuning experiment. Specifically, the system prompt was formulated as follows: "*You are a friendly and brilliant medical chatbot, designed to provide concise and accurate answers with regards to all ophthalmology-related queries. Ensure your responses are clear, to the point, and avoid repetition.*" This prompt was strategically crafted to instill a knowledgeable yet approachable persona in the chatbot, focusing on delivering precise and succinct medical advice.

These models were rigorously configured with fine-tuning parameters tailored to enhance their ability to generate precise and dependable responses. The configuration included:

- Maximum Prompt Length: Set to 256 tokens, this parameter defines the upper limit for the length of input prompts.

- Maximum Answer Length: Also set to 256 tokens, dictating the maximum length for generated responses.

- Overall Maximum Length: Capped at 512 tokens to encompass both prompt and response lengths.

- Loss Function: Employed TokenAveragedCrossEntropy, a variant focusing on the average token's contribution to the loss, working in tandem with the AdamW optimizer.

- Learning Rate: Maintained at 0.0001, ensuring gradual and stable model adjustments.

- Batch Size: Fixed at 2 to balance between computational demand and learning effectiveness.

- Epochs: Consistently set to 3, providing the models with a uniform duration for adaptation.

- Low-Rank Adaptation (Lora): Activated with settings of R=4, Alpha=16, and Dropout=0.05 to facilitate nuanced modifications to the model's internal representations[16].

- Gradient Checkpointing and Mixed Precision: Incorporated to elevate training efficiency without compromising output quality.

- Temperature and Repetition Penalty: Adjusted to 0.3 and 1.2, respectively, to foster response variety and precision.

The fine-tuning of LLAMA2 models predominantly utilized the default parameters provided by H2O's LM Studio, representing a streamlined and effective strategy. This approach secured a dependable performance foundation and eased the fine-tuning procedure. Utilization of these default settings enabled a concentrated investigation into the models' versatility and performance in the specialized context of ophthalmology, with the goal of achieving accurate and reliable results. The domain-specific knowledge used for fine-tuning was elaborated earlier in Section 2.1.

Fine-tuning time taken for each model was 6:25 for LLAMA2-7b, 6:30 for LLAMA2-7b-Chat, 11:58 for LLAMA2-13b, 12:12 for LLAMA2-13b-Chat, and notably longer at 38:32 for GPT-3.5 (in minutes:seconds; in increasing order; **Table 3(A)**).

## 2.4 Experimental Platform

The experimental platform for this study was a Windows 11 operating system utilizing Windows Subsystem for Linux 2 (WSL2). The hardware comprised a single GPU system with an NVIDIA RTX 4090, with 24 GB VRAM, and powered by a 12th Gen Intel(R) Core(TM) i9-12900K CPU. Python programming language Version 3.10 was used within the Linux subsystem for executing the LLAMA-2 model fine-tuning jobs.

## 2.6 Evaluation Metrics

We employed GPT-4[17] as an automated evaluator for the responses generated by the 5 fine-tuned LLMs (GPT-3.5, LLAMA-2-7b, LLAMA-2-13b, LLAMA-2-7b-chat, LLAMA-2-13b-chat). First, we generated responses to the test questions using GPT-4 (**Supplementary Material Table 1**), which were qualitatively verified as clinically accurate by a Senior Consultant Ophthalmologist. Next, a customized evaluation rubric was developed in collaboration with a team of ophthalmologists, detailed in **Table 2**, as the system prompt to guide the evaluation process using GPT-4. To minimize deviation and ensure focused assessments, the temperature parameter was set to 0.1. The evaluation output consisted of an overall score out of 100, and further elaboration based on factors including clinical accuracy, relevance to question intent, patient safety, and ease of understanding. This method provided a streamlined and automated alternative to traditional evaluation techniques, significantly reducing the dependency on extensive manual human grading.

To ensure high-quality evaluation standards, we subsequently sought to validate the alignment of GPT-4's automated grading with clinical judgment. To this end, 5 clinicians were tasked with ranking the responses from the 5 fine-tuned LLMs from 1st to 5th, with the 1st indicating the highest quality and 5th the lowest. The clinician panel consisted of 3 ophthalmology residents (1 junior resident, 2 senior residents who were referred to as board-certified ophthalmologists (i.e. certified fellows of The Royal College of Ophthalmologists), and 2 ophthalmology PhD candidates. The concordance between the clinicians' rankings and GPT-4's evaluations was statistically examined using Cohen's Kappa[17], Spearman and Kendall Tau correlation coefficients[18, 19]. For Spearman's correlation, the p-value was determined by converting the correlation coefficient (ρ) into a t-statistic using the formula: $t = \frac{p\sqrt{(n-2)}}{\sqrt{(1-p^2)}}$, where n is the sample size, and then comparing this t-statistic to a t-distribution with n-2 degrees of freedom to find the p-value. For Kendall Tau, the p-value was calculated by converting the tau coefficient (τ) into a z-score using the formula: $z = \frac{3\tau\sqrt{(n(n-1))/2}}{\sqrt{(2(2n+5))}}$, where n is the number of pairs, and then comparing this z-score to the standard normal distribution to obtain the p-value.

These calculations were critical for assessing the statistical significance of the observed correlations, providing a quantitative measure to support or refute the hypotheses under consideration. This dual-layered approach, combining automated assessment with expert validation, aimed to provide a comprehensive and reliable measure of the LLMs' performance in generating clinically relevant and accurate responses within the domain of ophthalmology.

## 3. Results

### 3.1 LLM Inference Calibration

In the model fine-tuning inference process for both the GPT-3.5 and LLAMA models, specific parameters were set to standardize the responses. The temperature was fixed at 0.2 to minimize response variability and ensure predictability. The context length was established at 256 tokens, allowing adequate contextual information for accurate responses. Additionally, a repetition penalty of 1.0 was implemented to reduce redundant content in the outputs. These settings were chosen to optimize the consistency and relevance of the LLMs' responses in a specialized medical domain.

### 3.2 Fine-Tuning Results

There were 200 GPT-4 evaluations and 200 human clinician rankings pairs for the 200 LLM-generated responses to the test set of 40 ophthalmology-related patient queries. GPT-3.5 scored the highest in accuracy (87.1), followed by LLAMA2-13b (80.9) and LLAMA2-13b-chat (75.5), with LLAMA2-7b-Chat and LLAMA2-7b scoring lower at 70.0 and 68.8, respectively (**Table 3(A)**).

In the evaluation comparing GPT-4's rankings of LLM responses with those of human clinicians, notable agreement was demonstrated based on Spearman's correlation coefficients and Kendall's Tau values. For 4 out of 6 clinician comparisons, Spearman's Correlation exceeded 0.70, with values around 0.90 for the Junior Resident, Ophthalmology PhD candidate-1, Board-certified Ophthalmologist-1, and the combined clinician group. These results suggest a strong alignment in ranking order. The Kendall's Tau values, aligned with Spearman's coefficients, further confirmed this strong correlation, particularly with Tau values near 0.80 for the same clinicians and the combined group. However, the moderate Cohen's Kappa values, up to 0.50 in 3 comparisons, may be less reflective of agreement in this context of ranking evaluations. Despite this, Kappa was included to provide a comprehensive view of the agreement, acknowledging the complexity inherent in comparing algorithmic and human assessments in a ranking format (**Table 3(B)**).

Additionally, we aimed to investigate the agreement between human clinicans and GPT-4's rankings of LLMs, framed around two hypotheses: H0, suggesting no significant correlation (implying agreement by chance), and H1, proposing a significant correlation (indicating meaningful agreement). The findings, critically supported by p-values significantly below the 0.05 threshold for both Spearman's correlation and Kendall's tau, robustly refuted H0 in favor of H1, indicating a strong statistical agreement between clinical and GPT-4 assessments. This significant correlation, as detailed in **Table 3(B)**, highlights the substantial alignment in evaluations, primarily based on the compelling evidence provided by the low p-values across various comparisons with clinicians.

Overall, GPT-3.5 responses were ranked first by both GPT-4-based evaluation as well as human clinician ranking in most cases. On the other hand, both GPT-4 and human clinicians had the least preference for LLAMA2-7b-generated responses. The responses by GPT-3.5 to most test questions were noted to be clinically accurate and succinct. On the other hand, responses from the LLAMA2 series were noted on several occasions to contain unnecessary repetition of phrases with redundant or irrelevant content. Particularly, one of the responses was flagged up to contain dangerously erroneous information. For the test question *"What is the management of cataracts in children?"*, the LLAMA2-13b response alarmingly included that cataract surgery *"is performed when the child is still in the womb…called foetal cataract surgery and is usually performed when the fetus is at least 24 weeks old"* (**Supplementary Material Table 2**). The confabulation of the non-existent *"foetal cataract surgery"* presented in a seemingly factual manner may be misleading for patient and caregivers. This highlighted that while LLAMA2-13b responses were ranked relatively higher overall, closer scrutiny of each response still uncovered clinical inaccuracies. Reassuringly, the GPT-4 evaluation specific to this response correctly identified the error and accurately outlined the management principle of pediatric cataracts in preventing amblyopia (**Supplementary Material Table 2**).

We observed that the GPT-4 to human correlation was particularly poor for only 1 (Board-certified Ophthalmologist-2) of the 5 clinicians. The gradings were vastly different from the other clinicians and also inconsistent when compared with the fellow board-certified Ophthalmologist with the same years of clinical experience. This highlights the inter-grader variability that is inevitable in human evaluation. Our study included human clinician graders representative of different levels of experience, and attempted to minimize the variability in

GPT-4's evaluation by setting the temperature to 0.1, to limit the randomness of evaluation by GPT-4. Further work would be needed to explore the variability of GPT-4 responses.

**3.3 Sub-Analysis on External Glaucoma Test Questions**

For the sub-analysis focusing on glaucoma-related test questions, we assessed the efficacy of LLMs in handling these subspecialty queries despite not being explicitly trained for it. Based on GPT-4 evaluation, a significant shift was noted with LLAMA2-13b and LLAMA2-13b-Chat exceeding the performance of GPT-3.5, scoring 90 and 76.2 respectively, as opposed to a score of 73.1 for GPT-3.5. These findings, detailed in **Table 4(A),** illustrated a notable variation in the models' proficiency when dealing with specialized medical content. LLAMA2-7b-Chat and LLAMA2-7b similarly performed the poorest, scoring 69.4 and 53.1 respectively.

For the test question *"What is microinvasive glaucoma surgery (MIGS)?"*, the LLAMA-13b and GPT-3.5 responses accurately described the procedure, and further highlighted its surgical indication for patients with mild to moderate glaucoma (and not for advanced glaucoma) which is aligned with clinical practice **(Table 5)**. This was accurately identified and appropriately scored higher by the GPT-4 evaluation (**Supplementary Material table 2**). On the other hand, medically inaccurate information was again noted in the LLAMA2-7b responses to glaucoma test questions. For the same test question on MIGS, the LLAMA2-7b response incorrectly identified that MIGS is indicated for patients who are not suitable for conventional glaucoma surgery, and contained unwarranted repetition of sentences with redundant information. In another example, the objective of the test question *"Why is it that my eye pressure is not high but the doctor says that I have glaucoma?"* was on normal tension glaucoma where patients have glaucomatous damage despite having normal eye pressure. The LLAMA2-7b response incorrectly diagnosed this clinical scenario as ocular hypertension, which instead refers to another condition with high eye pressure and no glaucomatous damage. GPT-4 evaluation accurately identified the incorrect diagnosis made in the LLAMA2-7b response, and highlighted the potential confusion to the patient.

In addition to these observations, the 'unseen' glaucoma test questions elucidated factual errors even in GPT-3.5 and LLAMA2-13b responses, despite their high overall rankings. For example, for the glaucoma test question *"What is the procedure of laser peripheral iridotomy like?"*, the GPT-3.5 response accurately identified its indication in the management of an angle closure glaucoma attack, however erroneously included that acute angle closure *"is a common condition that is usually not serious"*. Acute angle closure, if not promptly treated, may lead to irreversible glaucomatous optic nerve damage, corneal decompensation, and blindness[20]. Therefore, the medically inaccurate GPT-3.5 response could have spread misinformation and resulted in harm to patient care. This error was astutely identified by the GPT-4 evaluation and appropriately penalized. For the same test question on laser peripheral iridotomy, the LLAMA2-13b and LLAMA-13b responses described the procedure requiring the patient to *"lie on your side"*, which was incorrect as the procedure should be performed in upright sitting position at the laser machine. While the overall response was relevant and this error was arguably 'minor' without threatening patient safety, this demonstrated the lack of accuracy in the finer details which was not in keeping with clinical practice. Notably, this error was not identified by the GPT-4 evaluation.

In the further evaluation for alignment by comparing GPT-4's rankings of LLM responses with those of clinicians', disparities in statistical correlations were again observed in the glaucoma sub-analysis (**Table 4(B)**). These metrics suggest that while there was an overarching correlation, the exact alignment of specific glaucoma-related responses varied significantly among the clinical evaluators. This disparity underscored the complexities inherent in aligning algorithmic evaluations with human expertise, particularly in a specialized field such as ophthalmology.

Additionally, GPT-4 assigned a lower ranking to the fine-tuned GPT-3.5 model, prompting an investigation into the impact of fine-tuning on GPT-3.5. This led to a comparative analysis between the native GPT-3.5 responses and those of the fine-tuned GPT-3.5, as evaluated by GPT-4. The analysis revealed a decline in performance for the native GPT-3.5 model in the context of glaucoma test questions, where the average score attributed by GPT-4 to the native GPT-3.5 model was significantly higher at 94.5, compared to 73.1 for the fine-tuned GPT-3.5 model. This outcome interestingly suggests that fine-tuning may contribute to a degradation of the original knowledge base, potential compromising the model's ability to generalize. A summary of all the generated responses by the native and fine-tuned GPT-3.5 models and their respective GPT-4 evaluation is detailed in **Supplementary Material Table 3**.

## 4. Discussion

Based on our initial findings, we found that responses generated by fine-tuned LLMs for ophthalmology-related patient queries were largely clinically relevant and easy to understand. However, on closer scrutiny of these LLM responses, factual errors of varying clinical significance to patient safety were noted on several occasions. This limits the feasibility of deploying LLMs, even after fine-tuning with domain-specific knowledge, as patient-fronting medical chatbots in the healthcare setting till more work is done to investigate and improve their performance. Furthermore, we demonstrated that GPT-4 has potential to automatically and effectively evaluate these LLM-generated responses. By targeting its evaluation using our proposed rubric prompt, GPT-4 demonstrated alignment with human clinician evaluation, and appropriately identified inaccuracies within LLM responses in keeping with clinical practice.

While high congruence between GPT-4 evaluations and human clinician ranking was noted based on Spearman correlation and Kendall Tau, we observed that correlation on Cohen's kappa was less aligned. One possible reason could be that Spearman correlation and Kendall Tau reflect the correlation of the general order and relative differences between the rankings between GPT-4 and human clinicians. While uncovering insights on exact matches in ranking, Cohen's kappa may not be ideal for ranking correlation as it regards each rank as a separate category and ignores the ordinal nature of the data.

Our fine-tuning efforts were predominantly directed toward the LLAMA2 series models due to the limited customization capabilities offered by GPT-3.5's fine-tuning parameters. Additionally, we did not explore the LLAMA2-7b models due to GPU memory constraints, which limited our capacity to conduct extensive fine-tuning experiments. This consideration was crucial in determining the scope of our research and the selection of models suitable for our computational resources and objectives. Despite these constraints, our study revealed interesting trends. Out of all the fine-tuned LLMs selected for this study, we observed that GPT-3.5 was overall ranked first by both GPT-4 and human clinicians. This is consistent with the strong performance of GPT-3.5 in adapting to downstream tasks demonstrated in prior work. However, there have also been observations suggesting that GPT-4 tended to prefer outputs from its own model, potentially favoring the GPT-3.5 responses. Further work can be explored to assess GPT-4 evaluation in other models with the exclusion of GPT-based models. Within the LLaMA2 models, responses from LLAMA2-7b-chat and LLAMA2-13b-chat which comprised chat base models that were fine-tuned with conversational data by MetaAI, were preferred over base LLAMA2 models. Responses from both LLAMA2-13b and LLAMA-2-13b-chat were preferred over the 7B models with fewer parameters.

In striving for automated and reference-free evaluation, similar efforts have been explored to leverage LLM as an evaluator, two notable proposals were G-EVAL[21] and LLM-EVAL[22]. G-EVAL was proposed as a framework using GPT-4 and chain-of-thought prompting to evaluate LLM-driven text summarization tasks. LLM-EVAL comprised a single-prompt unified multi-dimensional automatic evaluation of open-domain LLM conversations. It encompassed conversation quality, such as content, grammar, relevance, and appropriateness, which demonstrated a high correlation with human evaluation. There were key differences between these examples and our proposed GPT-4 evaluation framework. First, our key evaluating metrics is tailored specifically to the healthcare domain, taking into account factors including clinical accuracy, patient safety, and ease of understanding of medical jargon from a layman's perspective. These are specific dimensions crucial to consider in the clinical setting, in addition to prior work that focused more on fluency and grammaticality. Second, in the experimental setup for the GPT-4 evaluation, we optimized the performance of LLMs tested via fine-tuning with domain-specific knowledge. Third, our approach of numerical ranking comparisons between LLM and human evaluation facilitated stratification of results to uncover subtle differences.

As more LLM-driven healthcare use cases are developed, there is a push for more intricate evaluation beyond the assessment of accuracy such as with benchmarking datasets. While evaluation based on language features such as readability, fluency, and grammar is intrinsic to an LLM, qualitative metrics looking specifically at clinical application need to be established. In addition to medical reasoning and contextual understanding, emphasis has been placed on the evaluation of potential harm and perpetuating biases. Proposed evaluation frameworks outline multi-dimensional aspects for consideration. This is especially crucial in the healthcare domain where it is fundamental to safeguard patient safety and mitigating biases are crucial. To our knowledge, the current framework is the first design of LLM as an evaluator application in clinical scenarios. This concept could be easily adopted and translated to other specialties in medicine.

## 5. Conclusion

Therefore, our results suggest the potential of GPT-4 for the automatic evaluation of responses from LLM-powered medical chatbots, indicating a promising avenue for future LLMs research. This initial investigation lays the groundwork for extensive validation of GPT-4's evaluation capabilities across larger datasets, a broader range of LLMs, and varied healthcare applications. Additionally, the integration of diverse multidimensional evaluation frameworks into our evaluation rubric could further refine the depth and precision of GPT-4's outputs. The effectiveness of LLMs, particularly those enhanced through fine-tuning experiments, in creating robust chatbots has been demonstrated. Moreover, the ability to develop these advanced chatbots using consumer-level GPUs is a significant stride, making this cutting-edge technology increasingly accessible for broad-scale deployment.

# Figures

# Figure 1: Overview of the developmental pipeline of this study

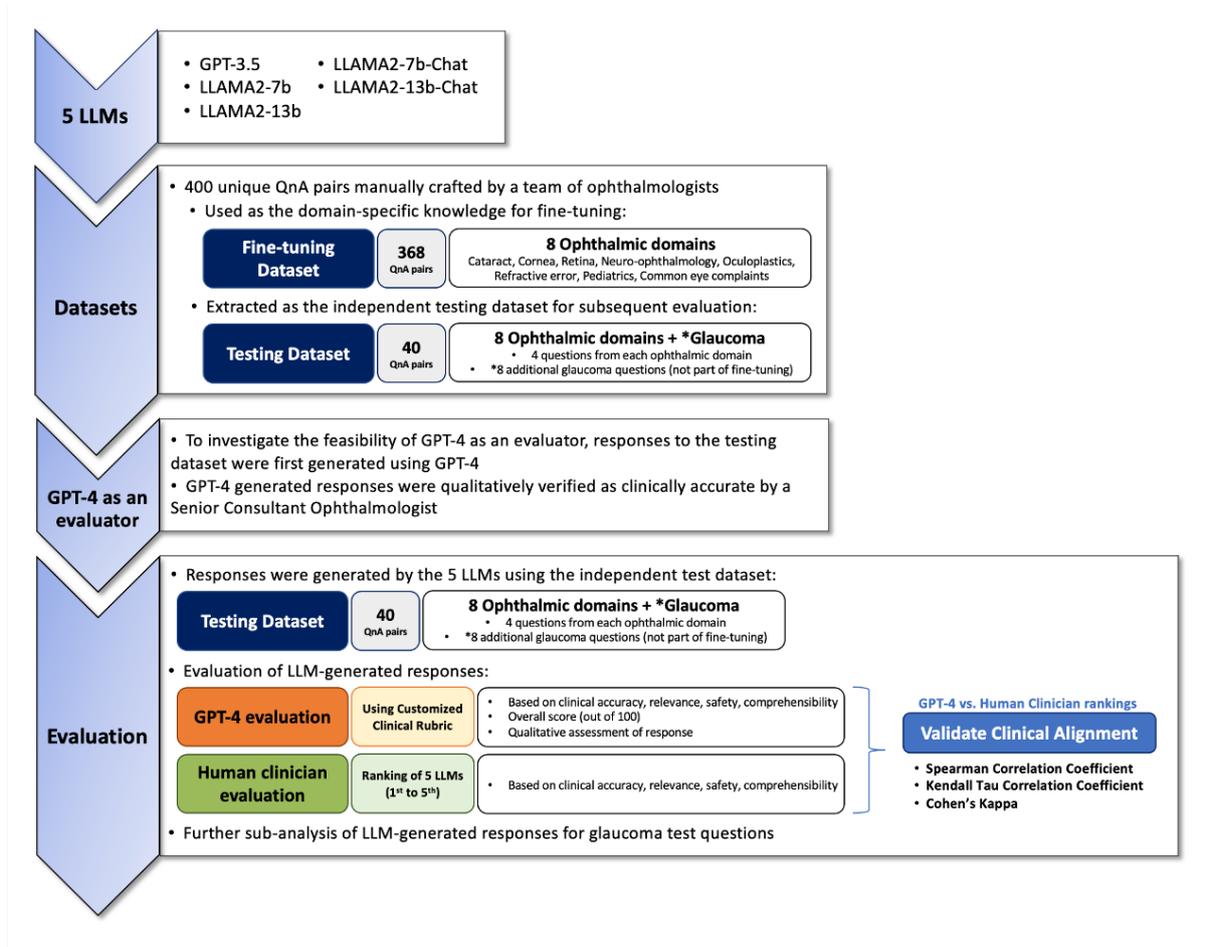

**Tables**

**Table 1. Distribution of the 408 question and answer (QnA) pairs in the fine-tuning and testing dataset, across the 8 ophthalmic domains**

| Ophthalmic Domains | Examples | Number of QnA Pairs (Fine-tuning) | Number of QnA Pairs (Testing) |
|---|---|---|---|
| Cataract | Cataract, cataract surgery | 22 | 4 |
| Cornea | Keratoconus, corneal infection, corneal transplant, burns or chemical injury | 60 | 4 |
| Retina | Diabetic retinopathy, age-related macular degeneration, epiretinal membrane, retinal tear or detachment | 51 | 4 |
| Neuro-ophthalmology | Cranial nerve palsies, myasthenia gravis, migraine | 32 | 4 |
| Oculoplastics | Chalazion, entropion, ptosis, nasolacrimal duct obstruction | 46 | 4 |
| Refractive error | Myopia, astigmatism | 60 | 4 |
| Paediatrics | Retinopathy of prematurity, amblyopia | 39 | 4 |
| Common eye complaints | Blur vision, eye redness, pain | 58 | 4 |
| Glaucoma | Primary open angle glaucoma, acute angle closure glaucoma, normal tension glaucoma | - | 8 |

**Table 2. GPT-4 evaluation rubric system prompt used.**

| Criteria | GPT-4 Evaluation Rubric |
|---|---|
| **Evaluation Focus** | Please act as an impartial judge and evaluate the quality of the response provided by an AI chatbot to the user questions pertaining to ophthalmology and eye health as displayed below . Your evaluation should consider factors including clinical accuracy (i.e. response contains medically accurate information that is aligned with clinical practice), relevance to question intent (i.e. response addresses the specific question intent), patient safety (i.e. no dangerously erroneous information), and easy to understand from the layman patient's perspective. |
| **Evaluation Method** | Begin your evaluation by providing a short explanation. Be as objective and consistent as possible. After providing your explanation, you must rate the response on a scale of 1 to 100 by strictly following this format. Both the question and chatbot answer will be shared sequentially. |

**Table 3(A). Overall GPT-4 evaluation (Scores and Rankings) and Clinician Rankings**

| LLM | Fine-Tuning Time Taken (min:sec) | GPT4 Score | GPT4 Ranking | Junior Resident | Ophthalmology PhD candidate-1 | Ophthalmology PhD candidate-2 | Board-certified Ophthalmologist-1 | Board-certified Ophthalmologist-2 | Overall Clinical Ranking |
|---|---|---|---|---|---|---|---|---|---|
| GPT-3.5 | 38:32 | 87.1 | 1st | 1st | 1st | 1st | 1st | 2nd | 1st |
| LLAMA 2-13b | 11:58 | 80.9 | 2nd | 2nd | 3rd | 3rd | 3rd | 4th | 3rd |
| LLAMA 2-13b-CHAT | 12:12 | 75.5 | 3rd | 3rd | 2nd | 4th | 2nd | 1st | 2nd |
| LLAMA 2-7b-CHAT | 6:30 | 70.0 | 4th | 5th | 4th | 2nd | 4th | 3rd | 4th |
| LLAMA 2-7b | 6:25 | 68.8 | 5th | 4th | 5th | 5th | 5th | 5th | 5th |

**Table 3(B). Overall comparison of rankings between GPT-4 and Clinical Graders based on Spearman, Kendall Tau Correlation Coefficients and Cohen's kappa**

| Human Clinician | Spearman (p-value) | Kendall Tau (p-value) | Cohen Kappa |
|---|---|---|---|
| Junior Resident | 0.90 ($p < 0.01$) | 0.80 ($p < 0.01$) | 0.50 |
| Ophthalmology PhD candidate-1 | 0.90 ($p < 0.01$) | 0.80 ($p < 0.01$) | 0.50 |
| Ophthalmology PhD candidate-2 | 0.70 ($p < 0.01$) | 0.60 ($p < 0.01$) | 0.25 |
| Board-certified Ophthalmologist-1 | 0.90 ($p < 0.01$) | 0.80 ($p < 0.01$) | 0.50 |
| Board-certified Ophthalmologist-2 | 0.62 ($p < 0.01$) | 0.53 ($p < 0.01$) | 0.25 |
| **Clinical Combined Grading** | **0.90** ($p < 0.01$) | **0.80** ($p < 0.01$) | **0.50** |

**Table 4(A). Glaucoma-related GPT-4 evaluation (Scores and Rankings) and Clinician Rankings**

| LLM | GPT-4 Score | GPT4 Ranking | Junior Resident | Ophthalmology PhD candidate-1 | Ophthalmology PhD candidate-2 | Board-certified Ophthalmologist-1 | Board-certified Ophthalmologist-2 | Overall Clinical Ranking |
|---|---|---|---|---|---|---|---|---|
| GPT-3.5 | 73.1 | 3rd | 2nd | 1st | 1st (tied) | 1st | 1st | 1st |
| LLAMA2-13b-Chat | 76.2 | 2nd | 3rd | 3rd | 1st (tied) | 2nd | 2nd | 3rd |
| LLAMA2-13b | 90.0 | 1st | 1st | 2nd | 1st (tied) | 3rd | 4th | 2nd |
| LLAMA2-7b-CHAT | 69.4 | 4th | 5th | 4th | 4th | 4th | 3rd | 4th |
| LLAMA2-7b | 53.1 | 5th | 4th | 5th | 5th | 5th | 5th | 5th |

**Table 4(B). Comparison of rankings between GPT-4 and Clinical Graders on glaucoma-related test questions based on Spearman, Kendall Tau Correlation Coefficients and Cohen's kappa**

| Human Clinician | Spearman | Kendall Tau | Cohen Kappa |
|---|---|---|---|
| Junior Resident | 0.80 | 0.60 | 0.0 |
| Ophthalmology PhD Candidate-1 | 0.70 | 0.60 | 0.25 |
| Ophthalmology PhD Candidate-2 | 0.89 | 0.84 | 0.50 |
| Board-certified Ophthalmologist-1 | 0.60 | 0.40 | 0.50 |
| Board-certified Ophthalmologist-2 | 0.30 | 0.20 | 0.25 |
| **Overall Clinical Grading** | **0.70** | **0.60** | **0.25** |

**Table 5. LLM-generated responses and overall scores given by the GPT-4 evaluation for the test question on minimally invasive glaucoma surgery** *(Green highlights accurate information consistent with clinical practice; while Red highlights the inaccurate information or irrelevant repetitions)*

| Test Question | Response generated by each LLM | | | | |
|---|---|---|---|---|---|
| | GPT-3.5 | LLAMA2-13b-Chat | LLAMA2-7b-Chat | LLAMA2-13b | LLAMA2-7b |
| **What is minimally invasive glaucoma surgery (MIGS)?** | Minimally invasive glaucoma surgery MIGS is a new and rapidly evolving field in glaucoma treatment. MIGS procedures are performed using microscopic equipment and tiny incisions, and are designed to be safer and less invasive than traditional glaucoma surgery. They are usually performed as day surgery, and have a faster recovery time. MIGS procedures are usually performed in patients with mild to moderate glaucoma, and are often combined with cataract surgery. They are not suitable for patients with advanced glaucoma. | Minimally invasive glaucoma surgery MIGS is a new type of glaucoma surgery that is performed through a small opening in the eye. It is less invasive and has less tissue trauma compared to traditional glaucoma surgery. MIGS is performed as a standalone procedure or in combination with cataract surgery. MIGS is performed to lower the intraocular pressure to a level that is safe for the eye. It is usually performed in patients with mild to moderate glaucoma. MIGS is not suitable for patients with severe glaucoma. Your ophthalmologist will perform a complete assessment and recommend the appropriate treatment for you. | Minimally invasive glaucoma surgery MIGS is a new type of glaucoma surgery that is less invasive than traditional glaucoma surgery. MIGS is performed through small incisions and does not require the removal of any tissue from the eye. MIGS is usually performed as a combined procedure with cataract surgery. MIGS is designed to reduce the risk of complications and speed up the recovery time. MIGS is suitable for patients with mild to moderate glaucoma and is usually recommended in advanced stages of glaucoma. Your ophthalmologist will perform a detailed assessment and recommend the appropriate surgery for you. You can also visit your ophthalmologist for more information. | Minimally invasive glaucoma surgery MIGS is a new class of glaucoma surgery that is performed through the use of small devices that are inserted into the channels of the eye. These devices help to lower the intraocular pressure by either reducing the amount of fluid produced by the eye or by improving the outflow of fluid from the eye. MIGS procedures are performed in conjunction with or in place of the traditional glaucoma surgery. The advantage of MIGS is that it is less invasive and has a lower risk of complications. Examples of MIGS procedures include XEN, Trabectome, iStent, Kahook Dual Blade and Drainage Implant. | Minimally invasive glaucoma surgery MIGS is a new type of surgery that is performed through a small hole in the sclera the white of the eye. It is a minimally invasive procedure that is performed under local anaesthesia. It is a good option for patients who are not suitable for conventional glaucoma surgery. It is also a good option for patients who are not suitable for medical treatment. MIGS is a good option for patients who are not suitable for conventional glaucoma surgery. It is also a good option for patients who are not suitable for medical treatment. MIGS is a good option for patients who are not suitable for conventional glaucoma surgery. It is also a good option for patients who are not suitable for medical treatment. MIGS is a good option for patients who are not suitable for conventional glaucoma surgery. It is also a good option for patients who are not suitable for medical treatment. MIGS is a good option for patients who are not suitable for conventional glaucoma surgery. It is also a good option for patients who are not suitable for medical treatment. MIGS is a good option for patients who |
| **GPT-4 Evaluation** | | | | | |
| **Overall score** | 95 | 95 | 75 | 95 | 40 |

**Supplementary Table 1. Responses generated by GPT-4 for testing dataset questions**

| No. | Testing Dataset Questions | Response generated by GPT-4 |
|---|---|---|
| 1 | **Can I choose to have multifocal lens for my cataract surgery?** | Yes, you can choose to have multifocal lenses implanted during cataract surgery, provided you are a suitable candidate. Multifocal intraocular lenses (IOLs) are designed to provide clear vision at different distances – near, intermediate, and far. They can reduce or eliminate the need for glasses or contact lenses after cataract surgery.<br><br>However, not everyone is a good candidate for multifocal IOLs. Your ophthalmologist will evaluate factors such as your eye health, lifestyle, and vision needs to determine if multifocal lenses are appropriate for you. It's important to discuss the potential benefits and drawbacks, as multifocal lenses can sometimes cause visual disturbances like glare and halos around lights.<br><br>It's essential to have a thorough discussion with your eye surgeon about your expectations and to understand the risks and benefits before deciding on multifocal IOLs for your cataract surgery. |

| **Can I be glasses free after my cataract surgery?** | Yes, many patients can achieve glasses-free vision after cataract surgery, especially for distance vision. During cataract surgery, the cloudy natural lens of the eye is replaced with an artificial intraocular lens (IOL). The power of the IOL can be selected to correct refractive errors like nearsightedness, farsightedness, and astigmatism.

However, whether you will be completely independent of glasses after surgery depends on several factors, including the type of IOL implanted, your eye health, the presence of any other vision issues, and your personal vision goals. For example, if you receive a monofocal IOL, you may still need glasses for reading or close work. Multifocal or accommodating IOLs can provide a broader range of vision correction, potentially reducing your dependence on glasses for both near and distance vision.

It's important to discuss your expectations with your ophthalmologist before surgery to understand what kind of visual outcome is realistic for you. |
|---|---|

| # | Question | Answer |
|---|---|---|
| 3 | **What are the different types of cataracts and how do they affect my vision?** | Cataracts are a common eye condition where the lens of the eye becomes cloudy, leading to a decrease in vision. There are several types of cataracts, each affecting vision in specific ways:1. Nuclear Sclerotic Cataracts: This is the most common type. It involves the gradual hardening and yellowing of the central zone of the lens, known as the nucleus. Over time, this can lead to a general decrease in vision clarity and can cause myopia, or nearsightedness. In advanced stages, the lens can turn brown, significantly affecting color perception.2. Cortical Cataracts: These cataracts start at the outer edge of the lens and extend inward in a spoke-like fashion. They can cause problems with glare and light scatter, making it difficult to see in bright light and causing halos around lights at night.3. Posterior Subcapsular Cataracts (PSC): This type forms at the back of the lens, right in the path of light. PSC cataracts can cause significant glare and halo effects, as well as difficulty reading, as they often affect reading vision earlier than other types of cataracts.4. Congenital Cataracts: Present at birth or developing during childhood, these can be hereditary or associated with some intrauterine infection or trauma. They vary in severity; some may not affect vision much, while others can lead to significant visual impairment.5. Secondary Cataracts: These develop as a result of other medical conditions, such as diabetes, or from exposure to toxic substances, certain drugs (like steroids), ultraviolet light, or radiation.6. Traumatic Cataracts: These form after an injury to the eye. The cataract may develop immediately after the trauma or several years later.The impact on vision depends on the type, size, and location of the cataract. Common symptoms include blurry vision, difficulty with glare, reduced intensity of colors, difficulty seeing at night, and sometimes double vision in the affected eye. If cataracts significantly impair vision, they can be surgically removed, with the cloudy lens replaced by an artificial one to restore clear vision. |
| 4 | **What are toric intraocular lenses?** | Toric intraocular lenses (IOLs) are a type of lens implant designed to correct astigmatism at the time of cataract surgery. Astigmatism is a common refractive error caused by an irregular curvature of the cornea or lens, leading to blurred or distorted vision. Toric IOLs have different powers in different meridians of the lens to compensate for the uneven curvature of the cornea. They are positioned in a specific orientation inside the eye to align with the astigmatism, thereby providing clearer vision postoperatively. These lenses are tailored to the individual's eye measurements and the degree of astigmatism. |
| 5 | **What is the treatment for keratoconus?** | The treatment for keratoconus varies depending on the severity of the condition. Here are the common approaches:1. **Eyeglasses or Soft Contact Lenses**: For mild keratoconus, eyeglasses or soft contact lenses can help correct the mild refractive errors.2. **Hard Contact Lenses**: If the condition progresses, rigid gas permeable (RGP) contact lenses may be prescribed. These lenses are more effective than soft lenses at correcting the irregular cornea because they maintain their shape, allowing the eye's surface to take on the shape of the lens.3. **Hybrid Contact Lenses**: These lenses have a rigid center with a soft outer ring, which can be more comfortable than traditional hard lenses.4. **Scleral and Semi-Scleral Lenses**: These are larger contact lenses that rest on the sclera, the white part of the eye, and vault over the cornea. They can be more stable and comfortable for people with keratoconus.5. **Collagen Cross-Linking (CXL)**: This is a treatment aimed at stopping the progression of keratoconus. It involves the application of vitamin B2 (riboflavin) to the cornea, which is then activated with UV light to strengthen the corneal fibers.6. **Intacs**: These are small curved implantable corneal devices that are surgically placed in the cornea to help flatten the corneal surface.7. **Topography-Guided Conductive Keratoplasty**: This is a non-invasive procedure that uses radio frequency energy to reshape the cornea.8. **Corneal Transplant**: In severe cases, when vision can no longer be corrected with lenses and the cornea becomes dangerously thin or scarred, a corneal transplant may be necessary. There are two types: full-thickness (penetrating keratoplasty) and partial-thickness (lamellar keratoplasty).9. **Custom Soft Contact Lenses**: Some manufacturers offer custom soft contact lenses for keratoconus, though these may not be suitable for all patients.10. **Prosthetic Lenses**: In cases where contact lens wear is difficult or not providing adequate vision, a prosthetic lens that includes both a white (scleral) and clear (corneal) portion can be an option.The choice of treatment is individualized based on the patient's specific needs and the stage of the disease. Regular monitoring by an |
| 6 | **Why must I continue my eye drops after corneal transplant surgery?** | After a corneal transplant surgery, it is essential to continue using prescribed eye drops for several reasons:1. **Prevention of Rejection:** Your immune system can recognize the new cornea as foreign and may try to reject it. Eye drops, often corticosteroids, help to suppress the immune response and reduce the risk of rejection.2. **Control of Inflammation:** Surgery can induce inflammation. Anti-inflammatory eye drops help to control this response and promote healing.3. **Infection Prevention:** Antibiotic eye drops may be prescribed to prevent bacterial infections, which are a risk after any surgical procedure.4. **Maintaining Graft Clarity:** To ensure the best possible outcome, it's crucial to keep the graft clear and healthy, which is facilitated by the appropriate use of eye drops.5. **Intraocular Pressure (IOP) Control:** Some eye drops may be used to manage IOP, as fluctuations can affect the health of the optic nerve and the success of the transplant.It is important to follow your ophthalmologist's instructions regarding the type, dosage, and duration of eye drop use to ensure the best possible outcome for your corneal transplant. |
| 7 | **What should I do if detergent went into my eye?** | If detergent gets into your eye, it's important to act quickly to minimize irritation and potential damage. Follow these steps:1. Rinse your eye immediately. Use lukewarm tap water or a sterile saline solution if available. Open your eyelids with your fingers and allow the water to flush your eye thoroughly for at least 15 to 20 minutes. Do not use hot water.2. Blink frequently while rinsing to help remove the detergent from your eye.3. Do not rub your eye, as this can cause further irritation or damage.4. After rinsing, if you wear contact lenses and they did not fall out during the rinse, remove them carefully and do not reinsert them until your eye has no more symptoms.5. Seek medical attention. Even if your symptoms improve after rinsing, it's important to consult a healthcare professional or an ophthalmologist to ensure that your eye has not been damaged and to receive further treatment if necessary.6. Follow any additional instructions given by the healthcare provider, which may include using lubricating eye drops or other medication to help soothe the eye.Remember, chemical injuries to the eye can be serious, and prompt action is essential to minimize the risk of long-term damage. |
| 8 | **What is the treatment for dry eyes?** | The treatment for dry eyes can vary depending on the underlying cause and severity of the condition. Here are some common approaches to managing dry eye syndrome:1. **Artificial Tears**: Over-the-counter lubricating eye drops can help alleviate the symptoms of dry eyes.2. **Prescription Eye Drops**: Certain prescription eye drops like cyclosporine (Restasis) or lifitegrast (Xiidra) can help increase tear production.3. **Punctal Plugs**: Tiny silicone plugs can be inserted into the tear ducts to prevent tear drainage, which helps keep the eyes more moist.4. **Warm Compresses and Eyelid Washing**: Applying warm compresses to the eyes and washing the eyelids can help release oil from the glands in the eyelids, improving tear quality.5. **Omega-3 Fatty Acid Supplements**: Dietary supplements containing omega-3 fatty acids may help improve dry eye symptoms.6. **Medications**: Adjusting medications that may be contributing to dry eye symptoms can be helpful.7. **Environmental Modifications**: Using a humidifier, avoiding wind and dry environments, and taking breaks during tasks that require prolonged visual attention can help.8. **Specialty Eyewear**: Glasses or goggles designed to create a moisture chamber around the eye can be beneficial in certain environments.9. **Tear Conservation**: In some cases, conserving the patient's own tears is a strategy, which may involve blocking the tear ducts through which tears normally drain.10. **Eyelid Treatments**: If eyelid conditions, such as blepharitis, are contributing to dry eyes, treating those conditions can help.11. **Surgery**: In severe cases, surgical options may be considered to correct eyelid abnormalities or to permanently close the tear ducts.It's important for individuals with dry eyes to consult with an eye care professional to determine the most appropriate treatment for their specific situation. |
| 9 | **What are the signs of retinal detachment?** | Signs of retinal detachment include:<br><br>1. Sudden appearance of floaters, which are small specks or threads that drift through your field of vision.<br>2. Flashes of light in one or both eyes.<br>3. A shadow or curtain effect over a portion of your visual field that develops as the detachment progresses.<br>4. Blurred vision.<br>5. A sudden decrease in visual acuity.<br><br>If you experience any of these symptoms, it's important to seek immediate medical attention, as retinal detachment is a serious condition that can lead to permanent vision loss if not treated promptly. |

| | How is diabetic retinopathy diagnosed? | Diabetic retinopathy is diagnosed through a comprehensive eye examination by an ophthalmologist or optometrist. Key diagnostic procedures include:1. **Visual Acuity Test**: This test measures the patient's ability to see at various distances using an eye chart.2. **Dilated Eye Exam**: Drops are placed in the eyes to widen the pupils, allowing the doctor to examine the retina and optic nerve for signs of damage and other eye problems.3. **Fundus Photography**: This involves taking photographs of the retina to document the current state of the retinal blood vessels and to detect changes over time.4. **Optical Coherence Tomography (OCT)**: This non-invasive imaging test provides cross-sectional images of the retina, revealing its thickness, which can indicate the presence of fluid or swelling (edema).5. **Fluorescein Angiography**: In this test, a fluorescent dye is injected into the bloodstream. The dye highlights the blood vessels in the back of the eye so they can be photographed to detect leaking blood vessels, retinal changes, or new blood vessel growth.Early detection and management of diabetic retinopathy are crucial to prevent vision loss. Patients with diabetes should have regular eye exams as recommended by their eye care professional. |
|---|---|---|
| 10 | | |
| 11 | **My doctor says I have age related macular degeneration and I have to get injections into my eye. What are the risks?** | Intravitreal injections, which are commonly used to treat age-related macular degeneration (AMD), involve injecting medication directly into the vitreous near the retina of the eye. These injections can help slow the progression of the disease, particularly for the wet form of AMD, by delivering medication that inhibits the growth of abnormal blood vessels and reduces fluid leakage. However, as with any medical procedure, there are potential risks associated with intravitreal injections, including: 1. **Infection (Endophthalmitis):** This is a rare but serious complication that can lead to severe vision loss. It occurs when bacteria enter the eye, leading to an infection. 2. **Retinal detachment:** This is another rare complication where the retina separates from the back of the eye. 3. **Cataract formation:** The procedure may accelerate cataract development in some patients, particularly if the lens is accidentally touched during the injection. 4. **Increased intraocular pressure (IOP):** The injection can temporarily increase the pressure inside the eye, which usually normalizes within a few hours. 5. **Bleeding:** There may be bleeding on the surface of the eye or within the eye, which often resolves on its own. 6. **Eye inflammation:** The eye may become red and inflamed after the injection. 7. **Floaters:** Some patients may notice new floaters in their vision after the injection. 8. **Allergic reactions:** Rarely, patients may have an allergic reaction to the medication used. 9. **Pain or discomfort:** Some patients may experience pain or discomfort in the eye after the injection. It's important to discuss the potential risks and benefits of intravitreal injections with your ophthalmologist to fully understand the implications of the treatment. Your doctor will take precautions to minimize risks, such as using antiseptic techniques to reduce the risk of infection and monitoring your eye after the procedure. If you experience any unusual symptoms after an injection, such as increased pain, decreased vision, or increasing redness, you should contact your ophthalmologist immediately. |
| 12 | **How do I use the amsler grid?** | To use the Amsler grid, follow these steps:1. **Position the Grid**: Place the Amsler grid at eye level in a well-lit room. The standard distance is about 14 to 16 inches away from your eyes.2. **Wear Your Glasses**: If you normally wear reading glasses or have glasses for distance, make sure you wear them while using the grid.3. **Cover One Eye**: Cover one eye with your hand without applying pressure to the eyelid. You will test each eye separately.4. **Focus on the Center Dot**: Look directly at the center dot on the grid with the uncovered eye. Keep your focus on this dot throughout the test.5. **Observe the Grid**: While looking at the center dot, take note of the appearance of the grid lines and the squares. They should appear straight and the squares should be the same size throughout.6. **Check for Distortions**: Look for any areas where the grid lines appear wavy, blurry, or are missing. Also, notice if any of the squares appear to be different in size or shape compared to the others.7. **Repeat with the Other Eye**: Cover the other eye and repeat the process.8. **Record Any Changes**: If you notice any changes from the last time you used the grid or any distortions, mark them on the grid if possible and contact your eye care professional promptly.The Amsler grid is a tool used to detect vision problems resulting from damage to the macula (the central part of the retina) or the optic nerve. It is often used by individuals with macular degeneration, a common eye condition that can cause vision loss. Regular monitoring with the Amsler grid can help detect changes in vision early on. |
| 13 | **I can see double images. Is it dangerous?** | Double vision, also known as diplopia, can be a symptom of various conditions, some of which may require urgent medical attention. It can occur in one eye (monocular) or both eyes (binocular). Monocular diplopia persists even when the other eye is closed, while binocular diplopia resolves when one eye is closed.Causes of diplopia can range from relatively benign conditions to serious health issues. Some possible causes include:- Refractive errors (like astigmatism)- Dry eye syndrome- Cataracts- Strabismus (misalignment of the eyes)- Diabetes- Neurological conditions (such as multiple sclerosis or a stroke)- Trauma to the head or eye- Cranial nerve palsies- Myasthenia gravis- Brain tumors or other massesIt's important to determine the underlying cause of your double vision. If you experience sudden double vision, it's crucial to seek medical attention promptly, as it could indicate a serious underlying condition, especially if it's accompanied by other symptoms like headache, weakness, or numbness.An ophthalmologist or neurologist will be able to assess your symptoms, perform necessary tests, and provide a diagnosis and appropriate treatment plan. |
| 14 | **Tell me about botox injections for eyelid twitching.** | Botulinum toxin, commonly known as Botox, is a neurotoxic protein that is used for various medical and cosmetic purposes. When it comes to eyelid twitching, or blepharospasm, Botox injections can be an effective treatment. Blepharospasm is a condition where there is an abnormal, involuntary blinking or spasm of the eyelids. Botox injections work by temporarily paralyzing the muscles that are causing the spasms. The toxin blocks the release of the neurotransmitter acetylcholine, which is responsible for muscle contraction, thus relaxing the muscles and stopping the twitching. The procedure is minimally invasive and is typically done in an outpatient setting. The effects of Botox for eyelid twitching usually last for about three to four months, after which the treatment can be repeated. Most patients experience significant relief from their symptoms following the injections. It's important to have Botox injections administered by a qualified healthcare professional who is experienced in treating blepharospasm to ensure the correct muscles are targeted and to minimize the risk of side effects. Possible side effects can include drooping of the eyelid, dry eye, or double vision, but these are typically temporary. Patients considering Botox injections for eyelid twitching should discuss the potential risks and benefits with their ophthalmologist or a neuro-ophthalmologist. |
| 15 | **What are the tests needed to confirm that I have myasthenia gravis?** | To confirm a diagnosis of myasthenia gravis, a combination of the following tests and evaluations may be used:1. **Acetylcholine Receptor Antibody Test**: This blood test checks for the presence of antibodies that may affect the neurotransmitter acetylcholine, which is crucial for communication between nerves and muscles. Most people with myasthenia gravis have an elevated level of these antibodies.2. **Anti-MuSK Antibody Test**: Some individuals who have myasthenia gravis have antibodies against a muscle-specific kinase (MuSK). Testing for these antibodies can help diagnose cases that don't have acetylcholine receptor antibodies.3. **Edrophonium Test**: This test involves the administration of a drug called edrophonium, which can temporarily relieve weakness in people with myasthenia gravis. The test can provide quick, though not definitive, evidence for myasthenia gravis.4. **Electromyography (EMG)**: An EMG measures the electrical activity of muscles. A special form of EMG, called repetitive nerve stimulation, can show a decline in the muscle's ability to respond to nerve signals in myasthenia gravis.5. **Single-Fiber EMG (SFEMG)**: This is a more sensitive version of EMG and can detect problems with nerve-to-muscle transmission that are characteristic of myasthenia gravis.6. **Computed Tomography (CT) or Magnetic Resonance Imaging (MRI)**: These imaging tests can be used to check for a thymoma (a tumor of the thymus gland), which is associated with myasthenia gravis.7. **Pulmonary Function Tests**: These tests can assess whether myasthenia gravis is affecting your breathing by measuring your respiratory muscle strength.A neurologist or a specialist in neuromuscular disorders typically conducts these tests to confirm a diagnosis of myasthenia gravis. It's important to note that not all tests may be necessary for every patient, and the choice of tests can depend on the individual's symptoms and clinical presentation. |
| 16 | **What is the management of hemifacial spasm?** | The management of hemifacial spasm typically involves both non-surgical and surgical options:1. **Medications**: Anticonvulsants such as carbamazepine or gabapentin may be used to alleviate symptoms, although they are often not very effective for hemifacial spasm.2. **Botulinum toxin injections (Botox)**: This is the most common first-line treatment for hemifacial spasm. Injections of botulinum toxin temporarily paralyze the muscles, reducing or eliminating the spasm. The effects usually last for several months, and the treatment must be repeated.3. **Microvascular decompression (MVD)**: This is a surgical procedure that may be recommended if botulinum toxin injections are ineffective or not tolerated. MVD involves relieving pressure on the facial nerve by repositioning or removing blood vessels that are compressing it. This is done through a small opening in the skull behind the ear. MVD has a high success rate and can potentially cure hemifacial spasm, but it carries risks associated with surgery and general anesthesia.4. **Other surgical options**: In rare cases where MVD is not possible or successful, other neurosurgical procedures might be considered.The choice of treatment depends on the severity of the symptoms, the patient's preferences, and their overall health. It's important for patients to discuss the potential benefits and risks of each treatment option with their healthcare provider. |

| # | Question | Answer |
|---|---|---|
| 17 | **What is the surgery procedure like for blepharoplasty?** | Blepharoplasty, also known as eyelid surgery, is a procedure to correct droopy eyelids by removing excess skin, muscle, and sometimes fat. The surgery can be performed on the upper eyelids, lower eyelids, or both. Here's a general outline of the procedure:**Preparation:**- Prior to surgery, a comprehensive eye examination and possibly a physical examination are conducted.- Patients may be asked to stop taking medications that can increase bleeding, such as aspirin, anti-inflammatory drugs, and certain herbal supplements.- The surgeon will mark the areas of the eyelids where incisions will be made.**Anesthesia:**- Blepharoplasty is typically performed under local anesthesia with sedation, but general anesthesia may be used in some cases.**Surgical Procedure:**- **Upper Eyelid Blepharoplasty:** The surgeon makes an incision along the natural crease of the upper eyelid. Through this incision, excess skin and fat are removed, and the muscle may be tightened. The incision is then closed with sutures, skin adhesive, or surgical tape.- **Lower Eyelid Blepharoplasty:** The incision can be made just below the lashes in the natural crease of the eye or inside the lower eyelid (transconjunctival approach). Excess fat, and sometimes excess skin and muscle, are removed or redistributed. The incision is then closed similarly to the upper eyelid.**Recovery:**- After the surgery, ointment may be applied to the eyes, and a cold compress may be recommended to reduce swelling and bruising.- The patient will receive instructions on how to care for the eyes, possible signs of complications, and when to follow up with the surgeon.- Stitches are typically removed within a week after surgery.- Swelling and bruising should subside within one to two weeks.**Risks and Complications:**- As with any surgery, there are risks, including infection, bleeding, scarring, inability to close the eyes, dry eyes, and, in rare cases, vision loss.It's important for patients to have realistic expectations and to discuss potential risks and benefits with their surgeon before undergoing blepharoplasty. |
| 18 | **Can shingles affect my eyes?** | Yes, shingles can affect your eyes. When the varicella-zoster virus, which causes chickenpox and shingles, reactivates, it can lead to a condition called herpes zoster ophthalmicus if it involves the ophthalmic branch of the trigeminal nerve. This can result in a painful rash around the eye and forehead, and it may also involve the eye itself, leading to conjunctivitis, keratitis, uveitis, or even glaucoma. If you suspect shingles affecting your eyes, it's important to seek medical attention promptly to reduce the risk of complications, including vision loss. |
| 19 | **How can my thyroid condition affect my eyes?** | Thyroid eye disease, also known as Graves' ophthalmopathy or thyroid-associated ophthalmopathy, is an autoimmune condition that can affect your eyes if you have a thyroid disorder, particularly hyperthyroidism due to Graves' disease. Here are some ways in which your thyroid condition can affect your eyes:1. **Eyelid Retraction**: The upper eyelid may retract higher than normal, and the lower eyelid may pull downwards. This can cause a staring appearance or excessive white (sclera) to be visible above or below the iris.2. **Exophthalmos (Proptosis)**: The eyes may bulge forward. This is due to inflammation and swelling of the muscles and tissues behind the eye, which can push the eyeball forward.3. **Dry Eye**: Symptoms include irritation, a gritty feeling, and dryness. This can occur due to eyelid retraction and proptosis, which can prevent the eyelids from closing properly, leading to increased exposure of the cornea.4. **Redness and Swelling**: Inflammation can cause redness and swelling of the eyes and eyelids.5. **Double Vision (Diplopia)**: Swelling and stiffening of the eye muscles can lead to misalignment of the eyes and double vision.6. **Vision Loss**: In severe cases, the optic nerve can be compressed due to the swollen tissues and muscles in the orbit, leading to vision loss.It's important to manage your thyroid condition and work closely with both your endocrinologist and ophthalmologist to monitor and treat any eye-related symptoms. Treatments can range from lubricating eye drops to, in severe cases, surgery to address the changes in the orbit and relieve pressure on the optic nerve. |
| 20 | **Why do I have a lump on my eyelid?** | A lump on the eyelid can be caused by several conditions, some of which are benign and others that may require medical attention. Here are a few possible causes:1. Chalazion: This is a painless lump that can develop when the oil gland (meibomian gland) in the eyelid becomes blocked. It often starts as a small, red, and tender area before turning into a lump.2. Stye (Hordeolum): A stye is a red, painful lump near the edge of the eyelid that may look like a boil or a pimple. It is caused by an infection of the eyelash follicle or oil gland.3. Cysts: Eyelid cysts, such as an epidermal inclusion cyst, can cause a lump. These are typically benign and filled with keratin.4. Xanthelasma: These are yellowish plaques that occur most commonly near the inner canthus of the eyelid. They are not painful and are associated with high levels of lipids in the blood.5. Infections: Bacterial or viral infections can lead to the formation of lumps on the eyelid.6. Neoplasms: Both benign and malignant tumors can appear as lumps on the eyelids. Examples include papillomas, basal cell carcinoma, squamous cell carcinoma, and melanoma.If you notice a lump on your eyelid, it's important to consult with an ophthalmologist or a healthcare provider to determine the cause and appropriate treatment. |
| 21 | **What are the risks of LASIK surgery?** | LASIK (Laser-Assisted In Situ Keratomileusis) surgery is generally safe, but like any surgical procedure, it carries certain risks and potential complications. Some of the risks associated with LASIK surgery include:1. **Dry Eyes**: LASIK surgery can cause a temporary decrease in tear production, leading to dry eye syndrome, which can be uncomfortable and may temporarily affect vision.2. **Glare, Halos, and Double Vision**: After surgery, patients may experience increased light sensitivity, glare, halos around bright lights, or double vision. These symptoms usually diminish over time but can be permanent in some cases.3. **Undercorrections**: If the laser removes too little tissue from the eye, the desired vision correction may not be achieved. This is more common in people who are nearsighted. Additional LASIK surgery may be required to remove more tissue.4. **Overcorrections**: Conversely, if the laser removes too much tissue from the eye, it can be more difficult to fix than undercorrections.5. **Astigmatism**: Uneven tissue removal can cause astigmatism, which may require additional surgery, glasses, or contact lenses.6. **Flap Problems**: Folding back or removing the flap from the front of the eye during surgery can lead to complications such as infection and excess tears. The outermost corneal tissue layer may also grow abnormally underneath the flap during the healing process.7. **Regression**: Over time, some patients experience a gradual return toward their original prescription, a phenomenon known as regression.8. **Vision Loss or Changes**: In rare instances, surgical complications can result in vision loss. Some patients may also not see as sharply or clearly as they did before the procedure.9. **Keratoconus or Ectasia**: There is a risk of developing keratoconus or corneal ectasia, where the cornea begins to bulge outward after surgery, due to weakening of the corneal structure.It's important for individuals considering LASIK to discuss these risks with their ophthalmologist, who can provide a thorough assessment and help determine if they are good candidates for the surgery. Preoperative screening can identify factors that may increase the likelihood of complications, and taking these into account can help minimize risks. |

| | What are the signs and symptoms that suggests my child has myopia? | If your child has myopia, also known as nearsightedness, you may notice the following signs and symptoms:1. Squinting: Your child may squint frequently in an attempt to see distant objects more clearly.2. Sitting Close: They may prefer to sit closer to the television, screen, or the front of the classroom to see better.3. Eye Strain: Complaining of tired eyes or eye strain, especially after prolonged periods of needing to focus on distant objects.4. Headaches: Frequent headaches can occur due to the strain of trying to focus on objects in the distance.5. Difficulty Seeing at a Distance: Your child may have trouble clearly seeing the board at school or objects that are far away.6. Blinking Excessively: Excessive blinking might be an attempt to clear up their blurred vision.7. Rubbing Eyes: They may rub their eyes frequently, which can be a sign of eye fatigue or strain.If you notice these symptoms, it's important to schedule an eye examination with an optometrist or ophthalmologist to determine if your child has myopia and to discuss potential corrective options such as glasses or contact lenses. Early detection and correction can help your child perform better in school and prevent the progression of myopia. |
|---|---|---|
| 22 | | |
| 23 | **My child has myopia, what are the available treatments to slow down the progression of myopia?** | To slow down the progression of myopia in children, several treatment options are available:1. **Atropine Eye Drops**: Low-dose atropine eye drops have been shown to slow myopia progression in some children. The exact mechanism is not fully understood, but it's believed to work by inhibiting the eye's focusing mechanism.2. **Multifocal Contact Lenses**: These contact lenses have different zones for near and far vision. They can reduce the stimulus for myopia progression by correcting peripheral vision and reducing the strain of focusing on close objects.3. **Multifocal Eyeglasses**: Similar to multifocal contact lenses, these glasses have multiple lens powers to help focus on objects at various distances, which may help slow the progression of myopia.4. **Orthokeratology (Ortho-K)**: This involves wearing specially designed rigid gas-permeable contact lenses overnight to temporarily reshape the cornea, which can reduce the need for glasses or contact lenses during the day and may slow myopia progression.5. **Outdoor Time**: Increasing the amount of time spent outdoors during daylight hours has been associated with a reduced risk of developing myopia and may slow its progression in children.6. **Lifestyle Changes**: Encouraging breaks from close work, maintaining a proper reading distance, and ensuring adequate lighting can help reduce eye strain.It's important to consult with an ophthalmologist or optometrist to determine the most appropriate treatment for your child, as individual responses to these treatments can vary. Regular eye examinations are crucial for monitoring myopia progression and adjusting treatment as needed. |
| 24 | **What causes myopia?** | Myopia, also known as nearsightedness, is a common vision condition where you can see objects near to you clearly, but objects farther away are blurry. It occurs when the eyeball is too long relative to the focusing power of the cornea and lens of the eye, or when the cornea and/or lens is too curved for the length of the eyeball. This causes light rays to focus at a point in front of the retina, rather than directly on its surface.Myopia can be due to a combination of genetic and environmental factors. Here are some potential causes and contributing factors:1. Genetics: A family history of myopia is a significant risk factor. If one or both parents are myopic, the chances of developing myopia are increased.2. Environmental factors: Activities that require prolonged near vision tasks, such as reading, using computers, or other electronic devices, can contribute to the development of myopia, especially if done excessively without regular breaks.3. Reduced time outdoors: Some research suggests that spending less time outdoors may increase the risk of developing myopia, possibly due to less exposure to natural light and fewer opportunities for focusing on distant objects.4. Age: Myopia typically develops in childhood or adolescence and can progress until the eye stops growing.5. Ethnicity: Myopia is more prevalent in certain ethnic groups, such as East Asian populations.Understanding the causes of myopia can help in developing strategies to manage its progression, such as regular eye exams, appropriate corrective lenses, and lifestyle modifications that include taking breaks during close work and spending more time outdoors. |
| 25 | **What are the treatment options for amblyopia?** | Treatment options for amblyopia, also known as lazy eye, typically include:1. **Corrective Eyewear**: Glasses or contact lenses are used to correct refractive errors such as nearsightedness, farsightedness, or astigmatism that may be contributing to amblyopia.2. **Patching**: Covering the stronger eye with an eye patch forces the brain to recognize the visual input from the amblyopic eye, thereby stimulating its use and improving vision.3. **Atropine Eye Drops**: These can be used to blur the vision in the stronger eye, which, like patching, encourages the use of the weaker eye.4. **Vision Therapy**: This involves a series of visual tasks and exercises prescribed and supervised by an eye care professional, designed to improve visual abilities and processing.5. **Surgery**: In cases where amblyopia is caused by issues such as cataracts or strabismus (misaligned eyes), surgical intervention may be necessary to correct the underlying problem.Early detection and treatment of amblyopia are crucial for the best outcomes, as the condition is most effectively treated during childhood when the visual system is still developing. |
| 26 | **What are eye problems that can occur in premature babies?** | Premature babies are at risk for several eye problems, the most notable being Retinopathy of Prematurity (ROP). ROP occurs when abnormal blood vessels grow and spread throughout the retina, the tissue that lines the back of the eye. These abnormal vessels are fragile and can leak, leading to retinal scarring or detachment.Other potential eye problems in premature babies include:1. Myopia (nearsightedness): Premature babies have a higher risk of developing myopia later in childhood.2. Strabismus (misaligned eyes): Muscle control of the eyes may not develop properly in premature infants, leading to strabismus.3. Amblyopia (lazy eye): If one eye develops better vision than the other, the weaker eye might become amblyopic.4. Refractive errors: Besides myopia, premature babies can also develop other refractive errors like hyperopia (farsightedness) or astigmatism.5. Visual-motor impairment: Premature infants may have difficulties with eye movements and hand-eye coordination.Regular eye examinations by a pediatric ophthalmologist are crucial for premature babies to monitor their visual development and address any issues early on. |
| 27 | **What is the management of cataracts in children?** | Management of cataracts in children, also known as pediatric cataracts, involves a combination of surgical and non-surgical approaches tailored to the child's age, the severity of the cataract, and any associated conditions. Here is a general outline of the management strategies:1. **Surgical Intervention**: The primary treatment for visually significant cataracts in children is surgery to remove the cloudy lens. The timing of surgery is crucial and often depends on whether the cataract is congenital (present at birth) or acquired, unilateral (affecting one eye) or bilateral (affecting both eyes), and the degree to which the cataract is interfering with visual development.   - **Congenital Cataracts**: Early surgery is typically recommended, often within the first few months of life, to minimize the risk of amblyopia (lazy eye). For unilateral cataracts, surgery may be performed as early as 4-6 weeks of age, while for bilateral cataracts, surgery may be staged for each eye and can be done within the first few months of life.   - **Acquired Cataracts**: The timing of surgery for acquired cataracts depends on the cause and the impact on vision.2. **Intraocular Lens (IOL) Implantation**: During cataract surgery, the natural lens is removed, and in many cases, an artificial lens called an intraocular lens (IOL) is implanted to help focus light onto the retina. The decision to implant an IOL in a child depends on the child's age, the health of the eye, and other factors. In very young children, the decision to implant an IOL may be deferred, and contact lenses or glasses may be used to correct vision after surgery.3. **Optical Correction**: After cataract removal, if an IOL is not implanted, the child will need other forms of visual correction such as glasses or contact lenses to focus light properly onto the retina.4. **Amblyopia Therapy**: If a cataract has caused amblyopia, additional treatments such as patching the stronger eye to force the use of the affected eye may be necessary. This therapy is critical for visual development and is most effective when started early.5. **Monitoring and Follow-up**: Regular follow-up with a pediatric ophthalmologist is essential to monitor the child's visual development, the position and condition of the IOL if present, and to check for complications such as glaucoma, which |
| 28 | **What should I do if my 2-year-old has watery eyes?** | If your 2-year-old has watery eyes, it could be due to a variety of reasons, such as a blocked tear duct, an infection like conjunctivitis, allergies, or simply irritation from wind or dust. Here's what you can do:1. **Observe**: Look for any signs of infection such as redness, swelling, or discharge. Also, check if there is any associated fever, irritability, or rubbing of the eyes.2. **Clean**: Gently clean the eyelids with a clean, warm, damp cloth to remove any crusts or discharge.3. **Avoid Irritants**: Keep your child away from potential irritants such as smoke, dust, or strong winds.4. **Tear Duct Massage**: If a blocked tear duct is suspected, you can gently massage the area near the corner of the eyes to help open it up. Your pediatrician can show you the proper technique.5. **Seek Medical Advice**: If the watery eyes persist, or if you notice any signs of infection or if your child seems to be in pain, consult your pediatrician. They may refer you to a pediatric ophthalmologist if necessary.6. **Follow Recommendations**: If an infection is diagnosed, follow the treatment plan prescribed by your healthcare provider, which may include antibiotic drops or ointment.Remember, it's important to consult a healthcare professional for an accurate diagnosis and appropriate treatment. |
| 29 | **Why do I have eye redness, itch, and discharge?** | Eye redness, itchiness, and discharge can be symptoms of several conditions, including:1. Conjunctivitis (Pink Eye): This is an inflammation or infection of the conjunctiva, the transparent membrane that lines your eyelid and covers the white part of your eyeball. It can be caused by viruses, bacteria, allergens, or irritants.2. Allergies: Allergic reactions to pollen, dust, pet dander, or other allergens can cause these symptoms.3. Dry Eye Syndrome: This occurs when your eyes do not produce enough tears or the right quality of tears, leading to inflammation and irritation.4. Blepharitis: Inflammation of the eyelids can lead to redness, itching, and discharge.5. Contact Lens-Related Issues: Overwear of contact lenses, poor hygiene, or an allergic reaction to contact lens solutions can cause these symptoms.6. Corneal Ulcer: An open sore on the cornea, often due to infection, can result in redness, pain, discharge, and reduced vision.7. Environmental Irritants: Smoke, fumes, or chemical exposure can irritate the eyes.If you are experiencing these symptoms, it is important to consult with an eye care professional for an accurate diagnosis and appropriate treatment. |
| 30 | **What are reasons for increased sensitivity to light?** | Increased sensitivity to light, also known as photophobia, can be caused by a variety of conditions, including:1. Ocular Conditions:   - Inflammation inside the eye (uveitis)   - Corneal abrasion or ulcer   - Conjunctivitis (pink eye)   - Cataracts   - Glaucoma, particularly acute angle-closure glaucoma   - Scleritis or episcleritis   - Dry eye syndrome2. Refractive Surgery:   - Postoperative sensitivity after procedures like LASIK or PRK3. Migraines:   - Light sensitivity is a common symptom before, during, or after a migraine headache.4. Neurological Conditions:   - Meningitis   - Encephalitis   - Brain injury or concussion5. Medications:   - Certain medications can increase light sensitivity as a side effect.6. Infections:   - Viral, bacterial, or fungal eye infections7. Systemic Diseases:   - Lupus   - Rheumatoid arthritis   - Fibromyalgia8. Genetic Conditions:   - Albinism, which reduces the amount of pigment in the eyes9. Withdrawal from Dark Adaptation:   - Prolonged time spent in dark environments can temporarily increase light sensitivity when re-exposed to normal light.10. Contact Lens Complications:   - Overwear or poorly fitting contact lenses can lead to increased sensitivity.11. Blepharospasm:   - A condition where there is an abnormal, involuntary blinking or spasm of the eyelids.If someone experiences a sudden increase in light sensitivity, it is important to consult an eye care professional, as it may indicate a more serious underlying condition that requires prompt attention. |
| 31 | **What are some reasons why I have sudden loss of vision in my left eye?** | Sudden loss of vision in one eye can be due to several serious conditions, and it is important to seek immediate medical attention. Some possible causes include:1. Retinal detachment: This occurs when the retina separates from the back of the eye, leading to a sudden, dramatic loss of vision.2. Retinal artery occlusion: This is caused by a blockage in one of the arteries that supply blood to the retina. It can lead to sudden, painless vision loss.3. Retinal vein occlusion: Similar to artery occlusion, but it involves a vein. This can lead to a sudden loss of vision due to the buildup of pressure from the blocked vein.4. Optic neuritis: Inflammation of the optic nerve can cause sudden vision loss, often accompanied by pain on eye movement.5. Glaucoma: Acute angle-closure glaucoma can lead to a sudden increase in intraocular pressure, resulting in pain, redness, and vision loss.6. Vitreous hemorrhage: Bleeding into the vitreous body of the eye can obscure vision suddenly.7. Macular hole or macular pucker: These conditions affect the central part of the retina and can cause sudden changes in vision.8. Migraine with aura: Some people experience a temporary loss of vision in one eye before or during a migraine headache.9. Stroke or transient ischemic attack (TIA): A blockage or bleed in the arteries supplying the brain can affect vision if the visual cortex or the pathways leading to it are involved.10. Giant cell arteritis (temporal arteritis): This is an inflammatory condition of the arteries that can lead to sudden vision loss, typically in older adults.Immediate evaluation by an ophthalmologist or a visit to the emergency room is crucial for any sudden loss of vision, as prompt treatment may be necessary to prevent permanent vision loss. |
| 32 | **Why do straight lines look wavy and distorted to me?** | Wavy or distorted straight lines can be a symptom of a condition called macular degeneration, which affects the central part of the retina and can lead to vision loss. Another possible cause is a macular hole or macular pucker. It's also characteristic of an ocular condition known as keratoconus, where the cornea becomes cone-shaped. Additionally, this symptom can be associated with an epiretinal membrane or complications from other eye surgeries. It is important to consult an ophthalmologist for a comprehensive eye examination to determine the exact cause and appropriate treatment. |
| 33 | **Can you explain the symptoms of glaucoma?** | Certainly. Glaucoma is a group of eye conditions that damage the optic nerve, which is vital for good vision. This damage is often caused by abnormally high pressure in your eye. The symptoms of glaucoma can vary depending on the type and stage of the condition:1. Open-Angle Glaucoma (Chronic):- Gradual loss of peripheral (side) vision, usually in both eyes- Tunnel vision in the advanced stages2. Angle-Closure Glaucoma (Acute):- Severe eye pain- Nausea and vomiting (accompanying the severe eye pain)- Sudden onset of visual disturbance, often in low light- Blurred vision- Halos around lights- Redness of the eye3. Normal-Tension Glaucoma:- Similar to open-angle glaucoma with gradual peripheral vision loss, but occurs despite normal eye pressure.4. Congenital Glaucoma:- Symptoms are noticeable in infants and include cloudiness of the eye's surface, excessive tearing, or sensitivity to light.It's important to note that in the early stages of open-angle glaucoma, there may be no noticeable symptoms. Regular eye exams are crucial for early detection and treatment to prevent vision loss. |
| 34 | **Why is it that my eye pressure is not high but the doctor says that I have glaucoma?** | Glaucoma is a group of eye conditions that can damage the optic nerve, which is vital for good vision. This damage is often caused by abnormally high pressure in your eye. However, glaucoma can occur even when intraocular pressure (IOP) is within the normal range. This is known as normal-tension glaucoma or low-tension glaucoma. In normal-tension glaucoma, your optic nerve becomes damaged even though your eye pressure is considered normal. The exact reason for this is not fully understood, but it is believed that the optic nerve may be more susceptible to damage in some individuals due to factors like reduced blood flow to the optic nerve or other vulnerabilities in the optic nerve's health. Your doctor may diagnose you with glaucoma based on the characteristic optic nerve damage and visual field loss, even if your eye pressure is not elevated. It's important to follow your doctor's recommendations for treatment and monitoring to help preserve your vision. |

| | **What are the different types of eye drops for treating glaucoma?** | There are several types of eye drops used to treat glaucoma, each working in a different way to lower intraocular pressure (IOP) in the eye. The main categories include:1. **Prostaglandin analogs**: These increase the outflow of aqueous humor from the eye, thereby lowering IOP. Examples include latanoprost, bimatoprost, and travoprost.2. **Beta-blockers**: These reduce the production of aqueous humor. Timolol and betaxolol are common beta-blockers used for glaucoma.3. **Alpha agonists**: These both decrease the production and increase the outflow of aqueous humor. Examples include brimonidine and apraclonidine.4. **Carbonic anhydrase inhibitors (CAIs)**: These eye drops reduce the production of aqueous humor by inhibiting an enzyme called carbonic anhydrase. Dorzolamide and brinzolamide are CAIs used in eye drop form.5. **Rho kinase inhibitors**: A newer class of medication that reduces intraocular pressure by increasing the outflow of aqueous humor through the trabecular meshwork. Netarsudil is an example.6. **Cholinergic or miotic agents**: These increase the outflow of aqueous humor by contracting the ciliary muscle and opening the trabecular meshwork. Pilocarpine is a commonly used agent in this category.7. **Combination drops**: These contain two different types of medication to lower IOP in different ways. Examples include dorzolamide-timolol (a combination of a CAI and a beta-blocker) and brimonidine-timolol (a combination of an alpha agonist and a beta-blocker).The choice of eye drop depends on the individual patient's type of glaucoma, the target IOP, the patient's response to medication, and any other medical conditions they may have. |
|---|---|---|
| 35 | | |
| 36 | **What is the procedure of laser peripheral iridotomy like?** | Laser peripheral iridotomy (LPI) is a procedure used to treat angle-closure glaucoma. Here's a step-by-step overview of what the procedure typically involves:1. **Preparation**: Before the procedure, you'll receive eye drops to prepare the eye. These may include a medication to prevent an increase in intraocular pressure and drops to constrict the pupil (miotics).2. **Anesthesia**: Anesthetic eye drops are applied to numb the eye to ensure you don't feel pain during the procedure.3. **Lens Application**: A special lens is placed on the eye to help the ophthalmologist focus the laser precisely on the iris.4. **Laser Application**: The ophthalmologist uses a laser to create a small hole in the peripheral part of the iris. This hole serves as a new drainage channel to allow the aqueous humor (the fluid within the front part of the eye) to flow from the posterior chamber to the anterior chamber, bypassing the narrowed or blocked area that's causing the angle closure.5. **Monitoring**: After the procedure, you'll be monitored for a short time to ensure that the intraocular pressure doesn't rise and that there are no immediate complications.6. **Post-Procedure Care**: You may be prescribed anti-inflammatory eye drops to reduce inflammation caused by the laser treatment. Your doctor will schedule a follow-up appointment to check the iridotomy's effectiveness and to monitor for any complications.The procedure is typically quick, often completed in a few minutes, and is usually performed on an outpatient basis. It's important to follow your ophthalmologist's instructions before and after the procedure to ensure the best possible outcome. |

| 37 | **When do I have to go for surgery if I have glaucoma?** | The decision to undergo surgery for glaucoma is typically made when other treatments, such as medications or laser therapy, have not been successful in controlling the intraocular pressure (IOP) or when the progression of the disease suggests a high risk of vision loss. Surgery is considered if:1. The IOP remains high despite maximum tolerated medical therapy and/or laser treatment.2. There is evidence of progressive optic nerve damage or visual field loss.3. The patient is not able to tolerate the medications due to side effects.4. The patient is not compliant or has difficulty with the medication regimen.The specific type of surgery will depend on the type of glaucoma, the stage of the disease, and other individual factors. Common surgical options include trabeculectomy, glaucoma drainage implants, and minimally invasive glaucoma surgeries (MIGS). It's essential to have regular check-ups with an ophthalmologist to monitor the condition and discuss the best treatment options. |
| 38 | **What is minimally invasive glaucoma surgery?** | Minimally Invasive Glaucoma Surgery (MIGS) refers to a group of surgical procedures that aim to lower intraocular pressure (IOP) in patients with glaucoma using microscopic-sized equipment and tiny incisions. These procedures are designed to reduce the risk of complications that are associated with traditional glaucoma surgeries such as trabeculectomy or tube shunt implantation. MIGS procedures work by improving the outflow of aqueous humor from the eye, thereby lowering IOP. They are often performed in conjunction with cataract surgery but can also be done as standalone treatments. MIGS devices target different parts of the aqueous humor outflow pathway, such as the trabecular meshwork, Schlemm's canal, or the suprachoroidal space. The benefits of MIGS include a safer profile with fewer complications, a more rapid recovery period, and often a reduction in the need for glaucoma medications postoperatively. However, while MIGS can be very effective for mild to moderate glaucoma, they may not be suitable for all patients, particularly those with more advanced disease who may require a more aggressive IOP reduction. |

| 39 | **What should I expect for the follow-up period after trabeculectomy surgery?** | After trabeculectomy surgery, the follow-up period is critical to ensure the success of the procedure and to monitor for any complications. Here's what you can generally expect:1. **Frequent Visits**: Initially, you will have frequent follow-up visits with your ophthalmologist. These may occur several times in the first week after surgery, then weekly, and less often as time goes on and your eye heals.2. **Medication**: You will likely be prescribed antibiotic and anti-inflammatory eye drops to prevent infection and control inflammation inside the eye.3. **Activity Restrictions**: Your doctor will advise you to avoid strenuous activities, heavy lifting, and swimming to prevent increasing pressure in the eye or introducing bacteria into the eye.4. **Monitoring IOP**: Intraocular pressure (IOP) will be closely monitored to ensure the surgery has successfully lowered it. Adjustments to medications or further interventions may be necessary if the IOP is not at the desired level.5. **Bleb Management**: The success of the surgery often depends on the formation and maintenance of a filtering bleb. Your doctor will check the bleb to ensure it is functioning properly.6. **Vision Changes**: Your vision may fluctuate during the first few weeks after surgery. This is normal and should stabilize over time.7. **Suture Adjustment**: Sometimes, sutures used in the surgery are adjustable. Your doctor may perform suture lysis or adjustment in the office to help regulate the flow of fluid from the eye.8. **Complications**: Signs of potential complications such as infection, cataract formation, or bleb failure will be assessed. Symptoms like severe pain, redness, or a sudden drop in vision should be reported immediately.9. **Long-term Monitoring**: Even after the initial healing period, regular check-ups are important to monitor the health of your eye and the long-term success of the surgery.The exact follow-up schedule and care plan will be personalized to your specific situation, so it's important to follow your surgeon's instructions and attend all scheduled appointments. |
|---|---|---|

| 40 | **What is selective laser trabeculoplasty?** | Selective Laser Trabeculoplasty (SLT) is a type of laser surgery used to lower intraocular pressure (IOP) in glaucoma. It is called "selective" because it primarily targets melanin-containing cells in the trabecular meshwork of the eye, which is the structure responsible for draining aqueous humor (the fluid inside the eye) to maintain proper eye pressure. During the procedure, a laser is used to apply a series of tiny, low-energy laser spots to the trabecular meshwork. This stimulates a biological response that improves the drainage of fluid through the meshwork, thereby reducing intraocular pressure. SLT is typically a quick, outpatient procedure and can be an alternative or adjunctive treatment to glaucoma medications. One of the advantages of SLT is that it has a favorable safety profile, with minimal side effects and complications. It can also be repeated if necessary, although the effect may diminish with subsequent treatments. |

**Supplementary Material Table 2. Responses generated by all 5 fine-tuned LLMs for the testing dataset questions, as well as GPT-4 evaluation of each response**

| No. | Test Questions | Responses generated by each LLM: | GPT-3.5 | LLAMA2-13B-Chat | LLAMA2-7b-Chat | LLAMA2-13b | LLAMA2-7b |
|---|---|---|---|---|---|---|---|
| 1 | Can I choose to have multifocal lens for my cataract surgery? | | Yes, you can choose to have multifocal lens for your cataract surgery. Multifocal lens can help you see better at different distances, and reduce your dependence on glasses. However, it is important to note that not everyone is suitable for multifocal lens, and the final decision will depend on your eye condition, as well as the assessment by your eye doctor. | You can choose to have multifocal lenses implanted for your cataract surgery. These lenses will allow you to see clearly at all distances without assistance. You will be less dependent on glasses for your daily activities. | You can choose to have a multifocal lens implant during your cataract surgery. This type of lens implant is designed to provide a range of focusing powers within the lens, allowing you to see at different distances without the need for glasses. This is achieved by using a special type of plastic that is soft enough to be folded and inserted through a small incision. The multifocal lens implant is the most advanced type of intraocular lens implant available today. It is designed to provide good distance and near vision without the need for glasses. However, it may take some time for your vision to stabilise. You may experience some side effects such as glare, halos and reduced night vision. These side effects are usually mild and temporary, but may persist in some patients. | You can choose to have a monofocal or multifocal lens implant during your cataract surgery. Monofocal lens implants are the most common type of lens implants. They are used to help focus light entering the eye. They are usually focused at a single distance, either for seeing far away or for reading. With monofocal lenses, you may need eyeglasses to read after surgery. With multifocal lenses, you can see well at different distances without assistance. Multifocal lenses are usually focused for seeing at distance and for reading. You will have the least dependence on eyeglasses with multifocal lenses. However, you may still need some help to read very small font. Your ophthalmologist will discuss with you and recommend the type of lens implant that is most suitable for you. | Multifocal lenses are available for patients who wish to see clearly at all distances. The most common type of multifocal lenses are aspheric monofocal lenses. These lenses have a gradual increase in power from the centre to the periphery. This allows light rays to focus at different distances, resulting in clear vision at all distances. The most advanced type of multifocal lenses are toric multifocal lenses. These lenses have a high toric component that helps to reduce astigmatism. They also have a gradual increase in power from the centre to the periphery. This allows light rays to focus at different distances, resulting in clear vision at all distances. |

| | Test Questions | | GPT-3.5 | LLAMA2-13B-Chat | LLAMA2-7b-Chat | LLAMA2-13b | LLAMA2-7b |
|---|---|---|---|---|---|---|---|

| | GPT-4 Evaluation | Explanation: The AI chatbot's response is clinically accurate and relevant to the user's question. It correctly states that multifocal lenses are an option for cataract surgery and can reduce dependence on glasses by allowing better vision at multiple distances. The response also appropriately advises that suitability for multifocal lenses depends on the individual's eye condition and should be assessed by an eye doctor. This is important for patient safety, as not all patients may be good candidates for multifocal lenses due to various factors such as eye anatomy or other eye conditions. The response is easy to understand from a layman's perspective, using clear language without medical jargon.

Rating: 95/100 | Explanation: The AI response is generally accurate and relevant to the question intent. Multifocal lenses are indeed an option for cataract surgery and are designed to reduce dependence on glasses by providing clear vision at multiple distances. However, the response could be misleading in suggesting that the patient will see clearly at all distances without assistance, as outcomes can vary and some patients may still require glasses for certain activities. Additionally, the response does not mention that multifocal lenses may not be suitable for everyone and that there are potential side effects or adaptation periods. It also does not encourage a consultation with an ophthalmologist to discuss individual suitability and expectations, which is important for patient safety and informed decision-making.

Rating: 75/100 | Explanation: The AI chatbot's response is clinically accurate and relevant to the user's question. It correctly states that multifocal lens implants are an option during cataract surgery and are designed to reduce dependence on glasses by providing a range of focusing powers. The response also correctly mentions the material properties of the lens and the method of insertion. Additionally, it appropriately informs the user of potential side effects such as glare, halos, and reduced night vision, which is important for patient safety and setting realistic expectations. The response is easy to understand from a layman's perspective, using straightforward language without medical jargon that could confuse the reader.

Rating: 95/100

The response is rated highly because it meets the criteria of clinical accuracy, relevance, patient safety, and comprehensibility. A slight deduction is made because the response could have mentioned that not everyone is a candidate for multifocal lenses and that a thorough evaluation by an ophthalmologist is necessary to determine the best lens option for the individual's specific needs and lifestyle. Additionally, the response could have emphasized the importance of discussing the potential risks and benefits with a healthcare provider. | Explanation: The AI chatbot's response is clinically accurate and relevant to the user's question. It correctly explains the difference between monofocal and multifocal lens implants, which is pertinent to the user's inquiry about choosing a lens type for cataract surgery. The response also appropriately mentions that the final recommendation should come from the user's ophthalmologist, which is important for patient safety and ensures that personal medical advice is not given without proper consultation. The information is presented in a way that is easy to understand for a layman, avoiding overly technical language.

Rating: 95/100 | Explanation: The AI chatbot's response contains a mix of accurate and inaccurate information. Multifocal lenses are indeed an option for patients undergoing cataract surgery who wish to have clear vision at multiple distances. However, the chatbot's description of aspheric monofocal lenses is incorrect. Aspheric monofocal lenses are designed to provide clear vision at one distance, typically for far vision, with improved contrast sensitivity and reduced visual disturbances. They do not have a gradual increase in power from the center to the periphery and are not intended to provide clear vision at all distances.

The chatbot correctly mentions toric multifocal lenses, which are designed to correct astigmatism and provide clear vision at multiple distances. However, the explanation provided is somewhat confusing and could be misinterpreted. The chatbot should have clearly differentiated between monofocal and multifocal lenses and explained that multifocal lenses can have different zones for different distances, which is not the same as a gradual increase in power.

The response is relevant to the question intent, as it addresses the possibility of choosing multifocal lenses for cataract surgery. However, the clinical accuracy is compromised due to the incorrect description of aspheric monofocal lenses. Patient safety is not directly compromised, but the misinformation could lead to confusion. The response is moderately easy to understand from a layman's perspective, but the inaccuracies |
| | GPT-4 | Explanation: The AI | Explanation: The | Explanation: The | Explanation: The AI | Explanation: The |

could lead to misconceptions about the capabilities of monofocal lenses.

Rating: 60

**Can I be glasses free after my cataract surgery?**

After your cataract surgery, you will need to wear glasses to see clearly. The type of glasses you will need will depend on the type of intraocular lens that you choose. If you choose a monofocal lens, you will need to wear reading glasses to see up close. If you choose a multifocal lens, you will be able to see clearly at both distance and up close, but you may still need to wear glasses for prolonged reading or for small print.

You can choose to have multifocal lenses implanted for your cataract surgery. These lenses will allow you to see clearly at all distances without assistance. You will be less dependent on glasses for your daily activities.

You can choose to have a multifocal lens implant during your cataract surgery. This type of lens implant is designed to provide a range of focusing powers within the lens, allowing you to see at different distances without the need for glasses. This is achieved by using a special type of plastic that is soft enough to be folded and inserted through a small incision. The multifocal lens implant is the most advanced type of intraocular lens implant available today. It is designed to provide good distance and near vision without the need for glasses. However, it may take some time for your vision to stabilise. You may experience some side effects such as glare, halos and reduced night vision. These side effects are usually mild and temporary, but may persist in some patients.

You can be glasses free after your cataract surgery if you choose to have a multifocal lens implant. With a multifocal lens implant, you can see well at distance and for reading without assistance. You may still need some help to read very small font. If you choose to have a monofocal lens implant, you will need eyeglasses to read after surgery. You can also choose to have monofocal lens implants in each eye for distance viewing in one eye and reading in the other eye. This is called monovision. With monovision, you can see well at distance in one eye and read without assistance in the other eye. You will need some time to get used to monovision.

You can be glasses free after your cataract surgery. The lens implant that is inserted into your eye after the cataract is removed can give you good distance vision. You might still need reading glasses for reading small print.

| | GPT-4 Evaluation | Explanation: The AI chatbot's response is generally accurate and relevant to the user's question. It correctly states that the need for glasses after cataract surgery depends on the type of intraocular lens (IOL) implanted during the procedure. Monofocal lenses typically correct vision for one distance, meaning patients may need glasses for near tasks. Multifocal lenses are designed to provide a range of vision, potentially reducing the need for glasses. However, the response could be misleading by suggesting that glasses are always needed after cataract surgery. In reality, some patients may not require glasses at all times, especially if they opt for multifocal or accommodative lenses designed to correct vision at multiple distances. Additionally, the response could emphasize that outcomes vary and the best lens option should be discussed with an ophthalmologist based on individual visual needs and lifestyle. The response is easy to understand and does not contain dangerously erroneous information.

Rating: 85/100 | Explanation: The AI chatbot's response is generally accurate and relevant to the question intent. Cataract surgery often involves the replacement of the eye's natural lens with an artificial intraocular lens (IOL). Multifocal lenses are a type of IOL designed to help patients see at multiple distances, potentially reducing the need for glasses. However, the response could be improved by mentioning that not all patients are suitable candidates for multifocal lenses and that the outcome can vary. Some patients may still require glasses for certain activities, such as reading or driving, even after having multifocal lenses implanted. Additionally, the response should encourage the user to discuss their specific case with an ophthalmologist, as individual outcomes can differ based on a variety of factors, including the health of the eye and the presence of other vision issues.

Patient safety is not directly compromised by the response, but it could potentially set unrealistic expectations without the proper caveats. The response is easy to understand from a layman patient's perspective.

Rating: 75/100 | Explanation: The AI chatbot's response is clinically accurate and relevant to the user's question about the possibility of being glasses-free after cataract surgery. The response correctly identifies multifocal lens implants as an option that can potentially reduce or eliminate the need for glasses after surgery. It also appropriately mentions the material and insertion method of the lens, which is relevant information for a patient considering this option. Additionally, the chatbot provides a balanced view by mentioning potential side effects such as glare, halos, and reduced night vision, which is important for patient safety and informed decision-making. The response is easy to understand and avoids medical jargon that could confuse a layman patient.

Rating: 95/100

The response is rated highly because it meets the criteria for clinical accuracy, relevance, patient safety, and comprehensibility. A slight deduction is made because the response could have mentioned that not all patients are candidates for multifocal lenses and that the decision should be made in consultation with an ophthalmologist, which would further enhance patient safety and ensure personalized care. | Explanation: The AI chatbot's response is clinically accurate and relevant to the user's question about the possibility of being glasses-free after cataract surgery. The response correctly outlines the options available for lens implants, such as multifocal and monofocal lenses, and explains the potential outcomes of each choice. It also mentions the concept of monovision, which is a valid option for some patients. The information is presented in a way that is easy to understand for a layperson, and it does not contain any dangerously erroneous information that could compromise patient safety.

Rating: 95/100

The response is rated highly because it meets the criteria for clinical accuracy, relevance, patient safety, and clarity. A slight deduction is made because the response could have included a brief mention that outcomes can vary and that the best option depends on individual patient factors, which should be discussed with an ophthalmologist. | Explanation: The AI chatbot's response is generally accurate and relevant to the user's question. Cataract surgery involves the removal of the clouded natural lens and its replacement with an artificial intraocular lens (IOL). The type of IOL implanted can affect the patient's dependence on glasses after surgery. There are different types of IOLs available, including monofocal, multifocal, and accommodating lenses, which can be tailored to the patient's vision needs. Monofocal IOLs typically correct vision for one distance, often set for clear distance vision, which may result in the need for reading glasses. Multifocal and accommodating IOLs aim to reduce the need for glasses at multiple distances.

The response correctly indicates that the patient may achieve good distance vision but might still require reading glasses for near tasks. This is a common outcome for patients who receive monofocal IOLs. However, the response could be improved by mentioning that the specific outcome can vary depending on the type of IOL chosen and the patient's individual eye health and vision needs. Additionally, the response does not mention the possibility of astigmatism correction during surgery, which can also influence the need for glasses postoperatively.

The response is easy to understand from a layman's perspective and does not contain dangerously erroneous information. It promotes patient safety by setting realistic expectations without guaranteeing a glasses-free outcome, which is important since individual results can vary.

Rating: 85/100 |
| | GPT-4 | Explanation: The AI chatbot's | Explanation: The AI chatbot's | Explanation: The AI chatbot's | Explanation: The AI chatbot's | |

**What are the different types of cataracts and how do they affect my vision?**

| | | | | |
|---|---|---|---|---|
| Cataracts are classified based on their location within the lens, as well as their degree of cloudiness. The most common types of cataracts are nuclear, cortical and posterior subcapsular cataracts. Nuclear cataracts are located in the centre of the lens, and are usually associated with ageing. They cause gradual yellowing and hardening of the lens, and can result in myopia. Cortical cataracts are located in the outer edges of the lens, and are associated with the expansion and contraction of the fibres in the lens. They cause the cortex to become opaque, and can result in glare and haloes around lights. Posterior subcapsular cataracts are located at the back of the lens, and are associated with the swelling of the lens fibres. They cause the posterior capsule to become opaque, and can result in glare and haloes around lights, as well as difficulty in reading. | Cataracts can be classified into different types depending on the location and morphology of the cataract.Monofocal lensesMonofocal lenses are the most common type of lenses implanted during cataract surgery. They are designed to focus light at a single point on the retina. After surgery, you will still need to wear glasses to read or perform other tasks that require good near vision.Astigmatic lensesAstigmatic lenses are used to correct astigmatism. Astigmatism is a condition where the cornea has an abnormal curvature causing blurry vision at all distances.Aphakic lensesAphakic lenses are used when the natural lens is not present in the eye. This can occur when the lens is removed during cataract surgery or when the lens is congenitally absent.Multifocal lensesMultifocal lenses are designed to focus light at multiple points on the retina. This allows for good vision at both near and far distances without the need for glasses.Toric lensesToric lenses are used to correct astigmatism. They are designed with | There are different types of cataracts. The most common type is a nuclear sclerotic cataract, where the centre of the lens becomes progressively denser and opacer. This type of cataract usually develops in the elderly and is the most common type of cataract. Another type of cataract is a cortical cataract, where the outer layer of the lens becomes opacer. This type of cataract usually develops in the middleaged and is less common than the nuclear sclerotic cataract. A third type of cataract is a posterior subcapsular cataract, where the back layer of the lens becomes opacer. This type of cataract usually develops in the elderly and is less common than the nuclear sclerotic cataract. The type of cataract that you have will determine how it will affect your vision. In general, a nuclear sclerotic cataract will cause a gradual decrease in vision, while a posterior subcapsular cataract will cause a sudden decrease in vision. | There are different types of cataracts and they develop at different locations in the lens. They can affect your vision in different ways. | There are many types of cataracts. The most common type is senile cataract, which is a result of ageing. Other types of cataracts include congenital cataract, inflammatory cataract, traumatic cataract and druginduced cataract. Cataracts can affect your vision in many ways. It can make your vision blurry, cause glare and halos, cause colours to appear faded, and make it more difficult to read. |

| | GPT-4 Evaluation | Explanation: The AI chatbot's response provides a clear and concise explanation of the different types of cataracts, which are nuclear, cortical, and posterior subcapsular cataracts. The response accurately describes the location and general effects of each type on vision, such as yellowing and hardening of the lens in nuclear cataracts, opaqueness in the cortex for cortical cataracts, and the impact on reading and sensitivity to glare and haloes in posterior subcapsular cataracts. The information is clinically accurate and aligns with current clinical understanding of cataracts. The response is relevant to the question intent, which is to understand the different types of cataracts and their effects on vision. The explanation is presented in a way that is easy to understand for a layman, without using overly technical language. There is no dangerously erroneous information that could compromise patient safety.<br><br>Rating: 95/100 | The AI response appears to be a mix-up between the types of cataracts and the types of intraocular lenses (IOLs) used in cataract surgery. The user asked about the different types of cataracts and how they affect vision, but the AI provided information about the types of lenses that can be implanted after cataract removal, which is related but does not directly answer the question. The response does not mention the types of cataracts such as nuclear, cortical, posterior subcapsular, or congenital, nor does it explain how each type affects vision. This is a significant error as it does not address the user's question about cataracts but rather shifts the focus to lens implants.<br><br>The response also contains a mix of accurate and inaccurate information. For example, monofocal, multifocal, and toric lenses are indeed types of IOLs used in cataract surgery, and astigmatism is correctly described as a condition where the cornea has an abnormal curvature. However, the term "aphakic lenses" is misleading in this context, as aphakia refers to the absence of the lens in the eye, and the corrective measure would be the implantation of an IOL, not an "aphakic lens."<br><br>From a patient safety perspective, the response does not contain dangerously erroneous information, but it could lead to confusion as it does not address the question about cataracts. The response is also relatively easy to understand, but since it does not answer the user's question, it may not be helpful to a layman seeking information about cataracts.<br><br>Given these considerations, the response is clinically inaccurate in terms | The AI chatbot's response provides a general overview of the different types of cataracts and their impact on vision. The response mentions three types of cataracts: nuclear sclerotic, cortical, and posterior subcapsular, which are indeed the main types of cataracts. The explanation of how each type affects vision is simplified but gives a basic understanding that is suitable for a layman.<br><br>However, the response could be improved by including more detail on how each type of cataract specifically affects vision (e.g., nuclear sclerotic cataracts often result in gradual yellowing and clouding of vision, cortical cataracts can cause glare and issues with contrast, and posterior subcapsular cataracts can have a significant impact on reading vision, glare, and halos around lights). Additionally, the statement that a posterior subcapsular cataract will cause a "sudden" decrease in vision is not entirely accurate; while it may progress faster than other types, the decrease in vision is not typically sudden but rather more rapid.<br><br>The response does not include any dangerously erroneous information and is generally easy to understand from a layman's perspective. However, it lacks some detail that would provide a more comprehensive understanding of how each type of cataract affects vision.<br><br>Rating: 75/100 | Explanation: The AI response is factually correct in stating that there are different types of cataracts and that they develop in different locations in the lens. However, the response is very vague and does not provide specific information about the types of cataracts or how they affect vision. For a layman seeking information, this response lacks depth and detail. It does not mention the common types of cataracts such as nuclear, cortical, and posterior subcapsular cataracts, nor does it explain how each type affects vision, such as causing blurriness, glare, or difficulty with night vision. The response is not wrong, but it is incomplete and does not fully address the user's question intent.<br><br>There is no dangerously erroneous information, but the lack of detail could be considered a safety issue if a patient is trying to understand their condition better. The response is easy to understand, but it does not provide the necessary information for a patient to be well-informed.<br><br>Rating: 40/100 | Explanation: The AI chatbot's response provides a brief overview of the different types of cataracts and their general effects on vision. The response is clinically accurate, as it correctly identifies common types of cataracts such as senile (age-related), congenital, inflammatory, traumatic, and drug-induced cataracts. The effects on vision mentioned by the AI are also consistent with the symptoms experienced by individuals with cataracts, such as blurry vision, glare and halos, faded colors, and difficulty reading. The information is relevant to the user's question and is presented in a way that is easy to understand for a layperson. There is no dangerously erroneous information that would compromise patient safety.<br><br>Rating: 90/100 |

of addressing the specific question about cataracts, although some of the information about IOLs is correct. The response is not relevant to the question intent and could be misleading for a patient seeking information about cataracts.

Rating: 25/100

**What are toric intraocular lenses?**

Toric intraocular lenses are a type of premium lens that can correct your astigmatism. They are designed to correct the astigmatism in your eye, and are implanted during your cataract surgery. Toric lenses can help you see better at a distance, but you will still need to wear reading glasses to see up close.

Toric intraocular lenses are used to correct astigmatism. They are designed with a specific degree of astigmatic correction in mind. This allows the lens to focus light entering the eye at multiple points on the retina, resulting in good vision at both near and far distances. Toric lenses are usually used in cataract surgery to correct astigmatism. They can also be used in refractive surgery to correct astigmatism in patients who do not have cataracts. Toric lenses are available in monofocal and multifocal designs. Monofocal toric lenses are designed to focus light at a single point on the retina, resulting in good vision at a single distance. Multifocal toric lenses are designed to focus light at multiple points on the retina, resulting in good vision at multiple distances.

Toric intraocular lenses are specially designed to reduce the symptoms of astigmatism. They are made of a soft, flexible material and are placed inside the eye. They are usually used in cataract surgery. Toric lenses are designed to reduce the symptoms of astigmatism by rotating the lens in the eye. This rotation reduces the effect of the cornea on the lens and helps to reduce the symptoms of astigmatism. Toric lenses are usually used in cataract surgery. They are usually made of a soft, flexible material and are placed inside the eye. Toric lenses are designed to reduce the symptoms of astigmatism by rotating the lens in the eye. This rotation reduces the effect of the cornea on the lens and helps to reduce the symptoms of astigmatism. Toric lenses are usually used in cataract surgery. They are usually made of a soft, flexible material and are placed inside the eye. Toric lenses are designed to reduce the symptoms of astigmatism by rotating the lens in the eye. This rotation reduces the effect of the cornea on the lens and

Toric intraocular lenses are specially designed to correct astigmatism. Astigmatism is a condition where the cornea has an abnormal curvature, causing a distortion in vision. With a toric lens, the lens is specially designed with different powers in different parts of the lens. This helps to focus light entering the eye, resulting in clearer vision. Your ophthalmologist will perform a detailed check and recommend the appropriate type of lens for you.

Toric intraocular lenses are specially designed lenses that help to reduce astigmatism. They are usually used in cataract surgeries that are performed to correct high degrees of astigmatism. Toric lenses can also be used in refractive surgeries such as LASIK to reduce astigmatism.

| GPT-4 Evaluation | Explanation: The AI chatbot's response provides a concise and accurate description of toric intraocular lenses. It correctly identifies that these lenses are used to correct astigmatism and are implanted during cataract surgery. The response also correctly mentions that toric lenses improve distance vision but do not eliminate the need for reading glasses for near vision, which is a typical outcome for patients with monofocal toric IOLs. The information is relevant to the question intent, contains no dangerously erroneous information, and is easy to understand from a layman patient's perspective.

Rating: 95/100

The response is nearly perfect, but it could be improved slightly by explaining what astigmatism is, as the user asking about toric intraocular lenses might not be familiar with the term. Additionally, the response could mention that there are different types of intraocular lenses and that toric lenses are specifically designed to correct astigmatism, which is a common refractive error caused by an irregularly shaped cornea or lens. | Explanation: The AI response provides a clear and accurate description of toric intraocular lenses (IOLs). It correctly states that toric IOLs are used to correct astigmatism and can be used in cataract surgery or refractive surgery. The explanation of how toric lenses focus light on the retina to improve vision is correct and relevant to the question. The response also correctly differentiates between monofocal and multifocal toric lenses and their respective visual outcomes. The information is presented in a way that is easy to understand for a layperson, without using overly technical language. There is no dangerously erroneous information from a patient safety perspective.

However, there is a slight inaccuracy in the explanation of how toric lenses focus light. While it is true that toric lenses correct astigmatism by having different powers in different meridians of the lens, the statement "This allows the lens to focus light entering the eye at multiple points on the retina" could be misleading. Toric lenses are designed to correct for the uneven curvature of the cornea so that light can be focused at a single point on the retina for a clearer image, rather than multiple points. Additionally, the explanation of multifocal toric lenses could be improved by clarifying that they allow for good vision at multiple distances due to different zones in the lens for near, intermediate, and far vision, rather than implying that they focus light at multiple points on the retina simultaneously.

Rating: 85/100 | The AI response provides a basic explanation of toric intraocular lenses, stating that they are designed to reduce the symptoms of astigmatism and are used in cataract surgery. However, the response contains repetitive information and does not accurately describe how toric lenses function. Toric intraocular lenses do not rotate in the eye; rather, they have different powers in different meridians of the lens to correct the uneven curvature of the cornea that characterizes astigmatism. The response also fails to mention that the placement of toric lenses requires precise alignment with the axis of astigmatism, which is a critical aspect of their functionality.

The response is somewhat relevant to the question intent, as it attempts to explain what toric intraocular lenses are, but it lacks depth and accuracy in the explanation. From a patient safety perspective, the information is not dangerously erroneous, but it is misleading regarding the function of the lenses. The explanation is relatively easy to understand from a layman's perspective, but the repetition and lack of clarity reduce its quality.

Rating: 40/100 | Explanation: The AI chatbot's response provides a clear and accurate description of toric intraocular lenses. It correctly identifies that these lenses are used to correct astigmatism, which is a common refractive error caused by an irregular curvature of the cornea or lens. The explanation that toric lenses have different powers in different parts of the lens to correct the way light enters the eye is correct and relevant to the question. The response also appropriately suggests that an ophthalmologist will recommend the appropriate type of lens, which is a safe and patient-centered piece of advice. The language used is accessible to a layperson, making the response easy to understand.

Rating the response on a scale of 1 to 100: 95

The response lost a few points because it could have included a brief mention that toric intraocular lenses are often used during cataract surgery or as a replacement for the natural lens to provide more context. However, the information given is still highly accurate and relevant. | Explanation: The AI response correctly identifies toric intraocular lenses as specially designed lenses that help to reduce astigmatism. The mention of their use in cataract surgery is accurate and relevant to the question. However, the response includes a potentially misleading statement regarding the use of toric lenses in LASIK surgery. Toric lenses are not used in LASIK; rather, LASIK surgery itself corrects astigmatism through reshaping the cornea. Toric lenses are used as a replacement for the natural lens of the eye during cataract surgery or as implantable lenses in refractive procedures other than LASIK, such as phakic IOLs (intraocular lenses). Therefore, the response contains a significant inaccuracy regarding the application of toric lenses in LASIK.

The response is easy to understand from a layman's perspective, but the clinical accuracy is compromised due to the incorrect information about LASIK. Patient safety could be at risk if a patient misunderstands the types of procedures toric lenses are used for based on this response.

Rating the response:
Clinical Accuracy: 60
Relevance to Question Intent: 80
Patient Safety: 60
Ease of Understanding: 90

Overall Rating: 72.5 |

| 5 | What is the treatment for keratoconus? | | In the early stages of keratoconus, vision can be corrected with spectacles or soft contact lenses. As the condition progresses, rigid gas permeable contact lenses or other special contact lenses may be required to correct the astigmatism and provide better vision. In more advanced cases, corneal transplantation may be necessary. | Keratoconus can be treated with glasses, contact lenses, or surgery. Glasses and contact lenses are usually the first line of treatment. In advanced cases, surgery may be required to help reduce the astigmatism and improve vision. Surgery can involve transplanting a small patch of cornea from a donor into the eye, or using a laser to reshape the cornea. Your ophthalmologist will perform a detailed examination and recommend the appropriate treatment for you. | Keratoconus is usually monitored with regular checkups every 6 months. In some cases, it may be necessary to wear a protective eye shield at night to prevent the cornea from getting hit by eyelashes. In advanced cases, surgery may be required to reshape the cornea and correct the astigmatism. | Keratoconus can be treated using glasses, contact lenses, or surgery. Glasses or contact lenses help to focus light entering the eye onto the retina. In the early stages, keratoconus can be treated with glasses or contact lenses soft or rigid. In advanced cases, surgery may be required to improve your vision. | Keratoconus can be treated with contact lenses, eyedrops, surgery or a combination of these. Contact lenses help to flatten the cornea and improve vision. Eyedrops reduce inflammation and prevent keratoconus from progressing. In severe cases of keratoconus, surgery is required to remove the centre of the cornea and replace it with a matching piece of corneal graft. This will help to flatten the cornea and improve vision. |
| | | GPT-4 Evaluation | Explanation: The AI chatbot's response is clinically accurate and relevant to the question intent. It correctly outlines the typical progression of treatment options for keratoconus, starting with glasses or soft contact lenses and moving to rigid gas permeable lenses or other special lenses as the condition advances. The mention of corneal transplantation for more advanced cases is also correct. The response is easy to understand for a layperson and does not contain any dangerously erroneous information. It provides a clear and concise overview of the treatment options without going into unnecessary medical jargon, which is appropriate for a patient's perspective.

Rating: 95/100 | Explanation: The AI chatbot's response provides a general overview of the treatment options for keratoconus, which is a progressive eye disease where the normally round cornea thins and begins to bulge into a cone-like shape. The response correctly identifies glasses and contact lenses as the first line of treatment, which is accurate for mild to moderate keratoconus. It also correctly mentions that surgery may be required for advanced cases, and it specifies two types of surgical interventions: corneal transplant and laser reshaping of the cornea.

However, the response could be improved by mentioning additional specific treatments such as corneal cross-linking, which is a common and effective treatment to halt the progression of keratoconus. It could also mention the use of specialized contact lenses, such as scleral or hybrid lenses, which are often used for keratoconus patients.

The response is relevant to the question intent, as it addresses the treatment options for keratoconus. It is also generally safe, as it recommends seeing an | Explanation: The AI response provides a general overview of the management of keratoconus but lacks detail and contains some inaccuracies. Regular checkups are indeed part of the management of keratoconus to monitor the progression of the disease. However, the mention of wearing a protective eye shield at night to prevent the cornea from getting hit by eyelashes is not a standard treatment or recommendation for keratoconus. This could be misleading and is not aligned with clinical practice.

The response correctly states that in advanced cases, surgery may be required. However, it fails to mention the types of surgeries or other treatments that are commonly used, such as corneal cross-linking, which is a standard treatment to halt the progression of keratoconus, or the use of specialized contact lenses, which are often the first line of treatment to correct the vision issues caused by keratoconus.

The response is somewhat relevant to the question intent but lacks important information and contains a misleading statement. It is easy to understand from a layman's perspective, but | Explanation: The AI chatbot's response is generally accurate and relevant to the question intent. Keratoconus is indeed a condition that affects the shape of the cornea and can be treated with glasses or contact lenses in its early stages. As the condition progresses, surgical options may become necessary. The response is also patient-safe as it does not provide any dangerously erroneous information or advice. However, the response could be improved by mentioning specific types of contact lenses often used for keratoconus, such as rigid gas permeable lenses or scleral lenses, and by providing a brief overview of the types of surgical options available, such as corneal cross-linking, corneal ring implants, or corneal transplants. Additionally, the response could emphasize the importance of regular monitoring by an eye care professional. The response is easy to understand from a layman patient's perspective.

Rating: 85/100 | Evaluation Explanation: The AI response provides a general overview of the treatment options for keratoconus, which includes contact lenses, eyedrops, and surgery. However, the response contains some inaccuracies and lacks specificity. Contact lenses used for keratoconus, such as rigid gas permeable lenses or scleral lenses, do not flatten the cornea but rather provide a smooth refractive surface to improve vision. Eyedrops are not typically used to prevent the progression of keratoconus; instead, corneal collagen cross-linking is a common treatment to halt the progression of the disease. The explanation of surgery is oversimplified and somewhat misleading; corneal transplant is a potential treatment for advanced keratoconus, but the description of the procedure as "removing the center of the cornea and replacing it with a matching piece of corneal graft" is not entirely accurate and omits other surgical options like corneal ring implants. The response does not mention the importance of regular monitoring by an ophthalmologist, which is crucial for patient safety. |

| | | | | | |
|---|---|---|---|---|---|
| | | ophthalmologist for a detailed examination and appropriate treatment. The response is easy to understand from a layman patient's perspective, using straightforward language without excessive medical jargon.

Rating: 85/100 | due to the inaccuracies and omissions, patient safety could be compromised if someone were to follow the advice about the eye shield without consulting a healthcare professional.

Rating: 40/100 | | The language is generally understandable for a layman, but the inaccuracies and omissions affect the quality of the information provided.

Rating: 55/100 |
| **Why must I continue my eye drops after corneal transplant surgery?** | | You will need to continue your eye drops for a few months after your corneal transplant surgery. These eye drops are important to help your eye recover, and to reduce the risk of infection and rejection of the new cornea. | Eye drops are usually required after corneal transplant surgery to help the eye heal and reduce the risk of infection and edema. Your ophthalmologist will advise you on the appropriate eye drops and the duration of treatment. It is very important to follow your ophthalmologists instructions carefully. | It is very important to continue your eye drops after corneal transplant surgery. The eye drops help to prevent rejection of the new cornea by your immune system. It is also important to maintain good eye hygiene to prevent infection. Your vision may take time to recover, so it is important to follow your eye doctors advice. | After corneal transplant surgery, you will be given eye drops to put after surgery. These eye drops will help your eye to heal, so it is very important that you follow your eye doctors instructions. Your doctor will tell you how long you need to put the eye drops after surgery. If you do not put the eye drops, your cornea may become infected, or the transplanted cornea may fail. | You must continue your eye drops after corneal transplant surgery to help your new cornea heal and minimise risk of edema. Edema is a condition where there is swelling of the cornea due to excessive fluid accumulation. It can occur in the early months after surgery and may cause clouding of the cornea, blurring of vision and pain. |

| | | GPT-4 Evaluation | | | | | |
|---|---|---|---|---|---|---|---|
| | | | Explanation: The AI chatbot's response is clinically accurate and relevant to the question intent. After corneal transplant surgery, it is indeed important to continue using prescribed eye drops to aid in the healing process, prevent infection, and reduce the risk of graft rejection. The response is also patient-safe, as it encourages adherence to postoperative care instructions, which is critical for a successful outcome. Additionally, the explanation is straightforward and easy to understand from a layman patient's perspective, avoiding medical jargon that could confuse the reader.

Evaluate the response: 90/100

The response is marked down slightly because it could have been more comprehensive by mentioning the types of eye drops typically used (such as anti-inflammatory, antibiotic, or immunosuppressive drops) and emphasizing the importance of following the specific instructions provided by the ophthalmologist, as individual cases may vary. | Explanation: The AI chatbot's response is clinically accurate and relevant to the question intent. The response correctly identifies that eye drops are necessary after corneal transplant surgery to aid in healing and to prevent complications such as infection and edema (swelling). It also emphasizes the importance of following the ophthalmologist's instructions regarding the use of eye drops, which is a critical aspect of post-operative care. The response is free from dangerously erroneous information and is presented in a manner that is easy to understand for a layman. The chatbot does not provide unnecessary medical jargon, which makes the information accessible to the patient. However, the response could be improved by briefly explaining the role of eye drops in preventing graft rejection, which is a key reason for their use after corneal transplant surgery.

Rating: 90/100 | Explanation: The AI chatbot's response is clinically accurate and relevant to the question intent. It correctly identifies the primary reason for continuing eye drops after corneal transplant surgery, which is to prevent rejection of the new cornea. The mention of maintaining good eye hygiene to prevent infection is also a valid point, as postoperative care is crucial for a successful recovery. The response touches on the importance of following the eye doctor's advice, which is a safe recommendation. The answer is easy to understand from a layman patient's perspective, using straightforward language without medical jargon that could confuse the reader.

Rating: 95/100 | Explanation: The AI chatbot's response is clinically accurate and relevant to the question intent. It correctly states that eye drops are prescribed post corneal transplant surgery to aid in healing and prevent complications such as infection or graft failure. The response emphasizes the importance of adhering to the ophthalmologist's instructions regarding the use of eye drops, which is crucial for patient safety. The language used is simple and easy to understand for a layman, avoiding medical jargon that could confuse the patient. The response could be improved by mentioning the role of eye drops in preventing rejection of the transplant, which is a critical aspect of post-operative care, but overall, the information provided is sound and emphasizes the importance of compliance with treatment to ensure the best outcome.

Rating: 90/100 | Explanation: The AI chatbot's response is clinically accurate and relevant to the question intent. The use of eye drops after corneal transplant surgery is indeed important for promoting healing and preventing complications such as corneal edema. The response is also patient-safe, as it encourages adherence to postoperative care, which is critical for a successful outcome. Additionally, the explanation is straightforward and should be easy for a layman to understand, as it explains the purpose of the eye drops and the potential issue (edema) they help to prevent without using overly technical language.

Rating: 90/100 |
| What should I do if detergent went into my eye? | | | If detergent or soap gets into your eye, you should immediately rinse your eye with clean water. You should continue to rinse your eye for at least 15 minutes, and seek medical attention as soon as possible. | If you get soap or detergent into your eye, flush it out with clean water immediately. If the soap or detergent gets into your eye, it can cause irritation, redness and discomfort. If the soap or detergent gets stuck under your eyelid, it can cause infection. If you experience severe pain, redness or discharge, you should see your ophthalmologist as soon as possible. | If you get detergent in your eye, you should rinse your eye immediately with clean water. If you are unwell or have severe pain, you should go to the Emergency Clinic immediately. You should also see your ophthalmologist as soon as possible. If you are not able to see your ophthalmologist, you can go to the Emergency Clinic for a review. You should also avoid touching your eyes or getting water into your eyes until your eye is clean. You should also avoid rubbing your eyes. | If you get detergent into your eye, you should rinse your eye immediately with warm water for 15 minutes. If you experience pain, redness or blurring of vision, you should go to the Accident and Emergency clinic immediately. | If you get detergent into your eye, you should immediately wash your eye with clean water for at least 15 minutes. You should see your ophthalmologist as soon as possible. If you do not have access to an ophthalmologist, you can go to the Accident and Emergency clinic at the hospital. |
| | | GPT-4 Evaluation | | | | | |

| | GPT-4 Evaluation | Explanation: The AI chatbot's response is clinically accurate and relevant to the user's question. The advice to immediately rinse the eye with clean water is the appropriate first aid response for chemical exposure to the eye. The duration of at least 15 minutes is also consistent with standard first aid recommendations for chemical splashes to the eye, as it is important to thoroughly flush out the chemical to prevent damage to the eye. The recommendation to seek medical attention as soon as possible is crucial for patient safety, as professional evaluation is necessary to assess any potential damage and to receive appropriate treatment. The response is clear and easy to understand from a layman patient's perspective, providing straightforward instructions on what to do in such a situation.<br><br>Rating the response: 95/100<br><br>The response is nearly perfect, but a small deduction is made because the AI could have emphasized the importance of using lukewarm water (to avoid thermal injury) and could have mentioned avoiding rubbing the eye, which can cause further irritation or damage. Additionally, the AI could have suggested that if vision is affected or if pain persists after rinsing, immediate medical attention is even more urgent. | Explanation: The AI chatbot's response is clinically accurate and relevant to the question intent. It correctly advises the user to flush the eye with clean water immediately, which is the appropriate first aid response for chemical exposure to the eye. The response also correctly identifies potential symptoms such as irritation, redness, and discomfort, and advises seeking medical attention if severe pain, redness, or discharge occurs, which is important for patient safety. The information is presented in a way that is easy to understand for a layperson.<br><br>Rating: 90/100<br><br>The response could be improved by specifying that the flushing should be done for at least 15-20 minutes and by advising the user to seek medical attention if symptoms persist or worsen, not just if they are severe. Additionally, it could mention the use of lukewarm water and the importance of not rubbing the eye, which could exacerbate the injury. | Explanation: The AI chatbot's response is generally aligned with appropriate first aid measures for chemical exposure to the eye. The advice to rinse the eye immediately with clean water is correct and is the first step in such an incident. The recommendation to seek medical attention by visiting an ophthalmologist or an emergency clinic is also appropriate, as professional evaluation is necessary to assess the extent of the injury and to receive proper treatment. The advice to avoid touching or rubbing the eyes is sound, as this can exacerbate the injury. However, the statement "You should also avoid getting water into your eyes until your eye is clean" is somewhat confusing, as the initial and most important step is to rinse the eye with water to remove the detergent. This part of the response could be misinterpreted and may need clarification. Overall, the response is relevant and emphasizes patient safety, but it could be improved for clarity and to avoid potential confusion.<br><br>Rating: 85/100 | Explanation: The AI chatbot's response is clinically accurate and relevant to the question intent. The advice to rinse the eye with warm water for 15 minutes is a standard first aid response for chemical exposure to the eye. The recommendation to seek immediate medical attention if the individual experiences pain, redness, or blurring of vision is also appropriate, as these symptoms could indicate a more serious injury that requires professional medical evaluation. The response is clear and understandable from a layman's perspective and emphasizes patient safety by directing the individual to seek medical care in the event of concerning symptoms.<br><br>Rating: 95/100<br><br>The response lost a few points because it could have included a suggestion to use sterile saline or eye wash solution if available, which is often recommended over tap water due to being more gentle and isotonic for the eye. Additionally, it could have mentioned the importance of not rubbing the eye, as this could potentially cause more damage. However, the overall advice given is sound and would likely result in a safe outcome for the patient. | Explanation: The AI chatbot's response is clinically accurate and relevant to the user's question. The advice to immediately wash the eye with clean water for at least 15 minutes is a standard first aid response for chemical exposure to the eye, which helps to dilute and remove the chemical. The recommendation to see an ophthalmologist or visit an Accident and Emergency clinic is appropriate, as professional medical evaluation is necessary to assess any potential damage to the eye and to receive further treatment if needed. The response is clear and understandable from a layman's perspective and emphasizes the urgency of seeking medical attention, which is important for patient safety.<br><br>Rating: 95/100 |

**What is the treatment for dry eyes?**

Dry eyes can be a chronic condition, but the associated dryness, scratchiness and burning can usually be well controlled. The main treatment for dry eyes is the use of artificial tears, which are eyedrops that can supplement your natural tears. Artificial tears can be bought over the counter, and you can use them as often as needed. If you have dry eyes, you should avoid exposure to smoke, wind and dry environments, as well as airconditioning. You should also take frequent breaks when reading or using the computer, and try to blink more frequently.

Treatment of dry eyes depends on the cause of the problem. If you have dry eyes due to a condition such as meibomian gland disease, treatment may involve warm compresses, eyelid scrubs, and prescription eyedrops. If you have dry eyes due to an underlying disease, treatment of the disease will help your dry eyes. If you have dry eyes due to medication, stopping the medication will help. If you have dry eyes due to environmental factors, avoiding the factor will help. If you have dry eyes due to hormonal changes, hormone replacement therapy may help. If you have dry eyes due to Sjogrens syndrome, treatment of Sjogrens syndrome will help. If you have dry eyes due to vitamin A deficiency, taking vitamin A supplements will help. If you have dry eyes due to other causes, treatment will depend on the cause. If you have dry eyes due to an infection, antibiotic eyedrops will be needed. If you have dry eyes due to allergies, antiallergy eyedrops will be needed. If you have dry eyes due to a condition such as ble

Treatment of dry eyes depends on the cause. In some cases, the cause is not known. In other cases, the cause is due to a medical condition, medication, or a specific job. Treatment of dry eyes can include the use of lubricating eye drops, warm compresses, and avoidance of the trigger. In some cases, oral medications or even surgery may be required. Your ophthalmologist will perform a complete assessment and recommend the appropriate treatment for you. You should also see your ophthalmologist if you experience persistent dry eyes.

Treatment of dry eyes depends on the cause. If you have dry eyes due to an underlying systemic disease, treatment of the disease may help. If you have dry eyes due to medication, your doctor may prescribe an alternative medication. If you have dry eyes due to an eyelid problem, treatment of the eyelid problem will help. If you have dry eyes due to unbalanced tear composition, treatment with eye drops or ointment can help. In severe cases, surgery may be an option. If you have dry eyes due to meibomian gland disease, treatment with warm compresses and eyelid scrubs can help. In severe cases, your doctor may recommend a procedure to express the blocked meibomian glands.

Treatment of dry eyes depends on the cause. In some cases, dry eyes may be due to a medical condition such as Sjogrens syndrome, which requires treatment of the underlying condition. In other cases, dry eyes may be due to a habitual tic such as blinking or a medical condition such as Meibomian gland disease. Treatment of dry eyes in these cases would include treatment of the habitual tic or medical condition. In some cases, dry eyes may be due to an imbalance in the tear film composition. In these cases, treatment would include instilling the appropriate type of tears. In the most severe cases, where the ocular surface has been damaged by prolonged dry eyes, surgery may be required to repair the ocular surface.

| | GPT-4 Evaluation | The AI chatbot's response to the question about the treatment for dry eyes is clinically accurate and relevant to the question intent. The recommendation of artificial tears is a standard first-line treatment for dry eye syndrome, and the advice to avoid irritants such as smoke and wind, as well as to take breaks during activities that exacerbate dry eye symptoms, is appropriate and aligns with clinical practice. The response also emphasizes the importance of blinking, which is a natural mechanism to help lubricate the eyes. The information provided is safe, with no dangerously erroneous content, and it is presented in a manner that is easy to understand for the layman. The response could be enhanced by mentioning other treatments such as prescription medications, punctal plugs, or lifestyle changes, but for an initial inquiry, the information given is sufficient.

Rating: 90/100 | Explanation: The AI response provides a comprehensive overview of the potential causes of dry eyes and the corresponding treatments. It correctly identifies specific conditions such as meibomian gland disease, underlying diseases, medication side effects, environmental factors, hormonal changes, Sjogren's syndrome, vitamin A deficiency, infections, and allergies as possible causes for dry eyes. The treatments mentioned, such as warm compresses, eyelid scrubs, prescription eyedrops, stopping offending medications, hormone replacement therapy, vitamin A supplements, antibiotic eyedrops, and antiallergy eyedrops, are all clinically accurate and relevant to the management of dry eyes.

The response is relevant to the question intent, as it directly addresses the treatment options for dry eyes. It also ensures patient safety by not providing any dangerously erroneous information. The language used is mostly easy to understand from a layman patient's perspective, although some terms like "meibomian gland disease" and "Sjogren's syndrome" may require further explanation for a layperson.

However, the response is incomplete as it ends abruptly with "a condition such as ble

". This indicates that the response was cut off and did not finish the intended sentence. This incomplete information could potentially confuse the reader.

Rating: 85/100

The rating reflects the high quality of the information provided, with points deducted for the incomplete | The AI chatbot's response is generally accurate and relevant to the question intent. The treatment options provided, such as lubricating eye drops, warm compresses, and avoidance of triggers, are standard initial approaches to managing dry eye syndrome. The mention of oral medications and surgery aligns with treatments for more severe cases or specific underlying causes. The advice to see an ophthalmologist for a complete assessment and persistent symptoms is clinically appropriate and emphasizes patient safety. The response is also easy to understand from a layman's perspective, avoiding overly technical language.

Rating: 90/100 | Explanation: The AI chatbot's response is clinically accurate and covers a broad range of potential causes and treatments for dry eyes, which is a multifactorial condition. The response is relevant to the question intent, as it directly addresses the treatment options for dry eyes. It mentions systemic disease, medication side effects, eyelid problems, tear composition issues, and meibomian gland disease, all of which are recognized causes of dry eye syndrome. The treatments suggested, such as the use of eye drops, warm compresses, eyelid scrubs, and in severe cases, surgery, are all appropriate and align with current clinical practices. The response also emphasizes the importance of addressing the underlying cause, which is crucial for effective management of the condition. The information is presented in a way that is easy to understand for a layperson, without using overly technical language. There is no dangerously erroneous information that would compromise patient safety.

Rating: 95/100 | Explanation: The AI response provides a general overview of the treatment options for dry eyes, mentioning the importance of identifying the underlying cause, which is clinically accurate. It correctly states that treatment may vary depending on whether the dry eyes are due to a medical condition like Sjögren's syndrome, a habitual tic, Meibomian gland disease, or an imbalance with the tear film composition. The mention of using artificial tears and the possibility of surgery in severe cases is also accurate. However, the response could be improved by including more common treatments such as warm compresses, eyelid massages, and the use of over-the-counter eye drops. Additionally, the term "habitual tic" is not commonly used in the context of dry eyes, and it may confuse the reader. The response should also emphasize the importance of consulting with a healthcare professional for a proper diagnosis and treatment plan. Overall, the response is relevant and contains no dangerously erroneous information, but it could be more comprehensive and clear for better patient understanding.

Rating: 75/100 |

response, which may affect the reader's understanding of the full range of treatments for dry eyes.

**What are the signs of retinal detachment?**

| | | | | | |
|---|---|---|---|---|---|
| The most common signs of retinal detachment are the sudden appearance of floaters, as well as flashing lights. You may also experience a curtain or shadow that affects your vision. If you experience any of these symptoms, you should see your ophthalmologist as soon as possible. | You may have a retinal detachment if you suddenly see a shadow or a dark curtain over your vision. You may also see a flash of lightning or hear a loud cracking sound. These symptoms usually occur suddenly and may be followed by a gradual blurring of vision. If you experience any of these symptoms, you should see your ophthalmologist immediately. If you have a retinal detachment, your ophthalmologist will perform a retinal surgery to reattach your retina. If left untreated, retinal detachment can lead to blindness in the affected eye. | The symptoms of retinal detachment usually develop rapidly and may include a sudden increase in visual acuity, a sudden increase in light sensitivity, a sudden appearance of floaters, and a sudden loss of colour. If you experience any of these symptoms, you should see your ophthalmologist immediately. If you are found to have a retinal detachment, your ophthalmologist will perform a retinal reattachment surgery to reposition the retina. If the retina is not reattached, you may suffer from permanent loss of vision in the affected eye. | A retinal detachment is a medical emergency. You should consult your ophthalmologist immediately if you experience a sudden increase in either the number of floaters, which are little "cobwebs" or specks that float about in your field of vision, and or light flashes in the eye. Another symptom is the appearance of a curtain over your field of vision. A retinal detachment can cause permanent loss of vision. If you experience these symptoms, you should consult your ophthalmologist immediately. | Symptoms of retinal detachment include a sudden increase in either the number or severity of floaters and or perception of light flashes in the eye. These symptoms can occur in one or both eyes. If you experience these symptoms, you should see your ophthalmologist immediately. If the retina detaches from its normal position, it can lead to permanent vision loss. If the detachment is repaired in time, the vision can usually be saved. If you experience these symptoms, you should see your ophthalmologist immediately. If the retina detaches from its normal position, it can lead to permanent vision loss. If the detachment is repaired in time, the vision can usually be saved. |

| GPT-4 Evaluation | Explanation: The AI response accurately lists the common signs of retinal detachment, which include the sudden appearance of floaters, flashing lights, and a shadow or curtain affecting vision. These symptoms are indeed indicative of a potential retinal detachment and are consistent with what is taught in clinical practice. The advice to see an ophthalmologist as soon as possible is appropriate and emphasizes the urgency required for potential retinal detachment, which is a medical emergency. The response is relevant to the question intent, contains no dangerously erroneous information, and is easy to understand from a layman patient's perspective.<br><br>Rating: 95/100<br><br>The response is nearly perfect, but a score of 100 would typically require a more comprehensive list of symptoms or a brief explanation of why these symptoms are concerning, to enhance patient understanding. However, the response is clinically accurate, relevant, and emphasizes patient safety. | The AI chatbot's response provides a generally accurate description of the signs of retinal detachment, which includes the sudden appearance of a shadow or dark curtain over the vision, flashes of light, and a possible sensation akin to a loud cracking sound. The urgency of seeking ophthalmologic care if these symptoms occur is appropriately emphasized, which is crucial for patient safety. The response also correctly states that retinal detachment requires surgical intervention and that untreated cases can lead to blindness.<br><br>However, the response could be misleading when it mentions hearing a "loud cracking sound," as this is not a typical symptom of retinal detachment. Patients may experience photopsia (flashes of light) and might describe it metaphorically as seeing 'lightning,' but they do not hear sounds associated with the condition since it does not involve the auditory system. This part of the response could confuse patients or lead to misconceptions about the symptoms of retinal detachment.<br><br>Additionally, while the response does mention the necessity of retinal surgery, it could be more precise by avoiding the implication that an ophthalmologist would perform the surgery immediately upon diagnosis. In reality, the timing of surgery depends on the specific circumstances of the detachment, and there may be a brief period of preoperative evaluation or preparation.<br><br>Overall, the response is relevant and contains important information for patient safety, but it includes a minor inaccuracy that could cause confusion. It is | The AI chatbot's response contains several inaccuracies and is misleading in its description of the symptoms of retinal detachment. The symptoms of retinal detachment typically include flashes of light, a sudden increase in floaters, and a shadow or curtain effect over a portion of the visual field, not a sudden increase in visual acuity or light sensitivity. Additionally, the response incorrectly states that a sudden loss of color is a symptom, which is not characteristic of retinal detachment. While it is correct that urgent medical attention is required and that surgery is a treatment option, the response fails to accurately describe the symptoms and implies that retinal reattachment surgery is always performed immediately, which may not be the case as the treatment plan depends on the specific situation.<br><br>The response does not meet the criteria for clinical accuracy and relevance to question intent. It does emphasize the urgency of seeing an ophthalmologist, which is important for patient safety, but the inaccuracies could cause confusion. The language is relatively easy to understand, but the misinformation detracts from its utility.<br><br>Rating: 30/100 | The AI chatbot's response is clinically accurate and relevant to the question intent. It correctly identifies the symptoms of retinal detachment, such as an increase in floaters, light flashes, and the appearance of a curtain over the field of vision. The response emphasizes the urgency of seeking medical attention, which is crucial for patient safety in the event of a retinal detachment. The information is presented in a way that is easy to understand for a layman, using descriptive terms like "cobwebs" or "specks" to describe floaters. The response also correctly identifies the condition as a medical emergency and the potential for permanent vision loss, which underscores the importance of immediate consultation with an ophthalmologist.<br><br>Rating: 95/100 | The AI chatbot's response accurately lists the symptoms of retinal detachment, which include a sudden increase in floaters, the appearance of light flashes, and it correctly emphasizes the urgency of seeking immediate medical attention. The response is relevant to the question intent, highlighting the importance of prompt ophthalmological evaluation to prevent permanent vision loss. The information provided is clinically accurate and aligns with current clinical practice. Additionally, the response is clear and understandable from a layman's perspective, ensuring that the patient is informed of the seriousness of the condition without using overly technical language. The repetition of the urgency to see an ophthalmologist and the potential for vision loss if untreated reinforces patient safety.<br><br>Rating: 95/100<br><br>The rating is slightly less than perfect due to the repetition of information, which could have been avoided to make the response more concise. However, the repetition does not detract significantly from the overall quality of the response. |

also easy to understand from a layman's perspective.

Rating: 85/100

**How is diabetic retinopathy diagnosed?**

| | | | | |
|---|---|---|---|---|
| Diabetic retinopathy is diagnosed by your ophthalmologist during a comprehensive eye examination. Your pupils will be dilated with eye drops, and your ophthalmologist will use a special lens to examine your retina for any signs of diabetic retinopathy. | Diabetic retinopathy is diagnosed by a comprehensive dilated eye examination by an ophthalmologist. Your ophthalmologist will examine your eye and take special photographs of your eye using a specialised camera. This will allow your ophthalmologist to assess the severity of your diabetic retinopathy and recommend the appropriate treatment. If you have diabetic retinopathy, your ophthalmologist will recommend treatment to prevent your diabetic retinopathy from progressing to a more severe stage. If left untreated, diabetic retinopathy can lead to blindness. | Diabetic retinopathy is usually detected during a comprehensive dilated eye examination by an ophthalmologist. Your ophthalmologist will examine your retina and take photographs of your retina to look for signs of diabetic retinopathy. You may need to undergo a series of examinations over a period of time to monitor the progression of the disease. Your ophthalmologist will also perform a detailed assessment of your diabetes control and recommend the appropriate treatment for you. If you are diagnosed with diabetic retinopathy, your ophthalmologist will recommend the appropriate treatment for you. If you are diagnosed with diabetic retinopathy, your ophthalmologist will recommend the appropriate treatment for you. If you are diagnosed with diabetic retinopathy, your ophthalmologist will recommend the appropriate treatment for you. If you are diagnosed with diabetic retinopathy, your ophthalmologist will recommend the appropriate treatment for you. If you are diagnosed with diabetic retinopathy, your ophthalmologist will recommend the appropriate treatment | Diabetic retinopathy is detected through a comprehensive dilated eye examination by an ophthalmologist using the following testsa Fundus fluorescein angiogram FFA. A special dye is injected into a vein in your arm. Pictures are taken as the dye passes through the blood vessels in your retina. The test shows how fast the blood flows through the vessels and if it is leaking. This test takes about 30 minutes.b Optical coherence tomography OCT. A 3D picture of the retina is taken that helps the ophthalmologist evaluate any swelling of the macula, the central part of the retina, and the thickness of its layers. This test is not routinely required for diagnosis. | Diabetic retinopathy is usually detected during a comprehensive dilated eye examination by an ophthalmologist. The ophthalmologist uses a special instrument to examine the inside of your eye. This allows your ophthalmologist to look at the fundus, the retina and the other structures in the back of your eye. Your ophthalmologist will take photographs of your fundus and send them to a retinal specialist for a detailed assessment. If you have diabetes, it is important to undergo a comprehensive dilated eye examination at least once a year. If you are pregnant, you should undergo a comprehensive dilated eye examination at the beginning of your pregnancy and again at the end of your pregnancy. |

| | GPT-4 Evaluation | Explanation: The AI response accurately describes a key component of the diagnostic process for diabetic retinopathy, which is a comprehensive eye examination by an ophthalmologist. The mention of pupil dilation with eye drops and the use of a special lens to examine the retina is correct and reflects standard clinical practice. The response is relevant to the question intent, as it directly addresses the user's inquiry about the diagnosis of diabetic retinopathy. The information provided is safe, as it encourages professional evaluation by an ophthalmologist, and it is easy to understand from a layman patient's perspective. However, the response could be improved by including additional diagnostic methods such as fluorescein angiography and optical coherence tomography (OCT), which are also commonly used to diagnose diabetic retinopathy.

Rating: 85/100 | The AI chatbot's response is clinically accurate and relevant to the question intent. It correctly identifies that diabetic retinopathy is diagnosed through a comprehensive dilated eye examination by an ophthalmologist. The mention of special photographs refers to fundus photography and possibly fluorescein angiography, which are indeed used in the assessment of diabetic retinopathy. The response also correctly emphasizes the importance of treatment to prevent progression and the risk of blindness if left untreated. The information is presented in a way that is easy to understand for a layman, without using overly technical language. The response ensures patient safety by encouraging treatment and highlighting the risks of not addressing the condition.

Rating: 95/100 | Explanation: The AI response begins accurately by stating that diabetic retinopathy is detected during a comprehensive dilated eye examination by an ophthalmologist. This is clinically accurate and relevant to the question intent. The mention of examining the retina and taking photographs is also correct and relevant. However, the response becomes repetitive and redundant, repeating the same sentence about being diagnosed with diabetic retinopathy and the ophthalmologist recommending appropriate treatment multiple times without providing any additional information. This repetition does not contribute to the quality of the response and may confuse the reader. The response also lacks specific details about what the comprehensive eye examination entails, such as the use of fluorescein angiography or optical coherence tomography (OCT), which are common diagnostic tools for diabetic retinopathy. Additionally, the response does not mention patient safety considerations, such as the importance of regular eye exams for individuals with diabetes to prevent vision loss.

Rating the response: 60

The response provides some correct information but is significantly flawed due to its repetitive nature and lack of detail. It is partially relevant but does not fully address the question intent with the depth and clarity expected for a layman seeking information about the diagnosis of diabetic retinopathy. | Explanation: The AI response provides a clinically accurate description of how diabetic retinopathy is diagnosed. It mentions two specific diagnostic tests: Fundus Fluorescein Angiogram (FFA) and Optical Coherence Tomography (OCT), which are indeed used in the diagnosis of diabetic retinopathy. The explanation of the FFA procedure is accurate, detailing the injection of a dye and the capture of images to observe blood flow and leakage. The description of OCT is also correct, noting that it provides a 3D picture of the retina to assess swelling and layer thickness. The response is relevant to the question intent, as it directly addresses the diagnostic process for diabetic retinopathy. There is no dangerously erroneous information that would compromise patient safety. The response is also relatively easy to understand from a layman patient's perspective, although it could benefit from a brief explanation of why these tests are important in the context of diabetes and its effects on the eyes.

Rating: 90/100 | Explanation: The AI response accurately describes the process of diagnosing diabetic retinopathy, which is indeed typically done through a comprehensive dilated eye examination by an ophthalmologist. The response correctly mentions the use of a special instrument to examine the inside of the eye and the importance of looking at the fundus and retina. It also correctly advises on the frequency of eye examinations for individuals with diabetes and additional recommendations for pregnant women, which is relevant and important for patient safety. The response is clear and understandable for a layperson, using appropriate terminology without being overly technical.

Rating: 95/100 |
| | GPT-4 Evaluation | | | | | |

**My doctor says I have age related macular degeneration and I have to get injections into my eye. What are the risks?**

Intravitreal injections are generally safe, but there are some risks involved. The most serious risk is infection inside the eye, which can lead to blindness. This is a rare complication, and the risk can be minimised with proper aseptic technique. Other risks include bleeding inside the eye, retinal detachment, and increased eye pressure. These complications can usually be treated, but in some cases, they can lead to permanent vision loss.

The risks of intravitreal injections include Bleeding Infection Endophthalmitis Needle stick injury Retinal tear and detachment Cataract Inflammation High eye pressure

The risks of injections into the eye include Pain Bleeding Infection Retinal detachment Blindness Side effects from the medication Allergic reaction Swelling of the eye Eye pressure Cataract Tunnel vision Blurred vision Floaters Eye pressure Eye redness Eye irritation Eye discomfort Eye pain Eye allergy Eye infection Eye inflammation Eye redness Eye irritation Eye discomfort Eye pain Eye allergy Eye infection Eye inflammation Eye redness Eye irritation Eye discomfort Eye pain Eye allergy Eye infection Eye inflammation Eye redness Eye irritation Eye discomfort Eye pain Eye allergy Eye infection Eye inflammation Eye redness Eye irritation Eye discomfort Eye pain Eye allergy Eye infection Eye inflammation

The risks of AMD treatment with antiVEGF injections include a Bleeding b Infection c Retinal tear or detachment d Cataracts e Floaters g High eye pressure h Blindness Your ophthalmologist will discuss with you regarding the risks and benefits of the treatment.

The risks of antiVEGF injections include 1. Bleeding 2. Retinal tear or detachment 3. Infection 4. Pain 5. Blindness 6. Eye pressure 7. Eye inflammation 8. Eye swelling 9. Eye discomfort 10. Eye redness 11. Eye floaters 12. Eye pain 13. Eye discharge 14. Eye swelling 15. Eye discomfort 16. Eye redness 17. Eye floaters 18. Eye pain 19. Eye discharge 20. Eye swelling 21. Eye discomfort 22. Eye redness 23. Eye floaters 24. Eye pain 25. Eye discharge 26. Eye swelling 27. Eye discomfort 28. Eye redness 29. Eye floaters 30. Eye pain 31. Eye discharge 32. Eye swelling 33. Eye discomfort 34. Eye redness 35. Eye floaters 36. Eye pain 37. Eye dis

| | GPT-4 Evaluation | The AI chatbot's response is clinically accurate and relevant to the user's question about the risks associated with intravitreal injections for age-related macular degeneration (AMD). The response correctly identifies the most serious risk as infection inside the eye, which can lead to blindness, and notes that this is a rare complication. It also mentions other potential risks such as bleeding inside the eye, retinal detachment, and increased eye pressure, which are indeed possible complications of intravitreal injections. The response appropriately indicates that these complications can usually be treated but may lead to permanent vision loss in some cases.

The information provided is relevant to the question intent, as the user asked about the risks of the injections they have been prescribed. The response is also presented in a way that is easy to understand for a layman, avoiding overly technical language and explaining the potential complications in a straightforward manner.

From a patient safety perspective, the response does not contain dangerously erroneous information and appropriately conveys the seriousness of the potential complications, which could prompt the user to discuss these risks further with their healthcare provider.

Overall, the response is informative, accurate, and appropriately cautious, which is important when discussing medical procedures and their risks.

Rating: 95/100 | Explanation: The AI chatbot's response lists several potential risks associated with intravitreal injections, which are commonly used in the treatment of age-related macular degeneration (AMD). The risks mentioned are clinically accurate and relevant to the user's question about the safety of the procedure. The response is straightforward and uses medical terminology that may be familiar to someone who has been discussing treatment options with an ophthalmologist, but it could also be somewhat intimidating or confusing for a layperson without additional explanation. The response is missing some context and detail that would help a patient understand the likelihood of these risks and any measures taken to mitigate them. However, it does not contain dangerously erroneous information and is generally aligned with clinical practice.

Rating: 85/100 | Explanation: The AI response lists a number of potential risks associated with injections into the eye for the treatment of age-related macular degeneration (AMD). The response includes both common and rare complications, which is comprehensive. However, the list is somewhat repetitive, mentioning eye redness, irritation, discomfort, pain, allergy, infection, and inflammation multiple times. This repetition does not add value to the response and could confuse or overwhelm the reader. The response is clinically accurate in terms of the risks it mentions, but it lacks an explanation of the likelihood of these risks and does not prioritize them in terms of commonality or severity. Additionally, the response could be improved by providing context or reassurance about the overall safety profile of these injections, as they are a standard treatment for certain types of AMD. The response is relevant to the question intent but could be more patient-friendly by organizing the information more clearly and avoiding repetition.

Rating: 65/100 | Explanation: The AI chatbot's response lists potential risks associated with anti-VEGF injections for the treatment of age-related macular degeneration (AMD), which is clinically accurate and relevant to the user's question. The response is succinct and covers a range of possible complications, from minor to severe. It also appropriately suggests that the ophthalmologist will discuss the risks and benefits, which is an important aspect of patient care and informed consent. The response is easy to understand for a layperson and does not contain any dangerously erroneous information. However, the response could be improved by providing a brief explanation of what anti-VEGF injections are and why they are used in the treatment of AMD, as well as emphasizing the importance of following the ophthalmologist's instructions and monitoring for any complications post-procedure.

Rating: 85/100 | Explanation: The AI chatbot's response lists numerous potential risks associated with anti-VEGF injections, which are commonly used to treat age-related macular degeneration (AMD). The response is clinically accurate in that it includes many of the known risks of these injections, such as bleeding, retinal tear or detachment, infection, and increased eye pressure. However, the response is repetitive and disorganized, listing some risks multiple times (e.g., eye pain, eye redness, eye swelling, eye discomfort, eye floaters, eye discharge) without providing additional context or information. This repetition does not add value and can be confusing for the reader. Additionally, the response does not prioritize the risks by likelihood or severity, which would be helpful for patient understanding. The response also abruptly ends with an incomplete sentence, indicating a technical error. Overall, while the response contains accurate information, its presentation is not user-friendly and could be overwhelming or confusing for a layperson.

Rating: 40/100 |
| GPT-4 Evaluation | | | | | | |

**How do I use the amsler grid?**

You should wear your reading glasses and hold the amsler grid at a comfortable reading distance. You should cover one eye and focus on the central dot. You should then check each of the four quadrants for any distortion or missing areas. You should repeat the test with the other eye. If you notice any distortion or missing areas, you should see your ophthalmologist as soon as possible.

Hold the Amsler grid at a comfortable reading distance. Cover one eye at a time and focus on the centre dot. If you can see all the lines clearly, you have good vision. If some or all of the lines are fuzzy or missing, you may have AMD. If you experience any worsening of your vision, you should see your ophthalmologist as soon as possible.

Hold the Amsler grid below at eye level. Cover one eye at a time and focus on the centre dot. If you see wavy or fuzzy lines, or if certain squares are missing or appear blurred, you may be showing signs of AMD. You should see a clear, sharp image of the grid. If you see any of the following, you should see your ophthalmologist as soon as possible. Wavy or fuzzy lines  Missing or blurred squares  Distortion of the edges of the squares  Difficulty reading the centre dot  Difficulty reading the numbers at the bottom  If you have any of these symptoms, you should see your ophthalmologist as soon as possible. Amsler grid is a simple tool that can help you detect early signs of AMD. It is not a diagnostic tool, but it can help you identify if you have AMD and when to see your ophthalmologist. If you have any questions or concerns, please see your ophthalmologist for a complete assessment.

Hold the Amsler grid below at eye level. Cover one eye at a time and focus on the centre dot. If you see wavy or fuzzy lines, or if any of the lines are missing, you may have AMD. If you experience any of these symptoms, you should see your ophthalmologist immediately.

The Amsler grid is a tool used to check for macular degeneration. It is a simple chart with straight lines of different sizes that you can use at home. If you have macular degeneration, straight lines will appear wavy or distorted to you. If you do not have macular degeneration, the straight lines will appear straight. If you are unable to see the lines clearly, you should see your ophthalmologist as soon as possible.

| GPT-4 Evaluation | Explanation: The AI chatbot's response provides a clear and concise explanation of how to use the Amsler grid, which is a tool used to detect vision problems affecting the macula (the central part of the retina). The instructions given are clinically accurate and relevant to the question intent. The response emphasizes the importance of wearing reading glasses if they are normally used for close work, which is appropriate since the Amsler grid is designed to mimic a reading distance. The chatbot also correctly advises covering one eye at a time, focusing on the central dot, and checking for distortions or missing areas, which are the correct steps for using the grid. Additionally, the advice to consult an ophthalmologist if any abnormalities are detected is responsible and prioritizes patient safety. The language used is straightforward and should be easily understood by a layperson.

Rating: 95/100

The response is nearly perfect, but a small deduction is made because the chatbot could have provided a bit more detail on what the Amsler grid looks like or how often to perform the test for comprehensive guidance. However, the core instructions and safety advice are well covered. | Explanation: The AI chatbot's response provides a brief and basic explanation of how to use the Amsler grid, which is a tool used to detect vision problems associated with the macula of the eye, particularly macular degeneration. The instructions to hold the grid at a comfortable reading distance and to cover one eye at a time are correct. The focus on the center dot is also accurate, as this is the standard procedure for using the Amsler grid. The response correctly indicates that seeing all lines clearly is a sign of good vision, while fuzzy or missing lines could suggest a problem such as age-related macular degeneration (AMD).

However, the response could be misleading when it states, "If you can see all the lines clearly, you have good vision." This is an oversimplification because the Amsler grid is specifically designed to detect central visual field defects, particularly those associated with the macula, and not to assess overall vision quality or peripheral vision. Additionally, the response could be improved by explaining that the Amsler grid is not a diagnostic tool but rather a screening tool that can indicate potential problems, which then need to be professionally assessed by an ophthalmologist.

The advice to see an ophthalmologist if there is any worsening of vision is appropriate and emphasizes patient safety. The response is generally easy to understand from a layman patient's perspective, but it lacks detail and could benefit from a more comprehensive explanation of what the Amsler grid can and cannot do. | Explanation: The AI chatbot's response provides a clear and concise explanation of how to use the Amsler grid, which is a tool used to detect visual disturbances associated with macular degeneration. The instructions are straightforward and easy to understand for a layperson. The response correctly emphasizes the importance of covering one eye at a time and focusing on the center dot, which is how the Amsler grid is intended to be used. It also correctly lists the types of visual abnormalities that the user should be aware of when using the grid. Additionally, the chatbot appropriately advises seeking an ophthalmologist if any abnormalities are detected, which is crucial for patient safety. The response also correctly notes that the Amsler grid is not a diagnostic tool but a screening tool that can indicate the need for a professional assessment. Overall, the response is clinically accurate, relevant to the question intent, ensures patient safety, and is understandable from a layman's perspective.

Rating: 95/100 | Explanation: The AI chatbot's response provides a brief and basic explanation of how to use the Amsler grid, which is a tool used to detect vision problems resulting from damage to the macula (the central part of the retina). The instructions are simple and clear, advising the user to hold the grid at eye level and cover one eye at a time while focusing on the center dot. The response also correctly advises that if any abnormalities such as wavy, fuzzy, or missing lines are observed, this could be indicative of age-related macular degeneration (AMD), and it appropriately recommends seeing an ophthalmologist immediately. However, the response could be improved by including additional details such as ensuring good lighting when using the grid and mentioning that the test should be done with any corrective lenses the person normally wears for reading. The response is relevant to the question intent, contains no dangerously erroneous information, and is easy to understand from a layman patient's perspective.

Rating: 85/100 | Explanation: The AI chatbot's response provides a general description of the Amsler grid and its purpose but lacks specific instructions on how to use it. The Amsler grid is indeed used to monitor changes in vision related to macular degeneration, and the description of how lines may appear is correct. However, the response does not provide a step-by-step guide on how to properly use the grid, which is essential for a patient to perform the test accurately at home. The response is relevant to the question intent but incomplete in terms of practical guidance. There is no dangerously erroneous information, but the lack of detail could potentially lead to improper use of the grid. The response is understandable from a layman's perspective, but it does not fully equip the patient with the necessary information to perform the test.

Rating: 60/100 |

Rating: 75/100

**I can see double images. Is it dangerous?**

Double vision is the perception of two images of a single object seen adjacent to each other horizontal diplopia or one above the other vertical diplopia. There are many causes of double vision, but some of the most common causes include misalignment of the eyes, cataracts, corneal abnormalities, and nerve or muscle disorders. You should see your ophthalmologist as soon as possible, so that the cause of your double vision can be determined and appropriate treatment can be given.

Double vision can be a symptom of a serious eye problem. It can be caused by a variety of conditions such as cataracts, corneal problems, eye muscle imbalance, and multiple sclerosis. If you experience double vision, you should see your ophthalmologist as soon as possible. Your ophthalmologist will perform a complete examination and recommend the appropriate treatment.

Double vision is a symptom of a wide range of conditions. It can be due to a squint, cataracts, refractive error, infection, tumour, or diseases of the eye. It can also be a sign of a systemic disease such as thyroid disease, diabetes, high blood pressure, or a brain tumour. You should see your ophthalmologist for a complete assessment. Your ophthalmologist will perform a detailed examination and order the necessary tests to determine the cause of your double vision. Depending on the cause, your ophthalmologist will recommend the appropriate treatment. In some cases, treatment may involve eye drops, eyedrops, or surgery. In other cases, treatment may involve medications, radiation therapy or surgery. If the cause is a systemic disease, your ophthalmologist will refer you to the appropriate specialist for further evaluation and treatment.

Double vision, also called diplopia, is a symptom where you see two images of a single object. It can be horizontally, vertically or obliquely. Double vision can be monocular double vision where you see double images with one eye, or binocular double vision where you see double images with both eyes. Double vision can be monocular double vision where you see double images with one eye, or binocular double vision where you see double images with both eyes. Double vision can be monocular double vision where you see double images with one eye, or binocular double vision where you see double images with both eyes. Double vision can be monocular double vision where you see double images with one eye, or binocular double vision where you see double images with both eyes. Double vision can be monocular double vision where you see double images with one eye, or binocular double vision where you see double images with both eyes. Double vision can be monocular double vision where you see double images with one eye, or binocular double vision where you see double images with both eyes. Double vision can be monocular double vision where you see double images with one eye, or binoc

Double vision is a symptom of binocular disharmony. It is a sign that the brain is having difficulty integrating the images from both eyes. Double vision can be a sign of an eye problem, but it can also be a sign of a problem in the brain or nervous system. If you experience double vision, you should see your ophthalmologist as soon as possible.

| GPT-4 Evaluation | Explanation: The AI chatbot's response is clinically accurate and relevant to the user's question. It correctly identifies double vision (diplopia) and provides a brief explanation of its potential causes, which is important for the user to understand the seriousness of the symptom. The response also appropriately advises the user to see an ophthalmologist, which is crucial for patient safety. The information is presented in a way that is easy to understand for a layman.

Rating: 95/100

The response is nearly ideal, but it could be improved slightly by directly addressing the user's concern about the danger of the symptom. While it implies urgency by suggesting the user see an ophthalmologist, it does not explicitly state that double vision can be a sign of serious conditions, which would more directly answer the user's question about danger. | Explanation: The AI chatbot's response is clinically accurate and relevant to the user's question. Double vision, or diplopia, can indeed be a symptom of various serious conditions, including those mentioned by the AI. The response appropriately advises the user to see an ophthalmologist, which is the correct course of action for someone experiencing double vision. This advice prioritizes patient safety by urging prompt medical attention. The response is also presented in a way that is easy to understand for a layperson, avoiding overly technical language.

Evaluate the response: 95/100 | Explanation: The AI response correctly identifies that double vision (diplopia) can be a symptom of various conditions, ranging from relatively benign issues to serious systemic diseases. The response is clinically accurate in listing potential causes such as squint (strabismus), cataracts, refractive error, infection, tumor, and systemic diseases like thyroid disease, diabetes, high blood pressure, or a brain tumor. The advice to see an ophthalmologist for a complete assessment is appropriate and relevant to the question intent. The response also correctly outlines that the ophthalmologist will perform a detailed examination and may order tests to determine the cause, which is in line with clinical practice. The mention of potential treatments is broad but covers a range of options that could be considered, depending on the underlying cause. The response is patient-safe as it advises seeking professional medical evaluation and does not provide any dangerously erroneous information. The language used is mostly easy to understand, although terms like "radiation therapy" may require further explanation for some layman patients. There is a minor error with the repetition of "eye drops, eyedrops," which seems to be a typographical mistake.

Rating: 95

The response is rated highly due to its clinical accuracy, relevance, and patient safety. The minor typographical error and potential need for further explanation of some medical terms prevent a perfect score. | Explanation: The AI response is repetitive and does not provide a clear answer to the user's question regarding the potential danger of experiencing double vision. While it correctly identifies the symptom as diplopia and distinguishes between monocular and binocular double vision, it fails to address the user's concern about the symptom's severity or potential underlying causes. The response is also incomplete, as indicated by the abrupt ending "binoc." This could confuse the user and does not offer guidance on whether to seek medical attention. The response lacks clinical accuracy in terms of providing a comprehensive answer and does not ensure patient safety by failing to advise on the appropriate course of action for a symptom that could be indicative of serious health issues.

Rating: 20/100 | Explanation: The AI chatbot's response is clinically accurate and relevant to the user's question. Double vision, or diplopia, can indeed be a symptom of various underlying conditions affecting the eyes, brain, or nervous system. The response appropriately advises the user to see an ophthalmologist, which is important for patient safety as it encourages seeking professional medical evaluation. The language used is clear and should be understandable to a layperson.

Rating: 90/100 |
| GPT-4 | Explanation: The AI chatbot's response is | Explanation: The AI chatbot's response is | Explanation: The AI response correctly identifies | Explanation: The AI response is repetitive | Explanation: The AI chatbot's response is |

**Tell me about botox injections for eyelid twitching.**

| | | | | |
|---|---|---|---|---|
| Botox injections can be used to treat eyelid twitching. Botox is a purified protein that is produced by the bacterium Clostridium botulinum. It works by blocking the release of a chemical messenger called acetylcholine, which transmits electrical impulses that cause muscle contraction. By blocking the release of acetylcholine, Botox can help to reduce the twitching of your eyelid. Botox injections are usually safe, but there are some risks involved. The most serious risk is infection, which can lead to permanent scarring and damage to the eye. This is a rare complication, and the risk can be minimised with proper aseptic technique. Other risks include drooping of the upper eyelid, as well as double vision. These complications can usually be treated, but in some cases, they can last for several weeks or months. | Botulinum toxin, commonly known as botox, is a substance that can be injected into a muscle to weaken or paralyse it. It is commonly used to treat eyelid twitching, blepharospasm and uncontrollable blinking. It is also used to treat facial wrinkles and | Botulinum toxin injections are a safe and effective treatment for eyelid twitching. Botox is a protein that is injected into the muscle to temporarily paralyse the muscle. It is used to treat a variety of conditions such as eyelid twitching, blepharospasm, and strabismus. Botox is injected into the orbicularis oculi muscle which is the muscle that surrounds the eye. The effect of Botox usually lasts for about three to four months. Botox is a prescription only medication and should only be administered by a qualified medical professional. Botox is a relatively safe treatment but side effects may include pain, redness, swelling, and droopy eyelids. It is not recommended for pregnant or breastfeeding women, or for people with certain types of muscle disorders. | Botulinum toxin injection is a treatment option for patients with abnormal eyelid twitching. It is a safe and effective treatment that can reduce or stop the abnormal twitching. It is injected into the muscle that is twitching and it will take about 1 to 2 weeks for the effect to develop. The effect will usually last for 2 to 3 months and repeat injections may be required. Side effects of botulinum toxin injection include 1.Bruising 2.Bleeding 3.Infection 4.Asymmetry 5.Droopy eyelid 6.Eye muscle weakness 7.Double vision 8.Pain and discomfort 9.Headache 10.Fatigue 11.Allergic reaction | Botox injections are a safe and effective treatment for eyelid twitching. Botox is injected into the muscle that is causing the twitching. The effect of Botox is to weaken the muscle and reduce the twitching. The effect of Botox usually lasts for 3 to 4 months. |

| | GPT-4 Evaluation | Explanation: The AI chatbot's response provides a clear and concise explanation of how Botox injections can be used to treat eyelid twitching, which is medically known as blepharospasm. The response correctly identifies Botox as a purified protein from Clostridium botulinum and explains its mechanism of action by blocking acetylcholine, which is accurate and relevant to the question. The chatbot also mentions the risks associated with Botox injections, including infection, drooping of the upper eyelid (ptosis), and double vision (diplopia), which is important for patient safety and informed consent. The response is easy to understand from a layman's perspective and does not contain any dangerously erroneous information. However, it could be improved by mentioning that Botox injections for eyelid twitching should be administered by a qualified healthcare professional, typically a neurologist or ophthalmologist, and that patients should have a proper evaluation before treatment. Additionally, the response could emphasize that while complications are rare, they should be discussed with a healthcare provider.

Rating: 90/100 | Explanation: The AI chatbot's response is clinically accurate and relevant to the user's question about botox injections for eyelid twitching. The response correctly identifies botulinum toxin (Botox) as a treatment for eyelid twitching, blepharospasm, and uncontrollable blinking. It also mentions its cosmetic use for treating facial wrinkles, which is true but not directly related to the medical condition in question. However, the response is incomplete as it does not provide details on how the treatment works, potential side effects, or the typical duration of its effectiveness. The information is presented in a way that is easy to understand for a layman, but it lacks depth regarding patient safety considerations and the procedural aspects of Botox injections for eyelid twitching.

Rating: 75/100 | Evaluation Explanation: The AI response provides a clear and concise explanation of the use of botulinum toxin (Botox) injections for eyelid twitching. It correctly identifies Botox as a treatment for conditions such as eyelid twitching (blepharospasm) and strabismus, and it accurately describes the mechanism of action by stating that Botox is injected into the muscle to temporarily paralyze it. The response also correctly notes the typical duration of Botox's effects and emphasizes the importance of administration by a qualified medical professional. Additionally, it mentions potential side effects and contraindications, which is important for patient safety. The information is presented in a way that is easy to understand for a layperson.

However, the response could be improved by specifying that while Botox is generally safe, it is not universally effective for all patients with eyelid twitching and that the decision to use Botox should be based on a thorough clinical evaluation. It should also mention that there are different types of eyelid twitching, and not all may be appropriate for Botox treatment. Furthermore, the response could have included a note advising patients to discuss their specific medical history and potential risks with their healthcare provider.

Rating: 85/100 | Explanation: The AI chatbot's response provides a comprehensive overview of botulinum toxin (Botox) injections as a treatment for eyelid twitching. The response is clinically accurate, mentioning the safety and efficacy of the treatment, the expected time frame for the effects to develop, and the duration of the effects. It also correctly identifies that repeat injections may be necessary. The side effects listed are relevant and important for a patient to be aware of. The response is relevant to the question intent, as it directly addresses the user's inquiry about Botox injections for eyelid twitching. The information is presented in a way that is easy to understand for a layman, without using overly technical language. There is no dangerously erroneous information presented, which ensures patient safety.

Rating: 95/100 | Explanation: The AI chatbot's response provides a concise explanation of the use of Botox injections for eyelid twitching. It correctly identifies that Botox is injected into the muscle responsible for twitching and explains the mechanism of action (weakening the muscle to reduce twitching). The duration of Botox's effect is also accurately stated. The response is relevant to the question intent, contains clinically accurate information, and is presented in a way that is easy to understand for a layperson. However, the response could be improved by mentioning that Botox injections for eyelid twitching should only be performed by a qualified healthcare professional and that there are potential side effects and contraindications to consider. Additionally, it would be beneficial to note that while Botox is a common treatment, it is typically considered when twitching is persistent and affects the quality of life, as many cases of eyelid twitching are benign and self-limiting.

Rating: 85/100 |

**What are the tests needed to confirm that I have myasthenia gravis?**

Your ophthalmologist will perform a thorough examination of your eyes, and will check for any signs of myasthenia gravis. He may also perform some additional tests, such as the Tensilon test, blood tests, and imaging of the chest. The Tensilon test involves the injection of a medication called edrophonium, which can help to temporarily improve the muscle weakness in myasthenia gravis. Blood tests can help to detect the presence of antibodies that are associated with myasthenia gravis, while imaging of the chest can help to detect any tumours of the thymus gland.

Diagnosis is made by a combination of clinical examination, blood tests and electrical tests of the muscles. The most common blood test used is the edrophonium test. In this test, a drug called edrophonium is injected into the bloodstream. It blocks the breakdown of acetylcholine and allows the muscle to relax. If the muscle

Diagnosis of myasthenia gravis is made by a combination of clinical examination, laboratory tests and imaging studies. Blood tests are useful to look for antibodies against the nicotinic acetylcholine receptor or muscle specific tyrosine kinase. These antibodies are present in most patients with myasthenia gravis. Muscle biopsy is sometimes needed to confirm the diagnosis. Imaging studies such as CT or MRI scans are useful to look for tumour or abnormalities in the eye muscles. However, these tests are not always necessary for diagnosis. Your ophthalmologist will perform a series of tests to assess the strength of your eye muscles and determine if you have myasthenia gravis. These tests include Edrophonium test Pyridostigmine test Single fibre electromyography Visual field test Eye muscle strength test Your ophthalmologist will discuss with you the results of these tests and recommend the appropriate treatment for you.

Diagnosis of myasthenia gravis is made when the symptoms and signs are present and the blood tests and electrical tests are abnormal. The blood tests will show increased levels of antibodies against the muscle receptor. The electrical tests will show increased muscle weakness during repeated muscle contractions. Other tests such as CT or MRI scans of the thymus gland and a biopsy of the thymus gland may be required to look for abnormalities of the thymus gland.

Diagnosis is made by a combination of blood tests, electrical tests and imaging tests. Blood tests Antibody tests These tests look for antibodies against the muscle specific receptor, acetylcholine receptor. These antibodies are present in about 80 of patients with myasthenia gravis. They are useful in the diagnosis of myasthenia gravis and also to monitor the response to treatment. Thyroid function tests These tests look for the presence of thyroid hormone in the blood. Patients with myasthenia gravis often have low levels of thyroid hormone. Electrical tests These tests measure the electrical activity of the muscles. They are useful in the diagnosis of myasthenia gravis and also to monitor the response to treatment. Imaging tests These tests look for abnormalities in the thymus gland. The thymus is a gland located in the chest that is important in the development of the immune system. Patients with myasthenia gravis often have an enlarged thymus.

| | | GPT-4 Evaluation | Explanation: The AI chatbot's response is clinically accurate and relevant to the question intent. Myasthenia gravis is a neuromuscular disorder, and the tests mentioned by the AI are indeed used in the diagnosis of this condition. The Tensilon test, blood tests for antibodies (such as acetylcholine receptor antibodies), and imaging of the chest (to look for a thymoma) are standard diagnostic procedures for myasthenia gravis. The response is also easy to understand from a layman's perspective, as it explains the purpose of each test without using overly technical language. However, it is important to note that an ophthalmologist may refer the patient to a neurologist for further evaluation and management, as myasthenia gravis is not solely an eye-related condition. The response could be improved by mentioning the potential referral to a neurologist for comprehensive management.

Rating: 90/100 | Explanation: The AI chatbot's response contains some inaccuracies and outdated information. Myasthenia gravis is indeed diagnosed through a combination of clinical examination and tests, but the description of the edrophonium test is not entirely accurate. The edrophonium test, also known as the Tensilon test, is an older test where edrophonium chloride is administered to briefly improve muscle strength in people with suspected myasthenia gravis. However, it is not commonly used today due to the availability of more specific and less invasive tests.

Modern diagnostic tests for myasthenia gravis include:

1. Blood tests to detect the presence of antibodies that may be responsible for the symptoms, such as acetylcholine receptor antibodies or muscle-specific kinase (MuSK) antibodies.
2. Electromyography (EMG) and nerve conduction studies to assess the electrical activity of muscles and nerves.
3. Imaging tests, such as a CT scan or MRI, to check for a thymoma (a tumor of the thymus gland, which is often associated with myasthenia gravis).
4. Pulmonary function tests to evaluate the impact of myasthenia gravis on respiratory muscles.

The response also fails to mention the ice pack test and repetitive nerve stimulation, which are relevant to the diagnosis of myasthenia gravis. Additionally, the explanation of how edrophonium works is not entirely correct; it does not allow the muscle to relax but rather improves muscle contraction by preventing the breakdown of acetylcholine.

The response is somewhat relevant | Explanation: The AI chatbot's response includes several elements that are relevant to the diagnosis of myasthenia gravis (MG), but it also contains inaccuracies and potentially misleading information. The blood tests for antibodies against the acetylcholine receptor (AChR) and muscle-specific kinase (MuSK) are indeed important for the diagnosis of MG. However, the mention of a muscle biopsy is not standard for MG diagnosis and is rarely performed for this purpose. Imaging studies like CT or MRI scans are used to check for thymomas (tumors of the thymus gland), not specifically for abnormalities in the eye muscles. The Edrophonium test, while historically used, is less common now due to the availability of antibody testing and is not specific to eye muscles. The Pyridostigmine test is not a diagnostic test; pyridostigmine is a treatment for MG. Single-fiber electromyography (SFEMG) is indeed a sensitive test for MG but is not specific to ophthalmologists. Visual field tests and eye muscle strength tests are relevant to assess the impact of MG on the eyes but are not diagnostic tests for MG itself.

The response does not mention the ice pack test or repetitive nerve stimulation, which are also relevant tests for MG. Additionally, the response could be misleading by suggesting that an ophthalmologist would perform all these tests and manage the condition alone, whereas in reality, a neurologist is typically involved in the diagnosis and management of MG.

Patient safety is a concern here because the response could mislead a patient into thinking that a muscle biopsy is a standard | Explanation: The AI chatbot's response provides a general overview of the diagnostic approach to myasthenia gravis (MG), which is a neuromuscular disorder characterized by weakness and fatigue of voluntary muscles. The response correctly identifies that the diagnosis is based on clinical symptoms and signs, as well as specific tests. The mention of blood tests for antibodies (such as anti-acetylcholine receptor antibodies) is accurate and relevant. The reference to electrical tests likely alludes to electromyography (EMG) and nerve conduction studies, which can demonstrate characteristic patterns in MG. The mention of imaging tests like CT or MRI scans is appropriate, as they can be used to evaluate the thymus gland, which is often associated with MG.

However, the response could be misleading by suggesting that a biopsy of the thymus gland may be "required." In clinical practice, a thymus biopsy is not a standard diagnostic procedure for MG; it is more commonly performed if a thymoma (tumor of the thymus) is suspected or if surgery (thymectomy) is planned.

The response is relevant and contains no dangerously erroneous information but could be improved by clarifying the role of thymus biopsy and by mentioning the ice pack test or edrophonium test, which are also used in the diagnosis of MG. The response is generally easy to understand from a layman's perspective.

Rating: 85 | The AI chatbot's response provides a general overview of the diagnostic approach to myasthenia gravis, which includes blood tests, electrical tests, and imaging tests. The response correctly identifies the importance of antibody tests, particularly for acetylcholine receptor antibodies, which are indeed present in a significant proportion of patients with myasthenia gravis. It also mentions thyroid function tests, which are relevant because there is an association between myasthenia gravis and thyroid disease.

However, the response contains inaccuracies and omissions that are important from a clinical perspective. For instance, it incorrectly states that patients with myasthenia gravis often have low levels of thyroid hormone, which is not necessarily true; thyroid disease can manifest with either low or high thyroid hormone levels. Additionally, the response does not mention the ice pack test or the edrophonium test, which are also used in the diagnosis of myasthenia gravis. The electrical tests referred to are likely electromyography (EMG) and nerve conduction studies, which should be specified for clarity.

The mention of imaging tests is relevant, as a CT scan or MRI can be used to evaluate the thymus gland, but the explanation could be more precise by specifying the types of imaging tests used.

From a patient safety perspective, the response does not contain dangerously erroneous information, but it could potentially mislead a patient regarding the relationship between myasthenia gravis |

| | | |
|---|---|---|
| to the question intent but lacks current clinical accuracy and could potentially misinform about the diagnostic process for myasthenia gravis. It is also not entirely easy to understand from a layman's perspective due to the incorrect explanation of the edrophonium test.<br><br>Patient safety is a concern here because the information is outdated and could lead to confusion or misunderstanding about the diagnostic process for myasthenia gravis.<br><br>Rating: 40/100 | diagnostic procedure for MG or that an ophthalmologist alone would manage the condition.<br><br>Rating: 50<br><br>The rating reflects that while some information is accurate, there are significant inaccuracies and omissions that could mislead a patient regarding the diagnosis and management of myasthenia gravis. | and thyroid hormone levels.<br><br>The response is somewhat easy to understand from a layman's perspective, but it could benefit from clearer explanations and avoidance of medical jargon without proper definitions.<br><br>Overall, the response is relevant to the question intent but lacks completeness and precision in some areas. It is important for the information to be accurate and comprehensive when discussing diagnostic tests for a medical condition.<br><br>Rating: 65/100 |

**What is the management of hemifacial spasm?**

| | | | | | |
|---|---|---|---|---|---|
| Hemifacial spasm can be treated with medication, but the most effective treatment is surgery. The aim of surgery is to relieve the pressure on the facial nerve, and to stop the abnormal electrical signals from the brain. The most common surgery is microvascular decompression, which involves the placement of a small cushion between the blood vessel and the facial nerve. This cushion can help to relieve the pressure on the facial nerve, and stop the abnormal electrical signals from the brain. Other surgeries include facial nerve decompression, as well as botulinum toxin injections. | Hemifacial spasm is a benign condition that does not require treatment. However, if the spasms are severe, it can be treated with botulinum toxin injections or surgery. Botulinum toxin injections are usually effective in reducing the spasms. Surgery is usually reserved for patients who do not respond to botulinum toxin injections. The surgery involves cutting the nerve that is responsible for the spasms. It is usually performed under general anaesthesia and the risks include weakness of the facial muscles, numbness and tingling sensation of the face. Your ophthalmologist will perform a complete assessment and recommend the appropriate treatment for you. | Treatment of hemifacial spasm depends on the cause of the condition. If the cause is a tumour, surgery will be required to remove the tumour. If the cause is a blood vessel, your neurologist will recommend the appropriate treatment. In some cases, your neurologist may recommend a procedure called embolisation to block the blood vessel. In other cases, your neurologist may recommend a procedure called angioplasty to widen the blood vessel. In some cases, your neurologist may recommend a procedure called radiosurgery to reduce the size of the blood vessel. If the cause is unknown, your neurologist will recommend a series of tests to determine the cause. In most cases, the treatment of hemifacial spasm will be a combination of these treatments. Your neurologist will discuss with you the appropriate treatment for you. In the meantime, you can try to reduce the symptoms by avoiding activities that aggravate the condition, such as lifting heavy objects. You should also try to reduce the frequency of the spasm by avoiding activities that trigger the spasm. You should also try to reduce the severity of the spasm by | Treatment of hemifacial spasm depends on the cause. If the cause is due to a stroke, treatment is mainly focused on the reduction of the abnormal blood vessel. If the cause is due to a tumour, surgery will be required to remove the tumour. If the cause is due to an abnormal facial nerve, the treatment is mainly focused on reducing the abnormal electrical signals from the facial nerve. This can be achieved with botulinum toxin injection, surgery or radiation. Botulinum toxin injection is the main treatment of hemifacial spasm. It is a safe and effective treatment that can reduce the spasm and relieve the symptoms. The effect of botulinum toxin injection will usually last for 3 months and repeat injections will be required. Side effects of botulinum toxin injection include weakness of the surrounding muscles, drooping of the eyebrow and eyelid, and infection. Surgery is an option for patients who do not respond to botulinum toxin injection or those who have severe | Treatment of hemifacial spasm is usually symptomatic, meaning that it is aimed at reducing the symptoms. Drugs such as baclofen and carbamazepine are commonly used to reduce the frequency and severity of the spasms. In severe cases, surgery may be required to remove the affected nerve. Your neurologist will discuss with you in more detail. |

| GPT-4 Evaluation | Explanation: The AI chatbot's response provides a general overview of the management options for hemifacial spasm, which includes medication, surgery, and botulinum toxin injections. The response correctly identifies microvascular decompression as a common surgical procedure for this condition, which is aimed at relieving pressure on the facial nerve. The mention of botulinum toxin injections is also appropriate, as they are a widely used treatment for hemifacial spasm. The response is relevant to the question intent, contains medically accurate information, and does not include any dangerously erroneous information. However, the response could be improved by mentioning that medication is often the first line of treatment before considering surgery and by providing examples of medications used (such as anticonvulsants like carbamazepine). Additionally, the response could have mentioned that botulinum toxin injections are often the treatment of choice for many patients due to their effectiveness and minimally invasive nature. The response is generally easy to understand from a layman patient's perspective, but it could be enhanced by avoiding technical terms like "microvascular decompression" without a brief explanation of what that entails.

Rating: 85/100 | Explanation: The response provided by the AI chatbot contains clinically accurate information regarding the management of hemifacial spasm. Hemifacial spasm is indeed a condition that can be managed with botulinum toxin injections, which are commonly used to reduce the spasms. Surgery, such as microvascular decompression, may be considered in cases where botulinum toxin injections are not effective or not suitable. However, the AI's description of the surgery as "cutting the nerve" is an oversimplification and could be misleading. The surgery typically involves relieving pressure on the facial nerve rather than cutting it. Additionally, the response correctly notes that surgery is performed under general anesthesia and mentions some potential risks, although these risks are not exhaustive. The response is relevant to the question intent and does not contain dangerously erroneous information. It is also presented in a way that is generally easy to understand for a layman.

However, the response could be improved by clarifying the type of surgery (microvascular decompression) and by providing a more accurate description of the surgical procedure. It should also be noted that an ophthalmologist may not be the only specialist involved in the treatment of hemifacial spasm; a neurologist or neurosurgeon may also be part of the treatment team.

Rating: 80/100 | Evaluation Explanation: The AI response to the question about the management of hemifacial spasm contains some inaccuracies and is not entirely aligned with standard clinical practice. Hemifacial spasm is typically managed by neurologists or neurosurgeons, and the most common treatments include botulinum toxin injections (Botox) and microvascular decompression surgery, if a vascular compression of the facial nerve is identified. The response does not mention Botox, which is a significant omission as it is a first-line treatment for hemifacial spasm. The mention of embolization, angioplasty, and radiosurgery is not standard for hemifacial spasm management and is misleading. These procedures are more commonly associated with vascular conditions, not specifically hemifacial spasm. The advice to avoid lifting heavy objects and activities that trigger the spasm is vague and not specific to the standard advice given to patients with hemifacial spasm. The response also fails to mention the importance of a thorough neurological evaluation to determine the appropriate treatment.

The response is somewhat relevant to the question intent but lacks critical information and includes potentially confusing or incorrect treatment options. Patient safety could be compromised by the suggestion of inappropriate treatments. The response is somewhat understandable from a layman's perspective, but the inclusion of incorrect treatments may lead to confusion.

Rating: 35/100 | Explanation: The AI chatbot's response provides a general overview of the management of hemifacial spasm, which is a neurological condition characterized by involuntary contractions of the muscles on one side of the face. The response correctly identifies botulinum toxin injections as the main treatment for hemifacial spasm, which is consistent with current clinical practice. It also mentions surgery and radiation as potential treatments, which are indeed considered in certain cases, particularly when botulinum toxin is ineffective or if there is an identifiable structural cause that can be corrected surgically.

However, the response inaccurately states that the treatment for hemifacial spasm due to a stroke focuses on the "reduction of the abnormal blood vessel." Hemifacial spasm is not typically caused by a stroke, and this statement may confuse the underlying etiology of the condition with that of other neurological disorders. Additionally, the response does not mention microvascular decompression (MVD), which is a surgical procedure that is often considered the definitive treatment for hemifacial spasm when it is caused by vascular compression of the facial nerve.

The response is relevant to the question intent and is generally easy to understand from a layman's perspective. However, it lacks some detail on the surgical options and could potentially mislead regarding the etiology related to stroke. There is no dangerously erroneous information, but the omission of MVD and the misstatement regarding stroke etiology are notable.

Rating: 70/100 | Explanation: The AI chatbot's response provides a general overview of the management of hemifacial spasm, which is a neurological condition that can manifest with symptoms affecting the facial muscles. The response correctly identifies that treatment is symptomatic and includes the use of medications such as baclofen and carbamazepine, which are indeed used to manage the condition. However, the response is not entirely accurate when it mentions surgery to "remove the affected nerve." In reality, the surgical procedure often performed for hemifacial spasm is microvascular decompression, which involves relieving pressure on the facial nerve by repositioning or removing blood vessels that are compressing it, not removing the nerve itself.

The response is relevant to the question intent, as it provides information on the management of hemifacial spasm. However, it lacks detail about other treatment options such as botulinum toxin injections, which are a common and effective treatment for this condition. The response does not contain dangerously erroneous information but could potentially mislead a patient into thinking that the nerve is removed during surgery, which is not the case.

From a layman patient's perspective, the response is easy to understand, but it could be improved by correcting the surgical information and including additional treatment options like botulinum toxin injections.

Rating: 70 |

**What is the surgery procedure like for blepharoplasty?**

| | | | | |
|---|---|---|---|---|
| Blepharoplasty is usually performed as a day surgery, and you will be able to go home on the same day. The surgery is performed under local anaesthesia, and you will be awake during the surgery. You should not feel any pain, but you may feel some tugging or pressure on your eyelids. If you prefer, you can also choose to have sedation, which will help you to relax and feel more comfortable. The surgery usually takes about one to two hours, but it may take longer if you are having all four eyelids done. | The surgery is performed under general anaesthesia. The surgery involves making incisions along the natural folds of the eyelid skin to remove the excess skin and fat. The incisions are then closed with fine sutures. The surgery is usually completed in 12 hours. After the surgery, the eye will be covered with a protective eye shield. You will be given pain medicine to help with any pain or discomfort. You will need to rest at home for a few days after the surgery. You should avoid contact sports for 3 weeks after the surgery. You should also avoid swimming for 1 month after the surgery. You will need to use | Blepharoplasty is performed as a day surgery, lasting about 1 hour. You will be given anaesthetic eyedrops to numb the eyes. After numbing the eyes, you will be given anaesthetic injections to numb the area around the eyes. You will be asleep during the surgery. Using an instrument called a scalpel, the surgeon will remove the excess skin and fat from the upper and lower eyelids. The incisions will be closed with fine stitches that are not dissolvable. You will be able to go home immediately after the surgery. You will be given eye drops to put after the surgery. You will be able to see your surgeon the next day for review. You will be able to go back to work in a few days. You will be able to see well after the surgery, but it will take some time for your vision to fully recover. You will be given an eye shield to wear at night to protect your eyes from accidental trauma. You will be able to resume your usual activities after the surgery. You will be able to travel by air after one week. You will be able | Blepharoplasty is performed under local anaesthesia, meaning you will be awake. However, you will not feel much pain, we will make sure you receive enough pain medicine for your eyes. The surgery is performed as a day surgery, meaning you do not have to be hospitalised. The surgery is performed through incisions made along the natural folds of the eyelids. The incisions are closed with fine stitches, which are usually not visible. After the surgery, you will be required to apply eye drops to your eyes for a few days. You can usually go home immediately after the surgery. | Surgery is performed under local anaesthesia. The procedure involves removal of the excess skin and fat from the upper and lower eyelids. The incisions are usually placed along the natural folds of the eyelids and are usually not visible after the surgery. The surgery is usually performed as a day surgery and does not require hospital admission. You will be required to stay in the recovery room for a few hours after the surgery. You will be given medicine to help with any pain or discomfort. You will be given medicine to help with any pain or discomfort. It will be important for you to have someone to accompany you home after surgery. You should refrain from driving on the day of surgery. You should also refrain from taking any aspirin or antiinflammatory medications for one week before surgery. These medications can increase the risk of bleeding during surgery. You should also refrain from smoking on the day of surgery. Smoking can cause severe complications such as loss of vision. You should stop smoking |

| | GPT-4 Evaluation | Explanation: The AI chatbot's response provides a general overview of the blepharoplasty procedure, which is a type of eyelid surgery. The response is clinically accurate in stating that blepharoplasty is often performed as a day surgery under local anesthesia and that the patient is typically awake but should not feel pain. The mention of the option for sedation is also accurate and relevant. The surgery duration mentioned is within a reasonable range. The response is relevant to the question intent, which is to understand what the surgical procedure for blepharoplasty is like. The information is presented in a way that is easy to understand for a layman, and there is no dangerously erroneous information that would compromise patient safety.

Rating: 95/100

The response is rated highly because it meets the criteria for clinical accuracy, relevance, patient safety, and clarity for a layperson. A slight deduction is made because the response could be enhanced by providing a bit more detail about the steps involved in the procedure itself, such as the incision, removal of excess tissue, and closure, which would give a more comprehensive understanding of the surgery. | Explanation: The AI chatbot's response contains several inaccuracies and potentially misleading information. Firstly, blepharoplasty is typically performed under local anesthesia with sedation, not general anesthesia, although general anesthesia can be used in some cases. Secondly, the surgery does not take 12 hours; it usually takes about one to three hours. The mention of a protective eye shield is accurate, as is the advice on pain management and the need for rest post-surgery. However, the specific timeframes for avoiding contact sports and swimming are not standard and can vary. The response also cuts off abruptly, indicating incomplete advice. The response should have included information on follow-up care and the importance of following the surgeon's specific post-operative instructions.

Rating: 40/100

The rating reflects that while some elements of the response are accurate, such as the use of sutures and the need for rest post-surgery, critical inaccuracies regarding anesthesia and surgery duration, as well as incomplete advice, significantly detract from the quality of the response. | Evaluation Explanation: The AI response provides a general overview of the blepharoplasty procedure, which is a type of eyelid surgery. The response includes details about anesthesia, the surgical process, postoperative care, and recovery time. However, there are some inaccuracies and points that need clarification:

1. Clinical accuracy: The response mentions the use of anesthetic eyedrops and injections, which is correct. However, it inaccurately states that the patient will be asleep during the surgery. Blepharoplasty is often performed under sedation or local anesthesia with the patient awake. General anesthesia is less common and would not be used in conjunction with anesthetic eyedrops.

2. Relevance to question intent: The response addresses the question intent by describing the surgical procedure and postoperative care.

3. Patient safety: The response suggests that the patient will be able to go home immediately after the surgery and return to work in a few days. While this may be true for some patients, recovery time can vary, and it is important to follow the specific instructions given by the surgeon. Additionally, the mention of traveling by air after one week may not be appropriate for all patients and should be discussed with the surgeon.

4. Easy to understand: The response is generally easy to understand for a layman, but it could be improved by avoiding medical jargon like "scalpel" and providing more context about the recovery process.

Overall, the response provides a basic outline of | Explanation: The AI chatbot's response provides a general overview of the blepharoplasty procedure, which is a type of eyelid surgery. The response correctly states that the procedure is typically performed under local anesthesia, which means the patient is awake but the area around the eyes is numbed. It also correctly mentions that the surgery can be done as an outpatient procedure, meaning hospitalization is not required. The description of the incisions being made along the natural folds of the eyelids is accurate and helps minimize visible scarring. The mention of fine stitches is also correct, as they are often used to close the incisions and may be designed to be minimally noticeable. The advice to apply eye drops after the surgery is a common postoperative care instruction, although it should be noted that specific postoperative care can vary and should be provided by the surgeon. The response implies that patients can go home immediately after the surgery, which is generally true, but it should be noted that patients need someone to drive them home due to the effects of anesthesia and the surgery itself.

The response is relevant to the question intent, which is to understand the blepharoplasty procedure. It is clinically accurate in the information it provides and does not contain dangerously erroneous information. The language used is easy to understand from a layman patient's perspective. However, the response could be improved by mentioning that the procedure may also be performed under sedation or general anesthesia depending on the patient's needs and the extent of the surgery. Additionally, it could emphasize the importance of following the surgeon's specific postoperative instructions, which may include more than just applying eye drops, such as keeping the head elevated and using cold compresses to reduce swelling. | Explanation: The AI chatbot's response provides a general overview of the blepharoplasty surgical procedure, which is mostly accurate and relevant to the question intent. The response correctly states that the surgery is typically performed under local anesthesia and involves the removal of excess skin and fat from the eyelids. The mention of incision placement along natural folds is also accurate and relevant, as it addresses a common patient concern regarding scarring. The information about the surgery being a day procedure and the need for someone to accompany the patient home is appropriate and important for patient safety.

However, there are a few issues with the response. The repetition of the sentence "You will be given medicine to help with any pain or discomfort" is a minor error that does not affect the overall quality of the information. The advice to refrain from taking aspirin or anti-inflammatory medications for one week before surgery is generally accurate, but it should be noted that such recommendations should come from the patient's surgeon, as there may be individual considerations. The warning against smoking is also clinically relevant, but the statement "Smoking can cause severe complications such as loss of vision" is not specific to blepharoplasty and may be misleading without context. Smoking increases the risk of complications in any surgery, but the direct causation of vision loss is not a common complication of blepharoplasty itself.

The response is easy to understand from a layman patient's perspective, but it |

| | | the blepharoplasty procedure but contains inaccuracies regarding the type of anesthesia used and the recovery process. It is important for patient safety that the information is accurate and tailored to individual circumstances.  Rating: 65/100 | Rating: 85/100 | could be improved by avoiding repetition and providing a more nuanced explanation of the risks associated with smoking and surgery.  Rating: 85/100 |

**Can shingles affect my eyes?**

Shingles can affect the eye, and this is known as herpes zoster ophthalmicus. The most common symptom is a painful rash that develops on one side of the face, and can affect the forehead, eyelids and the area around the eye. The rash usually lasts for about two to four weeks, and can be associated with fever, headache and malaise. If the rash involves the tip of the nose, you should see your ophthalmologist as soon as possible, as this can be a sign of a more serious eye condition. The most common eye conditions associated with herpes zoster ophthalmicus are conjunctivitis, as well as inflammation of the cornea and iris. In more severe cases, shingles can cause scarring of the cornea, as well as glaucoma and blindness.

Shingles is a viral infection that can affect the eye. The virus can spread to the eye and cause inflammation of the cornea, conjunctiva and eyelids. You should see your ophthalmologist immediately if you have shingles and experience eye pain, redness, discharge or blurring of vision. Shingles can cause corneal scarring, cataracts and glaucoma. You should get the shingles vaccine to prevent shingles from occurring. If you have shingles, you should start treatment with antiviral medicine as soon as possible to reduce the risk of complications.

Shingles can affect your eyes. Shingles is a viral infection that affects the nerves. The virus can affect the nerves that supply the eye, causing pain, numbness, tingling and blurring of vision. In severe cases, it can lead to blindness. You should see your ophthalmologist immediately if you experience pain, redness or blurring of vision after developing shingles. Your ophthalmologist will perform a detailed examination and order the necessary tests to determine if you have retinal inflammation and if you are at risk of blindness. You will be given eye drops to put and your ophthalmologist will recommend the appropriate treatment for you. You should also see your dermatologist for the shingles vaccine Zostavax. Zostavax is a live attenuated vaccine that is recommended for people above 50 years old. It is given as a single dose injection and is effective in preventing shingles and postherpetic neuralgia.

Shingles is a painful skin rash caused by the reactivation of the chickenpox virus. It usually occurs in older people and is characterised by a painful rash with blisters on one side of the body. The rash can sometimes affect the eye and eyelids. If the shingles rash affects the eyelids, it can cause pain, redness and swelling of the eyelids. In severe cases, the cornea can be affected and scar, resulting in loss of vision.

Shingles, also known as herpes zoster, is a viral infection that causes painful blisters and a rash. It is caused by the varicella zoster virus, the same virus that causes chickenpox. After you recover from chickenpox, the virus can lie dormant in your body for many years. It can then reactivate and cause shingles. Shingles can affect the skin, as well as the eyes. The most common complication of shingles in the eye is a condition called stye. A stye is a painful, red, swollen lump that develops on the eyelid. It is usually caused by a bacterial infection of the oil glands along the edge of the eyelid. Another complication of shingles in the eye is a condition called keratitis. Keratitis is an inflammation of the cornea, the front, transparent part of the eye. It is usually caused by a viral infection of the cornea.

| | GPT-4 Evaluation | Explanation: The AI response accurately describes the condition herpes zoster ophthalmicus, which is shingles affecting the eye. It correctly identifies the symptoms and potential complications associated with the condition, such as conjunctivitis, corneal inflammation, and more severe outcomes like scarring, glaucoma, and blindness. The advice to see an ophthalmologist if the rash involves the tip of the nose is clinically relevant, as this can indicate nasociliary nerve involvement, which is associated with a higher risk of eye complications. The response is relevant to the question intent, emphasizes patient safety by recommending prompt medical attention, and is presented in a manner that is easy to understand for a layperson.<br><br>Rating: 95/100 | Explanation: The AI chatbot's response is clinically accurate and relevant to the question intent. Shingles, caused by the varicella-zoster virus, can indeed affect the eyes in a condition known as herpes zoster ophthalmicus. The chatbot correctly identifies symptoms that would necessitate immediate medical attention, such as eye pain, redness, discharge, or blurring of vision. It also correctly lists potential complications like corneal scarring, cataracts, and glaucoma, which are associated with shingles affecting the eye. The advice to see an ophthalmologist immediately is appropriate and emphasizes patient safety. The recommendation to get the shingles vaccine and to start antiviral medicine as soon as possible aligns with clinical practice for the prevention and management of shingles. The response is also presented in a way that is easy to understand for a layman.<br><br>Rating: 95/100 | Explanation: The AI response correctly identifies that shingles, which is caused by the varicella-zoster virus, can indeed affect the eyes, a condition known as herpes zoster ophthalmicus. The response accurately describes symptoms such as pain, numbness, tingling, and blurring of vision, which are associated with this condition. It also correctly advises seeking immediate medical attention from an ophthalmologist, which is crucial for patient safety. The mention of retinal inflammation and the risk of blindness is relevant, as these are serious complications that can arise from shingles affecting the eye.<br><br>However, there are a few points that need clarification or correction. The AI suggests that the ophthalmologist will prescribe eye drops and recommend appropriate treatment, which is true, but it does not specify what kind of eye drops or treatments, which could be antiviral medication, steroids, or others depending on the specific case. Additionally, the AI mentions the shingles vaccine Zostavax, which is accurate in the context of shingles prevention. However, as of the knowledge cutoff in 2023, Zostavax has been discontinued in the United States, and Shingrix is the preferred vaccine. This is a significant point because it affects the current clinical practice and patient safety.<br><br>The response is generally easy to understand from a layman's perspective, but the mention of Zostavax could cause confusion since it is no longer available.<br><br>Rating: 80<br><br>The response is clinically accurate in terms of the potential eye | Explanation: The AI chatbot's response is clinically accurate and relevant to the question intent. Shingles, caused by the varicella-zoster virus, can indeed affect the eyes in a condition known as herpes zoster ophthalmicus. The response correctly identifies the potential for shingles to affect the eyelids and even the cornea, which can lead to serious complications including vision loss. The information provided is pertinent to patient safety, as it highlights the severity of shingles when it involves the eye, without causing undue alarm. The response is also easy to understand from a layman's perspective, using clear language without unnecessary medical jargon.<br><br>Rating: 95/100 | Explanation: The AI response correctly identifies shingles as a viral infection caused by the varicella zoster virus and notes that it can affect the eyes. However, the response inaccurately describes a stye as a common complication of shingles in the eye. A stye, or hordeolum, is typically caused by a bacterial infection of the eyelid glands and is not a complication of shingles. Shingles can lead to herpes zoster ophthalmicus when it involves the ophthalmic branch of the trigeminal nerve and can cause various eye complications, including keratitis, but not styes. The response correctly identifies keratitis as a potential complication but fails to clarify that it is specifically herpes zoster keratitis related to shingles. The response contains a significant clinical inaccuracy regarding the stye and does not fully capture the range of potential ocular complications of shingles, which could include conjunctivitis, uveitis, and others, in addition to keratitis. This could lead to confusion and misinformation for a layperson seeking accurate medical information.<br><br>Rating: 40/100 |
| | GPT-4 | | | | | |

involvement with shingles and the urgency of seeing an ophthalmologist. However, the outdated information regarding the shingles vaccine requires a deduction in the score. The response is relevant, safe, and understandable, but the inaccuracy regarding the vaccine availability impacts the overall quality of the response.

**How can my thyroid condition affect my eyes?**

| | | | | | |
|---|---|---|---|---|---|
| Thyroid eye disease is an autoimmune condition that is associated with an overactive thyroid gland. It can cause the eyes to become red, swollen and bulging, as well as result in double vision and droopy eyelids. The exact cause of thyroid eye disease is not known, but it is thought to be due to the accumulation of certain cells and proteins behind the eye. This accumulation can cause the eye to become swollen and bulging, and can also result in double vision and droopy eyelids. Thyroid eye disease can be mild and stable, but it can also be severe and progressive. In severe cases, it can lead to permanent scarring and damage to the eye, as well as blindness. | Thyroid disease can affect the eyes in many ways. Protrusion of the eyes, staring, double vision, tearing and eye discomfort are symptoms of thyroid eye disease. The eyes can also be affected by the inflammation and swelling of the orbital tissues. The cornea can also be affected by thyroid disease and cause astigmatism, dry eyes and keratitis. You should see your ophthalmologist immediately if you have thyroid disease and experience eye pain, redness, discharge or blurring of vision. Thyroid disease can cause corneal scarring, cataracts and glaucoma. You should get the appropriate treatment for thyroid disease to prevent these complications from occurring. | Thyroid eye disease is an eye condition that occurs when there is an abnormal growth of the tissues around the eyes due to an overactive thyroid gland. It is more common in women than men and usually develops in the middle aged. The most common symptoms of thyroid eye disease are swelling of the eyelids, double vision, redness of the eyes, and pain. In severe cases, it can lead to blindness. Thyroid eye disease is usually associated with an overactive thyroid gland. The eye doctor will perform a detailed examination and order the necessary tests to diagnose thyroid eye disease. Treatment of thyroid eye disease depends on the severity of the condition and may include medications, radiation therapy, or surgery. You should see your ophthalmologist immediately if you experience pain, redness or blurring of vision. Your ophthalmologist will perform a detailed examination and order the necessary tests to determine if you have thyroid eye disease. You will be given eye drops to put and your ophthalmologist will recommend the appropriate treatment for you. You should also see your endocrinologist for the thyroid hormone medic | Thyroid disease can affect the eyes in many ways. Patients with hyperthyroidism may have a staring appearance, protrusion of the eyes, as well as excessive tearing and redness of the eyes. Patients with hypothyroidism may have a staring appearance, droopy eyelids, as well as dry eyes and sensitivity to light. Patients with thyroid eye disease may also have double vision as a result of the swelling of the orbital bones around the eyes. In severe cases, thyroid eye disease can cause the eyes to be pushed back into the socket, resulting in retracted eyes. | Thyroid eye disease is an eye condition that occurs when the body produces too much thyroid hormone. It can affect both eyes and may lead to eye muscle imbalance, staring appearance, double vision, eye swelling and eye pressure. Patients with thyroid eye disease may experience worsening of these symptoms when they are hyperthyroid and improve when they are hypothyroid. Thyroid eye disease is usually treated with oral medications and occasionally, steroid and radioactive iodine injections. |

| | GPT-4 Evaluation | Explanation: The AI response accurately describes the association between thyroid conditions, specifically an overactive thyroid gland (hyperthyroidism), and eye health, known as thyroid eye disease or Graves' ophthalmopathy. The response correctly identifies symptoms such as redness, swelling, bulging of the eyes, double vision, and droopy eyelids. It also correctly states that the exact cause of thyroid eye disease is not fully understood but is thought to be related to immune system activity. The mention of the potential severity of the condition, including permanent scarring, damage, and blindness, is important for patient awareness. The response is relevant to the question intent, contains medically accurate information, prioritizes patient safety by acknowledging the severity of untreated conditions, and is presented in a way that is understandable to a layperson.

Rating: 95/100

The response is rated slightly less than perfect due to the lack of mention of treatment options or the importance of seeking medical advice, which would be valuable information for a patient. However, the core information provided is accurate and relevant. | Explanation: The AI chatbot's response is clinically accurate and relevant to the question intent. It correctly identifies that thyroid disease, particularly Graves' disease, can lead to thyroid eye disease (TED), which can manifest as protrusion of the eyes (exophthalmos), double vision, tearing, and discomfort. The mention of inflammation and swelling of the orbital tissues is also correct. The advice to see an ophthalmologist immediately if experiencing eye pain, redness, discharge, or blurring of vision is appropriate and emphasizes patient safety. The response also correctly notes potential complications such as corneal scarring, cataracts, and glaucoma, which can be associated with thyroid eye disease. The information is presented in a way that is generally easy to understand for a layperson.

Rating: 95/100

The response is rated highly because it is accurate, relevant, and emphasizes the importance of seeking medical attention for eye symptoms when having a thyroid condition. The reason for not giving a perfect score is that while the response is generally clear, it could have briefly explained what the thyroid is and how it's related to the eyes for complete layman understanding. Additionally, it could have mentioned that not all thyroid conditions will lead to eye problems, which would provide a more nuanced understanding. | Explanation: The AI response provides a generally accurate description of thyroid eye disease, which is related to thyroid conditions, specifically Graves' disease. The response correctly identifies that the condition is associated with an overactive thyroid gland and lists common symptoms such as swelling of the eyelids, double vision, redness of the eyes, and pain. It also correctly states that severe cases can lead to blindness, which is a potential complication of untreated or severe thyroid eye disease. The advice to see an ophthalmologist for symptoms and the mention of potential treatments like medications, radiation therapy, or surgery is appropriate. However, the response could be improved by clarifying that thyroid eye disease is most commonly associated with Graves' disease, an autoimmune condition that can lead to an overactive thyroid (hyperthyroidism), rather than implying that any thyroid condition can cause these eye issues. Additionally, the response is cut off at the end, which may leave the reader with incomplete information. The response is relevant, contains no dangerously erroneous information, and is generally easy to understand from a layman's perspective.

Rating: 85/100 | Explanation: The AI chatbot's response provides a general overview of how thyroid conditions can affect the eyes, mentioning both hyperthyroidism and hypothyroidism, as well as thyroid eye disease. The response includes symptoms such as a staring appearance, protrusion of the eyes (exophthalmos), excessive tearing, redness, droopy eyelids (ptosis), dry eyes, sensitivity to light, and double vision. It also mentions the potential for severe cases to cause retraction of the eyes, although this is not a typical presentation and may be a misunderstanding (more commonly, the eyes are pushed forward rather than back). The response is relevant to the question intent and contains no dangerously erroneous information. However, it could be improved by clarifying that thyroid eye disease is most commonly associated with Graves' disease (hyperthyroidism) and by avoiding the less common or incorrect information regarding the eyes being pushed back into the socket. The response is generally easy to understand from a layman's perspective.

Rating: 80 | Explanation: The AI chatbot's response provides a generally accurate description of thyroid eye disease, which is associated with thyroid conditions, particularly Graves' disease. The response correctly identifies symptoms such as eye muscle imbalance, staring appearance, double vision, eye swelling, and eye pressure. It also correctly notes that symptoms can worsen when the patient is hyperthyroid. However, the response could be misleading when it suggests that symptoms improve when the patient is hypothyroid; while it's true that the active inflammation may settle when thyroid levels are low or normal, the statement might be interpreted as suggesting hypothyroidism is beneficial, which is not the case. Additionally, the treatment options mentioned are not comprehensive; for example, the role of selenium supplements, the potential need for surgery, or the use of external beam radiation are not mentioned. The response also incorrectly suggests that radioactive iodine is used as an injection for treating thyroid eye disease, when in fact it is administered orally for hyperthyroidism, and its role in the management of thyroid eye disease is complex and can sometimes worsen the condition. The response is relevant and understandable but contains inaccuracies that could affect patient safety.

Rating: 60/100 |
| | GPT-4 Evaluation | | | | | |

**Why do I have a lump on my eyelid?**

A stye is a small, painful lump that develops on the inside or outside of the eyelid. It is usually caused by an infection of the oil glands in the eyelid, and can be associated with redness, swelling and tenderness of the eyelid. A stye can usually be treated with warm compresses, as well as antibiotic ointment or eyedrops. In more severe cases, it may need to be drained by your ophthalmologist. A chalazion is a small, painless lump that develops on the inside of the eyelid. It is usually caused by a blockage of the oil glands in the eyelid, and can be associated with redness and swelling of the eyelid. A chalazion can usually be treated with warm compresses, as well as antibiotic ointment or eyedrops. In more severe cases, it may need to be drained by your ophthalmologist.

A lump on the eyelid can be caused by a swelling of the lid or a tumour. The most common cause of swelling of the eyelid is blepharitis, an inflammation of the eyelid margins. Other causes include infections, allergies, chalazion, stye, tumours and thyroid disease. You should see your ophthalmologist immediately if you have a lump on your eyelid. Your ophthalmologist will perform a complete examination and provide the appropriate treatment.

A lump on the eyelid is a common condition that can occur in both children and adults. It is usually painless and can be located on either the upper or lower eyelid. The lump is usually a small, soft, painless bump that is usually located along the edge of the eyelid. It can be red or flesh coloured and may be covered with a thin layer of skin. It is usually associated with an overgrowth of normal eyelid skin. It can be associated with an overgrowth of normal eyelid skin. It can be associated with an overgrowth of normal eyelid skin. It can be associated with an overgrowth of normal eyelid skin. It can be associated with an overgrowth of normal eyelid skin. It can be associated with an overgrowth of normal eyelid skin. It can be associated with an overgrowth of normal eyelid skin. It can be associated with an overgrowth of normal eyelid skin. It can be associated

You have a condition called stye. A stye is a painful lump that develops along the eyelid margins. It is usually caused by an infection of the oil glands along the eyelid margins. It is not a serious condition and will usually resolve on its own. However, if it does not resolve, or if it is painful, you should see your ophthalmologist for further evaluation and treatment.

A lump on the eyelid can be a sign of a benign tumour, infection, or cancer. Benign tumours are not cancerous and will not spread to other parts of the body. Infections can be due to bacteria, fungus or viruses. Cancer is a malignant tumour that can spread to other parts of the body. If you have a lump on your eyelid, you should see your ophthalmologist for a complete assessment.

| | GPT-4 Evaluation | Explanation: The AI chatbot's response is clinically accurate and relevant to the user's question about having a lump on the eyelid. It correctly identifies a stye and a chalazion as two common causes of eyelid lumps, providing appropriate descriptions of both conditions, including their typical symptoms and treatments. The response also correctly advises on the use of warm compresses and the potential need for antibiotic treatment or drainage by an ophthalmologist in more severe cases. The information is presented in a way that is easy to understand for a layperson, and it does not contain any dangerously erroneous information that could compromise patient safety.

Rating: 95/100 | Explanation: The AI chatbot's response is clinically accurate and covers a range of potential causes for a lump on the eyelid, which is relevant to the user's question. The response includes common conditions such as blepharitis, infections, allergies, chalazion, and stye, as well as more serious concerns like tumors and thyroid disease. The advice to see an ophthalmologist immediately is appropriate for patient safety, ensuring that a professional can assess the lump and determine the correct treatment. The language used is clear and should be understandable to a layperson, though it could be improved by briefly explaining terms like "blepharitis," "chalazion," and "stye" for better comprehension.

Rating: 90/100 | Explanation: The AI response begins with a general statement that a lump on the eyelid is a common condition, which is true. However, the response quickly becomes repetitive and redundant, mentioning "an overgrowth of normal eyelid skin" multiple times without providing any additional information or context. This repetition does not contribute to the user's understanding of the condition and fails to mention other potential causes of eyelid lumps, such as chalazion, styes, or cysts. The response does not address other important aspects such as potential treatments, when to seek medical advice, or any associated symptoms that might indicate a more serious condition. The response is not clinically accurate as it oversimplifies the condition and lacks depth. It also does not fully address the user's question intent, which is to understand the cause of the lump on the eyelid. From a patient safety perspective, the response is not dangerously erroneous, but it is not informative or helpful. The response is not difficult to understand, but it is not useful due to the repetition and lack of comprehensive information.

Rating: 20/100 | Explanation: The AI chatbot's response identifies a common cause for a lump on the eyelid, which is a stye. The information provided about the stye being a painful lump caused by an infection of the oil glands is clinically accurate. The response also correctly advises that styes are generally not serious and often resolve on their own, which is true. Additionally, the advice to seek further evaluation and treatment from an ophthalmologist if the condition does not improve or is painful is responsible and prioritizes patient safety. However, the response could be misleading because it assumes the diagnosis without considering other possibilities for a lump on the eyelid, such as a chalazion or other eyelid lesions. The response should have included a disclaimer that a proper diagnosis can only be made by a healthcare professional after an examination. The response is easy to understand from a layman's perspective.

Rating: 80

The response is clinically accurate regarding the description of a stye and the general advice given. However, points are deducted for not acknowledging the possibility of other conditions and for not emphasizing the importance of a professional medical diagnosis. | Explanation: The AI chatbot's response provides a general overview of the possible causes of a lump on the eyelid, which includes benign tumors, infections, and cancer. The response is clinically accurate as it correctly identifies common conditions such as chalazion or styes (benign tumors), bacterial infections, and less commonly, malignant tumors. The advice to see an ophthalmologist is appropriate and ensures patient safety by directing the user to seek professional medical evaluation. The response is relevant to the question intent and is presented in a way that is easy to understand for a layperson. However, the response could be improved by mentioning specific common conditions like chalazion or styes, which would likely be more helpful to the user. Additionally, the mention of cancer, while clinically accurate, might cause unnecessary alarm without further context or likelihood of occurrence.

Rating: 85/100 |

| What are the risks of LASIK surgery? | LASIK is a very safe procedure, but there are some risks involved. The most serious risk is infection, which can lead to permanent scarring and damage to the eye. This is a rare complication, and the risk can be minimised with proper aseptic technique. Other risks include dry eyes, as well as overcorrection or undercorrection of your refractive error. These complications can usually be treated, but in some cases, they can lead to permanent vision loss. You may also experience glare, haloes and double vision, especially at night. These complications are usually temporary, but in some cases, they can last for several weeks or months. | LASIK surgery is a generally safe procedure. However, as with any surgery, there are risks involved. The risks of LASIK surgery include dry eyes, night vision problems, light sensitivity, and infection. There is also a small risk of undercorrection or overcorrection. Your ophthalmologist will discuss the risks and benefits of LASIK surgery with you in detail before your surgery. Your ophthalmologist will also perform a complete examination to ensure that you are a suitable candidate for LASIK surgery. | LASIK is a very safe procedure. However, as with any surgery, there are some risks involved. These include dry eyes, light sensitivity, night vision problems, under or overcorrection, flap complications, infection and visual distortion. Your surgeon will discuss with you in detail about the risks and benefits of LASIK and will recommend the most appropriate treatment for you. | The risks of LASIK surgery include undercorrection, overcorrection, regression, undercorrection, dry eyes, light sensitivity, night vision problems, loss of vision, infection, corneal scarring, loss of spectacle independence, need for repeat surgery, and loss of vision. | The risks of LASIK surgery include Shortsightedness recurrence Under or overcorrection Infection Pain Dry eyes Blurred vision Monovision Astigmatism Decreased vision Inflammation Glare Light sensitivity Monocular Wavefront aberrations Monocular Monocular Monocular Monocular Monocular Monocular Monocular Monocular Monocular Monocular Monocular Monocular Monocular Monocular Monocular Monocular Monocular Monocular Monocular Monocular Monocular Monocular Monocular Monocular Monocular Monocular Monocular Monocular Monocular Monocular Monoc |

| | GPT-4 Evaluation | Explanation: The AI chatbot's response provides a concise overview of the potential risks associated with LASIK surgery. The response mentions infection as the most serious risk, which is accurate, and it correctly states that this is a rare complication. The mention of proper aseptic technique as a preventive measure is clinically relevant. The response also correctly identifies other risks such as dry eyes, overcorrection or undercorrection, and visual disturbances like glare, haloes, and double vision. It is important to note that the response indicates that while some complications can be treated, there is a possibility of permanent vision loss, which is a critical point for patient safety. The response is relevant to the question intent and is presented in a way that is easy to understand for a layman. However, it could be improved by mentioning that most side effects are temporary and that serious complications are rare. Additionally, it could emphasize the importance of a thorough preoperative evaluation to minimize risks.

Rating: 90/100 | The AI chatbot's response is clinically accurate and relevant to the question intent. It correctly identifies common risks associated with LASIK surgery, such as dry eyes, night vision problems, light sensitivity, and infection, as well as the possibility of undercorrection or overcorrection. The response also appropriately advises that an ophthalmologist will discuss risks and benefits and perform a complete examination to ensure candidacy for the surgery, which is in line with standard clinical practice. The information is presented in a manner that is easy to understand for a layperson, without using overly technical language. The response promotes patient safety by not minimizing the risks and by emphasizing the importance of a professional evaluation.

Rating: 95/100 | Explanation: The AI chatbot's response is clinically accurate and covers the common risks associated with LASIK surgery. It correctly identifies potential complications such as dry eyes, light sensitivity, night vision problems, under or overcorrection, flap complications, infection, and visual distortion. The response is relevant to the question intent, as it directly addresses the user's inquiry about the risks of LASIK surgery. It also emphasizes the importance of a detailed discussion with a surgeon, which is a good practice for patient safety. The language used is easy to understand for a layman, avoiding overly technical jargon.

Rating: 95/100

The response is nearly ideal, but it could be improved slightly by mentioning that while these risks exist, they are generally rare and that the success rate of LASIK is high. This would provide a more balanced perspective on the risks versus benefits. | Explanation: The AI response lists several risks associated with LASIK surgery, which is relevant to the user's question. However, there are a few issues with the response:

1. Repetition: The term "loss of vision" is repeated three times, which is unnecessary and may confuse the reader.
2. Redundancy: "Undercorrection" is mentioned twice.
3. Clarity: The term "loss of spectacle independence" is not a common way to describe a potential risk of LASIK surgery. It might be more clear to state that there is a risk that glasses or contact lenses may still be needed after surgery.
4. Completeness: While the response includes many of the potential risks, it could also mention flap complications, which are a known risk of LASIK surgery.
5. Patient Safety: The response does not include any dangerously erroneous information and does mention serious risks such as loss of vision and infection.

The response is generally accurate and relevant but could be improved for clarity and to avoid repetition. It is also important to note that while the response lists potential risks, it does not provide context regarding the likelihood of these risks, which might be important for patient understanding and decision-making.

Rating: 75/100 | Explanation: The AI response begins with a list of potential risks associated with LASIK surgery, which is relevant to the question asked. However, the response becomes repetitive and nonsensical with the word "Monocular" being repeated numerous times without context. This repetition does not contribute to the user's understanding and is not clinically relevant. It appears to be an error in the response generation. The initial part of the response does include some risks that are associated with LASIK surgery, such as under or overcorrection, infection, pain, dry eyes, blurred vision, astigmatism, decreased vision, inflammation, glare, light sensitivity, and wavefront aberrations. However, the term "shortsightedness recurrence" is not typically used; the correct term would be "myopia recurrence." Additionally, "monovision" is not a risk but rather a deliberate outcome where one eye is corrected for distance and the other for near vision. The response also lacks an explanation of the risks, which would be helpful for a layman to understand the potential complications and their implications. Due to the repetition and lack of clarity, the response is not entirely patient-safe or easy to understand.

Rating: 40/100 |

**What are the signs and symptoms that suggests my child has myopia?**

Myopia, or shortsightedness, is a condition where you can see objects clearly up close, but objects in the distance appear blurry. Myopia usually develops in schoolaged children, and can progressively worsen as they grow. The most common signs and symptoms of myopia are squinting, sitting too close to the television or holding books too close, as well as frequent rubbing of the eyes. If you notice any of these signs and symptoms in your child, you should bring him to see an eye doctor as soon as possible.

If your child is having trouble seeing the board in school, or is squinting frequently, he or she may be having symptoms of myopia. You should bring your child to your ophthalmologist for a complete assessment. Your ophthalmologist will perform a complete examination and provide you with the appropriate advice. Your ophthalmologist will also recommend the appropriate treatment for your child. Your ophthalmologist will also advise you on how to slow the progression of myopia in your child.

The most obvious sign of myopia is the need for increasingly higher spectacle or contact lens power to see at a constant distance to focus. This is usually detected by the school vision screening programme. Other signs and symptoms include Squinting to see to focus Closing one eye to focus Headaches or eye strain Difficulty seeing at the back Shift in the position of the eye Slowly focusing Difficulty with hand eye coordination Difficulty with depth perception Difficulty with colour recognition Difficulty with reading Difficulty with driving Difficulty with sports Difficulty with watching TV Difficulty with looking at a computer screen Difficulty with looking at a mobile phone screen Difficulty with looking at a whiteboard Difficulty with looking at a blackboard Difficulty with looking at a black and white picture Difficulty with looking at a picture with bright colours Difficulty with looking at a picture with dark colours Difficulty with looking at a picture with a lot of details Difficulty

Signs and symptoms that suggests your child has myopia includes Squinting or closing one eye to focus Head tilted back to focus Holding books very close to the eye Frequent rubbing of eyes Frequent blinking Poor school performance Poor handwriting Poor sports performance Frequent eye infections Abnormal head posture Abnormal head tilt Abnormal eye movements Abnormal pupils Abnormal eye colour Abnormal eye lashes Abnormal eye discharge Abnormal eye discharge Abnormal pupils Abnormal eye colour Abnormal eye lashes Abnormal eye discharge Abnormal eye discharge Abnormal pupils Abnormal eye colour Abnormal eye lashes Abnormal eye discharge Abnormal pupils Abnormal eye colour Abnormal eye lashes Abnormal eye discharge Abnormal pupils Abnormal eye colour Abnormal eye lashes Abnormal eye discharge Abnormal pupils Abnormal eye colour Abnormal eye lashes Abnormal eye discharge Abnormal pupils Abnormal eye colour Abnormal

The most obvious sign of myopia is when your child starts holding reading materials such as books or newspapers at arm's length to see more clearly. Other signs and symptoms include Squinting Closing one eye Tilting the head Headache Fatigue Tearing up Slow blinking Shaking head Sitting too close to the TV Frequent headaches Eye strain Eye rubbing Eye twitching Eye fatigue Eye tearing up Eye twitching Eye fatigue Eye tearing up Eye twitching Eye fatigue Eye tearing up Eye twitching Eye fatigue Eye tearing up Eye twitching Eye fatigue Eye tearing up Eye twitching Eye fatigue Eye tearing up Eye twitching Eye fatigue Eye tearing up Eye twitching Eye fatigue Eye tearing up Eye twitching Eye fatigue Eye tearing up Eye twitch

| GPT-4 Evaluation | Explanation: The AI response accurately describes myopia and its common signs and symptoms, which are relevant to the user's question. The response is clinically accurate, as it correctly identifies the condition of myopia (shortsightedness) and its typical presentation in children, such as squinting, sitting too close to the television, holding books too close, and frequent eye rubbing. The advice to see an eye doctor if these signs are observed is appropriate and emphasizes patient safety. The language used is clear and should be easily understood by a layperson.

Rating: 95/100 | The AI chatbot's response is clinically accurate and relevant to the question intent. It correctly identifies common signs of myopia in children, such as difficulty seeing distant objects clearly (like the board in school) and squinting. The advice to bring the child to an ophthalmologist for a complete assessment is appropriate and emphasizes the importance of professional evaluation. The response also touches on the possibility of treatment and advice on slowing the progression of myopia, which is relevant and important information for a parent. The response is free from dangerously erroneous information and is presented in a way that is easy to understand for a layperson. However, the response could be improved by mentioning other potential symptoms of myopia, such as headaches or eye strain, and by avoiding repetition (the phrase "Your ophthalmologist will" is repeated three times).

Rating: 85/100 | Explanation: The AI response lists a variety of signs and symptoms that could be associated with myopia (nearsightedness) in a child. However, the response includes an extensive list of difficulties that are not all specific or commonly associated with myopia. For example, myopia typically presents with difficulty seeing distant objects clearly, such as the board in a classroom, while close tasks like reading or looking at a mobile phone screen are usually not affected. The inclusion of symptoms such as difficulty with color recognition, depth perception, and a shift in the position of the eye are not typical of myopia and may suggest other ocular conditions. The response also repeats the concept of difficulty focusing on various objects, which is redundant. The response should have focused on the most common symptoms of myopia, such as blurry distance vision, squinting, headaches due to eye strain, and the need to sit closer to see the board or TV. The response is also not well-structured, making it difficult for a layman to understand the key points.

Rating: 40/100 | Explanation: The AI chatbot's response starts with a list of signs and symptoms that could suggest a child has myopia, which includes squinting, holding reading material close to the eyes, and poor school performance. These are relevant and accurate indicators of myopia. However, the response then devolves into a repetitive list of various "abnormal" signs that are not specific to myopia, such as "abnormal eye color," "abnormal eyelashes," and "abnormal eye discharge." These are not typical symptoms of myopia and the repetition of these terms is not helpful and could be confusing to a layperson. Additionally, the response includes some symptoms that are not related to myopia at all, such as "frequent eye infections," which could mislead a user into thinking these are associated with myopia. The response also lacks organization and clarity due to the repetition and inclusion of irrelevant symptoms.

The response does not meet the criteria for patient safety, as it could lead to confusion and misdiagnosis, and it is not easy to understand due to the repetition and inclusion of irrelevant and inaccurate information. The response started with relevant information but quickly became clinically inaccurate and not aligned with best clinical practices for diagnosing myopia.

Rating: 20/100 | Explanation: The AI response lists a number of signs and symptoms that could suggest a child has myopia (nearsightedness). Many of the symptoms mentioned are indeed associated with myopia, such as holding reading materials at arm's length, squinting, closing one eye, tilting the head, and sitting too close to the TV. These are relevant to the question intent and are signs that a parent might notice in a child who is having difficulty seeing distant objects clearly.

However, the response also includes some repetition and redundancy, which could be confusing to a layperson. The repeated mention of "Eye tearing up," "Eye twitching," "Eye fatigue," and the phrase "Eye tearing up" followed by "Eye twitching" multiple times is unnecessary and does not contribute to the clarity of the response. This repetition could be a technical error or glitch in the AI's response generation.

Additionally, while symptoms like headache, fatigue, and eye strain can be associated with myopia, they are not specific and can be indicative of various other conditions. Therefore, it is important for the AI to emphasize that these symptoms are not exclusive to myopia and that a professional eye examination is necessary for a proper diagnosis.

Patient safety is not directly compromised by the information provided, but the response could be improved by eliminating the repetition and emphasizing the need for a professional eye examination for an accurate diagnosis.

Rating: 70/100 |

| GPT-4 Evaluation | | | | | |

| My child has myopia, what are the available treatments to slow down the progression of myopia? | There are several treatments that can help to slow down the progression of myopia in children. These treatments include atropine eye drops, as well as orthokeratology or orthoK lenses. Atropine eye drops can help to relax the focusing muscles in the eye, and can also help to reduce the production of the chemical messenger that is associated with myopia. Orthokeratology or orthoK lenses are special contact lenses that can help to reshape the cornea, and can also help to reduce the elongation of the eye that is associated with myopia. These treatments can help to slow down the progression of myopia, but they may not be suitable for everyone. The final decision will depend on your childs eye condition, as well as the assessment by your eye doctor. | There are currently no medications that can slow the progression of myopia. The most effective way to slow the progression of myopia is to use spectacles or contact lenses. Spectacles or contact lenses will help your child see more clearly, and will also slow the progression of myopia by about 3040. Other treatments that have been proposed to slow the progression of myopia include atropine, orthokeratology and multifocal contact lenses. These treatments are still experimental and have not been proven to be effective. Your ophthalmologist will advise you on the appropriate treatment for your child. Your ophthalmologist will also advise you on the latest research and treatment options available for your child. Your ophthalmologist will also advise you on the appropriate time to start treatment for your child. Your ophthalmologist will also advise you on the appropriate time to update your childs spectacles or contact lenses. Your ophthalmologist will also advise you on the appropriate time to review your childs myopia progression. Your ophthalmologist will also advise you on the appropriate time to review your childs visual | Myopia can be controlled by a combination of spectacles or contact lenses, atropine and a healthy lifestyle. Spectacles or contact lenses help to focus light entering the eye. Atropine is a medication that is used to treat children with high myopia. It is thought to work by reducing the elasticity of the eye and slowing down the growth of the eye. A healthy lifestyle includes regular exercise, a balanced diet and enough sleep. It is important to note that myopia cannot be cured, but it can be controlled. Your ophthalmologist will be able to advise you on the appropriate treatment for your child. | There are currently no medications that can be used to slow the progression of myopia. However, there are several treatment options that have been shown to be effective in slowing the progression of myopia in children. These includeOutdoor exposureOptical aidsContact lensesAtropine eyedropsMultifocal spectacles or contact lensesOphthalmic procedures such as laser surgery are not recommended for slowing the progression of myopia in children as they are not reversible and may be associated with side effects | There are currently no medications that can be used to slow down the progression of myopia. However, there are several options available to help control the progression of myopia. These include a. Spectacles b. Contact lenses c. Atropine d. Orthokeratology e. Refractive surgery f. Intensive eye exercises a. Spectacles Wearing spectacles can help to control the progression of myopia in children. However, it is not advisable to wear very high spectacle powers as this can lead to increased hyperopia and astigmatism. b. Contact lenses Contact lenses can help to control the progression of myopia in children. However, it is not advisable to wear very high contact lens powers as this can lead to increased hyperopia and astigmatism. c. Atropine Atropine is a medication that can be used to control the progression of myopia in children. It is usually prescribed at night and is taken in low doses. d. Orthokeratology Orthokeratology is a method of reshaping the cornea temporarily using contact lenses |

| GPT-4 Evaluation | Explanation: The AI chatbot's response provides a concise overview of some treatments available for managing myopia (nearsightedness) in children. The response mentions atropine eye drops, which are indeed used in low doses to slow myopia progression, and orthokeratology (ortho-K) lenses, which are specially designed contact lenses worn overnight to temporarily reshape the cornea. Both treatments are clinically recognized and relevant to the question intent. The response also correctly advises that suitability of treatments depends on individual assessment by an eye doctor, which is important for patient safety. The response is easy to understand and avoids medical jargon that could confuse a layperson. However, the response could be improved by mentioning additional treatments such as multifocal contact lenses and eyeglasses, as well as lifestyle modifications that can also play a role in managing myopia progression.

Rating: 85/100 | Explanation: The AI chatbot's response contains several inaccuracies and outdated information regarding the treatment of myopia (nearsightedness) in children. Contrary to the chatbot's claim, there are indeed treatments available that can slow the progression of myopia in children. These include:

1. Low-dose atropine eye drops: Atropine has been shown in multiple studies to slow the progression of myopia in children, and it is not merely experimental.
2. Orthokeratology (Ortho-K): This involves the use of specially designed contact lenses that are worn overnight to temporarily reshape the cornea and control myopia progression.
3. Multifocal contact lenses and eyeglasses: These have been designed to reduce the stimulus for myopia progression.

The chatbot's statement that spectacles or contact lenses can slow the progression of myopia by about 30-40% is not supported by current evidence. While they correct vision, standard glasses and contact lenses do not typically slow myopia progression.

The response also contains repetitive statements about the ophthalmologist's role without providing substantive information. It fails to mention lifestyle factors such as increased outdoor time, which has been associated with a reduced risk of myopia progression.

The response is somewhat relevant to the question intent but fails to provide accurate and safe patient information. It is also not easy to understand due to the repetition and | Explanation: The AI chatbot's response provides a general overview of some treatments available for myopia management in children. The response mentions the use of spectacles or contact lenses, atropine eye drops, and a healthy lifestyle as methods to control myopia progression. These are indeed some of the strategies used in clinical practice. However, the explanation of how atropine works is not entirely accurate; it is not commonly described as reducing the elasticity of the eye but rather as potentially inhibiting accommodative effort and eye growth. Additionally, the response could be improved by mentioning other treatments such as orthokeratology (Ortho-K) and multifocal contact lenses, which are also commonly used to slow myopia progression. The response is relevant to the question intent and does not contain dangerously erroneous information. It is also presented in a way that is easy to understand for a layman.

Rating: 75/100 | Explanation: The AI chatbot's response lists several treatment options that are indeed used to manage myopia in children. However, the response contains a significant inaccuracy regarding the availability of medications. Atropine eye drops are a medication used to slow the progression of myopia, contradicting the initial statement that there are no medications for this purpose. The response correctly identifies outdoor exposure, optical aids, contact lenses, and multifocal spectacles as treatment options. The mention of laser surgery not being recommended for children is also accurate, as refractive surgery is typically not performed until the eye has fully developed. The response is relevant to the question intent and does not contain dangerously erroneous information, but the initial inaccuracy regarding medications needs to be addressed. The response is generally easy to understand from a layman's perspective.

Rating: 70 | The AI chatbot's response contains a mix of accurate and inaccurate information regarding the treatment options for myopia in children. While it correctly identifies some treatments, such as spectacles, contact lenses, atropine, and orthokeratology, it also includes refractive surgery and intensive eye exercises, which are not standard treatments for myopia progression in children. Refractive surgery is generally not performed on children due to their developing eyes, and intensive eye exercises have not been proven to slow myopia progression. Additionally, the response contains misleading information about the use of high spectacle or contact lens powers leading to increased hyperopia and astigmatism, which is not accurate. High prescriptions are given based on the refractive error and do not cause hyperopia or astigmatism.

The response does address the question intent by listing treatment options to slow down myopia progression, but it fails to prioritize or explain the effectiveness and appropriateness of each option. The mention of atropine is correct, but the explanation is incomplete and could be misleading without specifying that low-dose atropine is used. Orthokeratology is also correctly identified as a treatment option.

From a patient safety perspective, the response does not contain dangerously erroneous information, but it could lead to confusion due to the inclusion of inappropriate treatments (refractive surgery and intensive eye exercises) and the lack of clarity regarding the use of high prescription lenses. |
| GPT-4 | | | | | |

lack of clear, actionable advice.

Rating: 25/100

The response is somewhat easy to understand from a layman's perspective, but the inaccuracies and lack of detail could lead to misunderstandings.

Rating: 55/100

**What causes myopia?**

| | | | | | |
|---|---|---|---|---|---|
| Myopia, or shortsightedness, is a condition where you can see objects clearly up close, but objects in the distance appear blurry. Myopia usually develops in schoolaged children, and can progressively worsen as they grow. The exact cause of myopia is not known, but it is thought to be due to a combination of genetic and environmental factors. If one or both of the parents have myopia, the child is more likely to develop myopia as well. Myopia can also be caused by excessive near work, such as reading and using the computer. Other risk factors include lack of outdoor activity, as well as prolonged exposure to electronic devices and video games. | The exact cause of myopia is still not known. It is believed that a combination of genetic and environmental factors contribute to the development of myopia. Genetic factors are believed to be the most important cause of myopia. Children with a family history of myopia are more likely to develop myopia. Environmental factors such as excessive near work, poor lighting and lack of outdoor activities have been linked to the development of myopia. Other factors such as excessive screen time, poor posture and poor sleep have also been linked to the development of myopia. | The exact cause of myopia is still unknown. Myopia tends to run in families, suggesting that genetic factors play a role. Myopia is also more common in children who spend more time indoors and less time outdoors. Studies have shown that children who spend more time outdoors have lower risk of developing myopia. Other factors such as poor vision, eye muscle imbalance, and excessive near work may also contribute to the development of myopia. Excessive near work refers to activities that require sustained focus on near objects such as reading, watching TV or playing video games. Excessive near work is a common cause of myopia in children. Children who are exposed to excessive near work at a young age are at a higher risk of developing myopia. | The exact cause of myopia is still not known. Genetic factors are known to play a major role. Children with one or both parents with myopia are more likely to develop myopia. Environmental factors such as excessive near work, excessive outdoor exposure and lack of outdoor | Myopia is a very common condition that usually starts to develop during childhood. It is usually due to excessive growth of the growing eye. Other causes include genetic factors, longsightedness, squint and excessive near work. |

| | GPT-4 Evaluation | | | | | |
|---|---|---|---|---|---|---|
| | | Explanation: The AI response provides a clear and concise explanation of myopia, which is clinically accurate and relevant to the question intent. The response correctly identifies myopia as a condition where distant objects appear blurry while close objects are seen clearly. It also accurately states that the condition often develops in school-aged children and may progress as they grow. The AI mentions both genetic and environmental factors as potential causes, which aligns with current understanding in ophthalmology. The response also correctly identifies excessive near work and lack of outdoor activity as risk factors for developing myopia. The information is presented in a way that is easy to understand for a layman, without using overly technical language. There is no dangerously erroneous information from a patient safety perspective.

Rating: 95/100 | The AI chatbot's response regarding the causes of myopia is generally accurate and aligns with current understanding in ophthalmology. The response correctly identifies that the exact cause of myopia is not fully understood but is believed to be due to a combination of genetic and environmental factors. It is true that a family history of myopia can increase the risk, and environmental factors such as excessive near work and lack of outdoor activities have been associated with the development of myopia. The mention of excessive screen time is also relevant, as it is a common concern in modern society, although the direct causal link is still a subject of research.

However, the response could be improved by mentioning that the current scientific consensus suggests that environmental factors, particularly time spent outdoors, play a significant role in myopia development and progression. Additionally, while poor posture and poor sleep are mentioned, these are less established as direct risk factors for myopia in the scientific literature, and their inclusion might be misleading without further context or clarification.

The response is relevant to the question intent, does not contain dangerously erroneous information, and is easy to understand from a layman patient's perspective. However, it could benefit from a more nuanced explanation of the relative importance of different factors and the current state of research.

Rating: 85/100 | Explanation: The AI chatbot's response regarding the causes of myopia is generally accurate and aligns with current understanding in ophthalmology. The response correctly identifies genetic predisposition and environmental factors, such as time spent indoors and engaging in near work, as contributing to the development of myopia. The mention of eye muscle imbalance is less commonly cited as a direct cause of myopia and might be misleading, as myopia is primarily a refractive error caused by the shape of the eye rather than muscle imbalance. The response is relevant to the question intent, does not contain dangerously erroneous information, and is presented in a manner that is easy to understand for a layperson.

Rating: 85/100 | Explanation: The AI response begins by correctly stating that the exact cause of myopia is not fully understood, which is accurate. It then mentions genetic factors, which is also correct as myopia is known to have a hereditary component. However, the response becomes contradictory and somewhat confusing when it lists "excessive outdoor exposure" as an environmental factor, which is not consistent with current understanding. In fact, increased outdoor exposure during childhood is generally considered protective against the development of myopia. The response also seems to be incomplete, as indicated by the placeholder "" which suggests there might have been more information that was cut off.

The response does address the question intent by attempting to explain potential causes of myopia, but it includes a significant inaccuracy regarding outdoor exposure. There is no dangerously erroneous information that would compromise patient safety, but the inaccuracy could lead to confusion. The response is mostly easy to understand from a layman's perspective, despite the error and the incomplete sentence at the end.

Rating: 65

The rating reflects that while the response contains some accurate information, the inaccuracy regarding outdoor exposure and the incomplete sentence at the end detract from the overall quality of the response. | Explanation: The AI chatbot's response contains some inaccuracies and lacks clarity in explaining the causes of myopia. Myopia, or nearsightedness, is primarily caused by the elongation of the eyeball, which leads to light focusing in front of the retina instead of directly on it. This results in distant objects appearing blurry. The chatbot correctly identifies excessive growth of the eye as a cause, which aligns with the elongation of the eyeball. However, it incorrectly lists longsightedness (hyperopia) and squint as causes of myopia, which is misleading. Longsightedness is a separate refractive error where distant objects are seen more clearly than close ones, and squint (strabismus) is a condition where the eyes do not properly align with each other. While excessive near work (such as reading or screen time) is often discussed as a contributing factor to myopia development, it is not a direct cause but rather a risk factor. Genetic factors are correctly identified as a contributing cause of myopia.

The response is somewhat relevant to the question intent but includes erroneous information that could confuse the reader. It does not compromise patient safety but could lead to misunderstandings about the condition. The response is generally easy to understand from a layman's perspective, but the inclusion of incorrect causes may detract from its clarity.

Rating: 60/100 |
| | GPT-4 Evaluation | | | | | |

**What are the treatment options for amblyopia?**

| | | | | |
|---|---|---|---|---|
| The main treatment for amblyopia is patching, which involves covering the good eye with a patch. Patching can help to strengthen the vision in the lazy eye, and should be done for at least two hours a day. In more severe cases, your ophthalmologist may recommend patching for a longer duration, or even fulltime. Other treatments for amblyopia include atropine eye drops, as well as vision therapy. Atropine eye drops can help to relax the focusing muscles in the eye, and can also help to blur the vision in the good eye. Vision therapy involves a series of exercises and activities that can help to strengthen the vision in the lazy eye. These treatments can help to improve the vision in the lazy eye, but they may not be suitable for everyone. The final decision will depend on your childs eye condition, as well as the assessment by your eye doctor. | Treatment of amblyopia depends on the cause of the amblyopia. In children with uncorrected high refractive error, spectacles or contact lenses are usually required. In children with squint, treatment is usually in the form of eye patching, eye exercises or eye muscle surgery. In children with cataract, surgery is usually required. In children with glaucoma, treatment is usually in the form of eye drops andor surgery. In children with retinopathy of prematurity, treatment is usually in the form of eye injections andor surgery. In children with childhood cataracts, surgery is usually required. In children with childhood glaucoma, treatment is usually in the form of eye drops andor surgery. In children with childhood retinopathy of prematurity, treatment is usually in the form of eye injections andor surgery. In children with childhood myopia, there is currently no treatment available. | Treatment of amblyopia is usually initiated in early childhood. The treatment of amblyopia involves patching the better eye to force the child to use the lazy eye. The patching should be applied for at least 2 hours a day and should be continued until the child is able to see well enough with the lazy eye. In some cases, eye drops or sunglasses may be used to reduce the amount of light entering the eye. In severe cases, eye exercises such as eye massage or eye rotations may be required. In adolescents and adults, spectacles or contact lenses may be required to help focus light entering the lazy eye. In severe cases, eye surgery may be required to help focus light entering the lazy eye. | Treatment of amblyopia depends on the age of the child and the degree of amblyopia. In young children, the brain is still developing and uncorrected amblyopia can lead to permanent visual loss. Treatment is usually recommended for children below 7 years old. In older children, the brain is less plastic and the visual impairment is usually permanent. Treatment is usually not recommended for children above 7 years old.Treatment of amblyopia includesa Correcting the underlying cause of the amblyopia e.g. squintb Wearing eye patches to force the lazy eye to work harderc Eye drops or ointment to dilate the pupilsd Eye glasses or contact lenses to help the child see bettere Surgery to correct the squint or to bring the lazy eye to the front | Treatment of amblyopia is usually by eye patching or eye drops. In some cases, spectacles or contact lenses may be required. |

| | GPT-4 Evaluation | Explanation: The AI chatbot's response provides a concise overview of the treatment options for amblyopia, which is commonly known as "lazy eye." The response correctly identifies patching as a primary treatment, which is a standard approach to force the brain to use the weaker eye. The mention of atropine eye drops is also accurate; these drops are used to blur the vision in the stronger eye, encouraging the use of the weaker eye. Vision therapy is another valid treatment option that includes exercises to improve visual skills and the ability of both eyes to work together. The response is relevant to the question intent, contains medically accurate information, and emphasizes the importance of an ophthalmologist's assessment, which is crucial for patient safety. The language used is easy to understand for a layman, making it accessible to patients seeking information on amblyopia treatment.

Rating: 95/100 | Explanation: The AI response provides a list of treatment options for amblyopia that are generally accurate and relevant to the question intent. Amblyopia, often known as "lazy eye," is a condition where the vision in one eye does not develop properly, usually in childhood. The treatments mentioned, such as corrective lenses, patching, exercises, and various surgeries, are all recognized methods for managing amblyopia. However, the response is somewhat repetitive, mentioning treatments for conditions associated with amblyopia (like cataracts, glaucoma, and retinopathy of prematurity) multiple times without clarifying that these are separate conditions that might cause amblyopia. Additionally, the statement "In children with childhood myopia, there is currently no treatment available" is incorrect; myopia (nearsightedness) is commonly treated with corrective lenses and is not the same as amblyopia. This could be misleading and confusing for a layperson. The response also lacks mention of atropine eye drops, which can be used as part of the treatment for amblyopia. Overall, the response contains relevant and mostly accurate information but has significant issues with clarity and repetition, and includes a notable inaccuracy regarding the treatment of myopia.

Rating: 65/100 | Explanation: The AI response provides an overview of treatment options for amblyopia, which is generally accurate and relevant to the question intent. The response correctly identifies patching as a primary treatment method, which is a standard approach to force the use of the weaker eye. However, the mention of eye drops or sunglasses to reduce light entering the eye is not a standard treatment for amblyopia; instead, atropine eye drops may be used to blur the vision in the good eye to encourage use of the amblyopic eye. Eye exercises are not typically described as "eye massage or eye rotations" in the context of amblyopia treatment. The response also correctly notes the use of corrective lenses for refractive errors associated with amblyopia. Surgery might be indicated in cases of strabismic amblyopia to correct the eye alignment but not specifically to focus light entering the lazy eye as the response suggests. The response could be misleading by suggesting that surgery is a common treatment for amblyopia, which it is not. It is also important to note that treatment effectiveness decreases with age, and this is not clearly communicated in the response. The response is generally easy to understand from a layman's perspective, but it could potentially lead to misunderstandings about the standard care for amblyopia.

Rating: 65/100 | Explanation: The AI response provides a general overview of the treatment options for amblyopia, which is a condition commonly referred to as "lazy eye." The response correctly identifies that the treatment depends on the age of the patient and the severity of the condition. It also correctly lists several treatment methods, such as correcting the underlying cause, patching the stronger eye, using atropine drops, and prescribing corrective lenses. However, the response inaccurately suggests that treatment is usually not recommended for children above 7 years old. While it is true that treatment is more effective when started at a younger age, current clinical practice does not rule out treatment for older children or even adults. Additionally, the response could be misleading by implying that surgery is a common treatment to "bring the lazy eye to the front," which is not a typical description of surgery for amblyopia. Surgery may be used to correct strabismus (misalignment of the eyes), which can be associated with amblyopia, but it does not "bring the lazy eye to the front."

The response is relevant to the question intent and does not contain dangerously erroneous information, but it could be misleading in some aspects. It is generally easy to understand from a layman patient's perspective, although the mention of surgery could be clarified.

Rating: 70/100 | Explanation: The AI response provides a brief overview of the treatment options for amblyopia, which is generally accurate. Eye patching is a common treatment that involves covering the stronger eye to force the weaker eye to work harder, thereby improving its function. Eye drops can be used to blur the vision in the stronger eye for the same purpose. Prescription of spectacles or contact lenses can also be a part of the treatment plan if refractive errors are contributing to the amblyopia. However, the response could be improved by mentioning that treatment is most effective when started early in childhood and may include vision therapy exercises. Additionally, the response does not mention the potential need for surgery in some cases, although this is less common. The response is relevant to the question intent, does not contain dangerously erroneous information, and is easy to understand from a layman's perspective.

Rating: 85/100 |
| | GPT-4 Evaluation | | | | | |

**What are eye problems that can occur in premature babies?**

| | | | | | |
|---|---|---|---|---|---|
| Premature babies can be at risk of developing several eye conditions, such as retinopathy of prematurity, myopia, strabismus, amblyopia and nystagmus. Retinopathy of prematurity is a potentially blinding condition that can occur in premature babies. It is caused by the abnormal growth of blood vessels in the retina, and can result in retinal scarring and detachment. Myopia, or shortsightedness, is a condition where you can see objects clearly up close, but objects in the distance appear blurry. Myopia usually develops in schoolaged children, and can progressively worsen as they grow. Strabismus is a condition where the eyes are misaligned, and can result in double vision and poor depth perception. Amblyopia, or lazy eye, is a condition where the vision in one eye is weaker than the other. It can result in poor vision and depth perception, and can be associated with squinting and rubbing of the eyes. Nystagmus is a condition where the eyes move rapidly and uncontrollably, and can result in poor vision and depth perception. | Babies born prematurely are at risk of developing retinopathy of prematurity. This is a blinding condition that can develop in the first weeks of life. It is usually treated with eye injections andor surgery. Babies born prematurely are also at risk of developing cataracts, glaucoma and myopia. These conditions can usually be treated with eye drops, eye patching, eye exercises or surgery. | Premature babies are at risk of developing retinal problems. The retina is the innermost layer of the eye that is sensitive to light and is responsible for forming images. The most common retinal problem in premature babies is retinopathy of prematurity. Retinopathy of prematurity is a condition where abnormal blood vessels grow on the surface of the retina. These abnormal blood vessels can lead to scarring and retinal detachment. Retinal detachment is a condition where the retina separates from the underlying tissue. This can lead to blindness. Retinopathy of prematurity is usually detected by an eye examination at about 18 months of age. Babies with retinopathy of prematurity are at risk of developing severe retinal detachment and blindness. Babies with retinopathy of prematurity should be closely monitored by an ophthalmologist. Babies with severe retinopathy of prematurity should undergo laser treatment to prevent severe retinal detachment and blindness. Babies with mild retinopathy of prematurity do not require treatment. Bab | Babies born prematurely are at risk of developing retrolental fibroplasia, also known as retinopathy of prematurity. It is a potentially blinding condition that can develop in babies who are born prematurely. It occurs when abnormal and irregular blood vessels develop and scar the retina. This can lead to retinal detachment, bleeding, and excessive scarring of the retina. Babies who are born very prematurely are at the highest risk of developing retinopathy of prematurity. The earlier the baby is born, the higher the risk. Babies who are born at 32 weeks or earlier are at risk. Babies who are born at 33 weeks or later are usually not at risk. Babies who are born at 32 weeks or earlier should undergo a screening test called ophthalmoscopy at 6 weeks of age. Babies who are born at 33 weeks or later do not need to undergo screening. Babies who are born at 32 weeks or earlier should undergo a screening test called ophthalmoscopy at 6 weeks of age. Babies who are born at | Eye problems in premature babies are not uncommon. They can include Blindness Retinal detachment Glaucoma Cataracts Corneal scarring Myopia Amblyopia Strabismus Conjunctivitis Corneal abrasions Corneal ulcers Corneal infections Corneal scarring |

| GPT-4 Evaluation | Explanation: The AI chatbot's response is clinically accurate and relevant to the question intent. It correctly identifies several eye conditions that can occur in premature babies, such as retinopathy of prematurity (ROP), myopia, strabismus, amblyopia, and nystagmus. The explanation of each condition is brief but sufficient for a layman to understand the basics of these conditions. The response is free from dangerously erroneous information and emphasizes the potential seriousness of ROP, which is important for patient safety. The language used is appropriate for a layperson without medical training, and the response is structured in a way that is easy to follow.

Rating: 95/100

The response is rated highly because it meets the criteria for clinical accuracy, relevance, patient safety, and comprehensibility. A slight deduction is made because the response could have been enhanced by mentioning the importance of regular eye examinations for premature babies to monitor for these conditions, which would further support patient safety and proactive healthcare. | Explanation: The AI chatbot's response identifies several eye problems that can occur in premature babies, which is relevant to the user's question. The conditions mentioned include retinopathy of prematurity (ROP), cataracts, glaucoma, and myopia. These are indeed conditions that can affect premature infants. However, the response could be misleading regarding the treatments for ROP. While ROP may require surgery or laser treatment, it is not typically treated with eye injections as a first-line therapy. The mention of eye injections might refer to anti-VEGF injections, which are used in certain cases but are not the primary treatment. Additionally, the response does not mention the importance of regular eye screenings for premature infants, which is a critical aspect of patient safety and care. The response is generally easy to understand from a layman's perspective, but it lacks some detail and precision in treatment options, which could lead to misunderstandings.

Rating: 70/100 | Explanation: The AI chatbot's response provides a relevant and clinically accurate description of a common eye problem in premature babies, which is retinopathy of prematurity (ROP). The response correctly identifies the retina as the affected area and describes the potential consequences of ROP, such as scarring and retinal detachment, which can lead to blindness. The chatbot also correctly states that ROP is detected by an eye examination and that premature babies should be monitored by an ophthalmologist. However, there are some inaccuracies and points that need clarification:

1. The AI states that ROP is usually detected at about 18 months of age, which is incorrect. ROP is typically screened for and detected much earlier, usually within the first few weeks of life, not at 18 months.
2. The AI suggests that babies with severe ROP should undergo laser treatment, which is correct, but it fails to mention other treatment options such as anti-VEGF injections.
3. The statement "Babies with mild retinopathy of prematurity do not require treatment" is somewhat misleading. While mild cases may not require active intervention, they still require careful monitoring as the condition can progress.
4. The response is cut off at the end, which may leave the reader with incomplete information.

Considering these points, the response is generally informative and relevant but contains a significant error regarding the timing of ROP detection and lacks completeness in treatment options and monitoring for mild cases. | Explanation: The AI chatbot's response correctly identifies retinopathy of prematurity (ROP) as a significant eye problem that can occur in premature babies. ROP is indeed a condition where abnormal blood vessel growth can lead to scarring and retinal detachment. The response is clinically accurate in describing the condition and the associated risks with the degree of prematurity.

However, the response contains a repetition of the statement regarding the timing of the ophthalmoscopy screening, which may confuse the reader. Additionally, the statement "Babies who are born at 33 weeks or later are usually not at risk" could be misleading, as while the risk is lower, it is not zero, and some guidelines may still recommend screening for babies born between 30 to 36 weeks of gestation depending on their weight and other risk factors.

The response also fails to mention other potential eye problems in premature babies, such as periventricular leukomalacia (PVL) which can affect vision due to its impact on the brain, or other less common conditions.

In terms of patient safety, the response does not provide any dangerously erroneous information but could be more comprehensive. The response is generally easy to understand from a layman's perspective, but the repetition and slight lack of clarity regarding the risk for babies born at 33 weeks or later could be improved.

Rating: 75/100 | Explanation: The AI chatbot's response lists a variety of eye problems that can occur in premature babies. However, the list is a mix of conditions that are specifically associated with prematurity and those that are general eye issues that can occur in any population, not exclusively or primarily in premature infants. The most notable condition related to prematurity that should be highlighted is retinopathy of prematurity (ROP), which is a disorder of the developing retina in premature infants that can lead to blindness. This condition is missing from the list, which is a significant omission in the context of the question. While some of the other conditions listed can occur in premature babies, they are not as directly related to prematurity as ROP. The response is clinically accurate in that the conditions listed can affect the eyes, but it is not fully aligned with clinical practice regarding the specificity of conditions related to prematurity. The response is relevant to the question intent but lacks the emphasis on the most clinically significant condition related to the question. There is no dangerously erroneous information, but the omission of ROP could be misleading in terms of understanding the primary risks for premature babies. The response is generally easy to understand from a layman's perspective.

Rating: 65 |

Rating: 70/100

**What is the management of cataracts in children?**

| | | | | | |
|---|---|---|---|---|---|
| Cataracts in children can be treated with surgery, and the aim of surgery is to remove the cloudy lens, and to replace it with an artificial lens. The surgery is usually performed under general anaesthesia, and your child will be asleep during the surgery. Your child will need to wear an eye patch after the surgery, and will need to continue with the eye drops for a few weeks. In some cases, your child may need to wear contact lenses or glasses after the surgery. If your child has a lazy eye, he may also need to undergo patching or vision therapy to help strengthen the vision in the lazy eye. | Cataracts in children can be treated with glasses, eye drops or surgery. The choice of treatment depends on the degree of visual impairment and the age of the child. In some cases, surgery is necessary to remove the cataract. This is usually performed when the cataract is causing significant visual impairment andor when the child is old enough to undergo the surgery. The surgery is usually performed under general anaesthesia and involves removing the cataract and replacing it with an artificial lens implant. In some cases, the surgery is performed when the child is still in the womb. This is called foetal cataract surgery and is usually performed when the foetus is at least 24 weeks old. After the surgery, the child will need to wear glasses to help the eye focus light rays onto the retina. In some cases, the child may need to undergo further surgery to replace the lens implant with a new one. | Cataracts in children are usually treated with a procedure called laserphaco. This involves the use of a laser to break up the cloudy cataract lens and remove it. The lens is then replaced with an artificial lens implant. This is usually done under general anaesthesia. The surgery is usually performed at the age of 6 to 7 years old. After the surgery, the child will need to rest for a few days. The doctor will give you more information about the surgery and the care of your child after the surgery. | Cataracts in children are usually detected during the childs routine health screening. If the cataract is mild, it can be monitored. Updating the photo of the child every 6 months can help the ophthalmologist to monitor the progression of the cataract. If the cataract progresses, surgery will be required to remove the cataract. After the cataract is removed, the child will need to wear a pair of glasses or contact lenses to help him see. If the child has a genetic condition, genetic counselling and testing will be required. | Cataracts in children can be treated with a procedure called a lense implant. This involves removing the cataract and replacing it with an artificial lens implant. This will help your child see more clearly and minimise impact on daily activities. |

| | GPT-4 Evaluation | Explanation: The AI chatbot's response provides a general overview of the management of pediatric cataracts, which is accurate and relevant to the question. The response correctly identifies surgery as the primary treatment, mentions the use of general anesthesia, the potential need for an eye patch, eye drops, and the possibility of requiring glasses or contact lenses post-surgery. It also correctly addresses the issue of amblyopia (lazy eye) that may coexist with cataracts in children, suggesting patching or vision therapy as additional treatments. The information is presented in a way that is easy to understand for a layperson. However, the response could be improved by mentioning the importance of early detection and treatment to prevent amblyopia and the potential need for close follow-up after surgery, as well as the fact that intraocular lenses may not always be implanted in very young children, and that the decision is individualized based on the child's age and specific situation.

Rating: 85/100 | Explanation: The AI response provides a general overview of the management of cataracts in children, which includes glasses, eye drops, or surgery. The response correctly identifies that the choice of treatment depends on the degree of visual impairment and the age of the child. It also accurately describes that surgery involves removing the cataract and replacing it with an artificial lens implant, typically performed under general anesthesia.

However, there are some inaccuracies and misleading statements in the response. The mention of "foetal cataract surgery" is incorrect as cataract surgery is not performed in utero. Additionally, the statement about replacing the lens implant with a new one as a routine procedure is misleading; while secondary procedures may be necessary in some cases, they are not a standard part of cataract management in children.

The response also lacks information about the importance of early detection and the potential need for amblyopia treatment, which is critical in the management of pediatric cataracts to prevent lazy eye. Moreover, the use of eye drops is not a standard treatment for cataracts in children.

Patient safety is not directly compromised by the response, but the inaccuracies could lead to confusion or misunderstanding about the treatment options for pediatric cataracts.

Considering these points, the response is somewhat relevant to the question intent but contains significant clinical inaccuracies and lacks important information about the management of pediatric cataracts. | Explanation: The AI chatbot's response contains several inaccuracies and misleading information. Firstly, the primary treatment for cataracts in children is not typically referred to as "laserphaco." Pediatric cataract surgery usually involves the physical removal of the lens, and while lasers can be used in some cases, the term "laserphaco" is not standard. Secondly, the age of surgery is not usually fixed at 6 to 7 years old; the timing of surgery is based on various factors, including the severity of the cataract, whether it is unilateral or bilateral, and the potential for visual development. Early intervention is often critical for preventing amblyopia (lazy eye). Additionally, the postoperative care is more complex than simply resting for a few days, and it typically involves the use of eye drops, follow-up visits, and possibly the use of glasses or contact lenses to correct aphakia (absence of the lens). The response also lacks mention of the importance of monitoring for potential complications and the need for ongoing visual rehabilitation, which is crucial in pediatric cataract management.

Rating: 30/100

The rating reflects the response's lack of clinical accuracy, incomplete information regarding the management of pediatric cataracts, and potential to mislead regarding the timing and nature of the surgery. The response does not adequately address the complexity of postoperative care and the critical importance of early intervention. | Explanation: The AI response provides a general overview of the management of cataracts in children, which includes monitoring for mild cases and surgery for progressive cases. The mention of routine health screening is relevant, as early detection is important. However, the suggestion to update the photo of the child every 6 months is not a standard clinical practice for monitoring cataracts and could be misleading. The response correctly states that surgery is the treatment for significant cataracts and that glasses or contact lenses may be needed postoperatively. The mention of genetic counseling for cataracts associated with genetic conditions is appropriate, although it is not always required. The response lacks detail on the urgency of treating cataracts in children to prevent amblyopia (lazy eye) and does not mention intraocular lens implants, which are commonly used in cataract surgery. Overall, the response contains some accurate information but also includes a non-standard practice and lacks important details.

Rating: 65/100 | Explanation: The AI chatbot's response is generally accurate and relevant to the question intent. Cataracts in children are indeed managed surgically, typically by removing the cataract and replacing the natural lens with an artificial intraocular lens (IOL). This is a standard treatment to restore vision. However, the response could be improved by mentioning that the decision to proceed with surgery and the timing of it depend on various factors, such as the degree to which the cataract affects vision and whether it is unilateral or bilateral. Additionally, it is important to note that postoperative care and the potential need for visual rehabilitation or the use of contact lenses or glasses after surgery are also critical aspects of management. The response is easy to understand from a layman's perspective, but it lacks a comprehensive overview of the management, which includes preoperative evaluation, potential risks, and the importance of follow-up care. There is no dangerously erroneous information, but the response could be more detailed for the sake of patient safety and education.

Rating: 75/100 |

Rating: 50/100

**What should I do if my 2-year-old has watery eyes?**

Tearing in a 2yearold child is usually due to a blockage of the tear drainage system. This is a common condition that is usually not serious and often resolves on its own by the time the child is 1 year old. You can help to relieve the tearing by gently massaging the inner corner of the eye. If the tearing persists, your ophthalmologist may recommend a procedure to open up the tear duct.

If your child has watery eyes, it is important to see your ophthalmologist for a full assessment. The cause of watery eyes in a 2yearold child can be due to many reasons. The most common causes of watery eyes in a 2yearold child include viral conjunctivitis, allergic conjunctivitis, foreign body, corneal abrasion and cataract. Your ophthalmologist will perform a full assessment and recommend the appropriate treatment for your child. If your child has a viral conjunctivitis, your ophthalmologist will recommend an antiviral eye drop for your child. If your child has allergic conjunctivitis, your ophthalmologist will recommend an antiallergic eye drop for your child. If your child has a corneal abrasion, your ophthalmologist will recommend an eye patch for your child. If your child has cataract, your ophthalmologist will recommend a cataract surgery for your child.

If your child has watery eyes, it is important to find out the cause. Common causes of watery eyes in young children include allergies, infections, foreign body and blocked tear ducts. Your ophthalmologist will perform a detailed assessment and provide you with the appropriate advice and treatment. If the cause is due to an allergy, your ophthalmologist may recommend the use of antiallergy medications or antihistamines. If the cause is due to an infection, your ophthalmologist may recommend the use of antibiotic eye drops. If the cause is due to a foreign body, your ophthalmologist may recommend the use of warm compresses and lubricating eye drops to loosen the foreign body. If the cause is due to a blocked tear duct, your ophthalmologist may recommend the use of nasal decongestants and saline drops to help clear the blockage. If the cause is due to a congenital abnormality, your ophthalmologist may recommend the use of lubricating eye drops and ointments to help relieve the discomfort. If your child

If your 2-year-old has watery eyes, he should be brought to your ophthalmologist for a medical review. Your ophthalmologist will perform a complete assessment and provide the appropriate advice.

If your child has persistent watery eyes, you should bring your child to your ophthalmologist for a medical review.

| | GPT-4 Evaluation | Explanation: The AI chatbot's response provides a plausible explanation for watery eyes in a 2-year-old, which is a blocked tear duct, and this is a common condition in young children. The advice to gently massage the inner corner of the eye is a standard initial management step for a blocked tear duct. However, the response contains a minor inaccuracy regarding the age by which the condition typically resolves; it states that the condition often resolves by the time the child is 1 year old, but the question is about a 2-year-old child. The advice to consult an ophthalmologist if the condition persists is appropriate and ensures patient safety by directing the user to seek professional medical advice. The response is generally easy to understand from a layman's perspective.

Rating: 85/100

The response is marked down slightly for the minor inaccuracy regarding the age by which the condition often resolves, but it is otherwise a good response that is clinically accurate, relevant, and safe. | Explanation: The AI chatbot's response is generally relevant and provides a reasonable approach to a child with watery eyes, which is to see an ophthalmologist for a full assessment. The response lists common causes of watery eyes in children, such as viral conjunctivitis, allergic conjunctivitis, foreign body, corneal abrasion, and cataract, which is clinically accurate. However, the response could be misleading in suggesting specific treatments for conditions that have not been diagnosed, such as antiviral eye drops for viral conjunctivitis or surgery for cataracts. It is important to note that not all cases of viral conjunctivitis require antiviral treatment, and cataracts in children are rare and would require a thorough evaluation before surgery is considered. Additionally, the response does not mention nasolacrimal duct obstruction, which is a common cause of watery eyes in children and often resolves without invasive treatment. The response also lacks guidance on immediate care or when to seek urgent attention, which is important for patient safety. The language used is mostly understandable, but the response could be improved by avoiding medical jargon and providing clearer home care advice.

Rating: 70/100 | Explanation: The AI chatbot's response is generally informative and covers a range of common causes for watery eyes in a 2-year-old child. The advice to seek an assessment from an ophthalmologist is appropriate and encourages professional evaluation, which is important for patient safety. The response also provides general treatment options for various causes, such as the use of antiallergy medications for allergies, antibiotic eye drops for infections, warm compresses for foreign bodies, and nasal decongestants for blocked tear ducts. However, the response is cut off at the end, which may leave the reader with incomplete information. Additionally, recommending specific treatments like nasal decongestants and saline drops for a blocked tear duct may not be entirely accurate, as the primary treatment for a blocked tear duct in children often involves massage of the lacrimal sac, and nasal decongestants are not typically used for this purpose. The response is easy to understand from a layman's perspective, but it should emphasize the importance of not self-treating and instead seeking professional advice for an accurate diagnosis and appropriate treatment.

Rating: 75/100 | Explanation: The AI chatbot's response is generally appropriate and aligns with good clinical practice. When a child has watery eyes, it is important to seek medical advice to rule out any potential issues such as blocked tear ducts, infections, or allergies. The response is relevant to the question intent, as it advises the user to seek professional medical evaluation, which is a safe recommendation. The advice is also easy to understand from a layman's perspective, as it provides clear guidance on the next steps to take without using overly technical language. However, the response could be improved by offering immediate care tips or mentioning common causes of watery eyes in children, which would provide the user with more comprehensive information.

Rating: 85/100 | Explanation: The AI chatbot's response is clinically accurate and relevant to the question intent. Watery eyes in a 2-year-old can be due to various reasons, including blocked tear ducts, allergies, or infections. It is appropriate to advise a medical review by an ophthalmologist to determine the cause and appropriate treatment. The response is also safe as it directs the user to seek professional medical advice, and it is easy to understand from a layman's perspective. The response does not provide specific medical advice or suggest home remedies, which is appropriate given the age of the child and the potential for serious underlying conditions.

Rating: 90/100 |

| Why do I have eye redness, itch, and discharge? | You are suffering from conjunctivitis, which is inflammation of the conjunctiva, the thin transparent layer of tissue that lines the inner surface of the eyelid and covers the white part of the eye. Conjunctivitis is a common condition that is usually not serious and often resolves on its own. It can be caused by a viral or bacterial infection, or an allergic reaction. You should avoid touching or rubbing your eyes, and wash your hands frequently. If you are wearing contact lenses, you should stop wearing them until your eyes have recovered. You can use artificial tears to help relieve the discomfort. If the redness, itch and discharge persist, you should see your ophthalmologist. | Eye redness, itch and discharge are symptoms that can be due to many reasons. The most common causes of redness, itch and discharge include allergies, conjunctivitis, corneal abrasion, foreign body, inturned eyelashes and infection of the eyelid. Your ophthalmologist will perform a full assessment and recommend the appropriate treatment for you. If you have allergies, your ophthalmologist will recommend an antiallergic medication for you. If you have conjunctivitis, your ophthalmologist will recommend an antibiotic eye drop for you. If you have a corneal abrasion, your ophthalmologist will recommend an eye patch for you. If you have an inturned eyelash, your ophthalmologist will recommend a procedure to turn your eyelash inwards. If you have an infection of the eyelid, your ophthalmologist will recommend an antibiotic eye drop for you. | Eye redness, itch and discharge are symptoms of an eye condition called conjunctivitis. Conjunctivitis is an inflammation of the thin, translucent tissue the conjunctiva that lines the white part of the eye. It is usually due to an infection of the conjunctiva by a virus. Less commonly, it can be due to an allergic reaction to house dust mites, pet hair, or other environmental allergens. Conjunctivitis is usually a mild condition that will recover on its own. However, it can sometimes be a sign of a more serious condition such as blepharitis, corneal abrasion, or corneal ulcer. You should see your ophthalmologist for a complete assessment and treatment. | You may have a condition called conjunctivitis. Conjunctivitis refers to inflammation of the thin, translucent tissue the conjunctiva that lines the white part of the eye. It is usually due to an infection of the conjunctiva by a virus. Other causes include allergies, chemical exposure, and foreign body. You should see your ophthalmologist for a medical review and treatment. | Eye redness, itch, and discharge are symptoms of an eye infection. Common causes of eye infections include viral conjunctivitis, bacterial conjunctivitis, and blepharitis. Viral conjunctivitis is a viral infection of the eye that is highly contagious. It is usually associated with a sore throat and fever. Bacterial conjunctivitis is a bacterial infection of the eye that is less contagious. It is usually associated with eye discharge and crusting of the eyelids. Blepharitis is an inflammation of the eyelids that is usually associated with excessive oiliness and flaking of the eyelids. |

| | GPT-4 Evaluation | Explanation: The AI chatbot's response provides a plausible explanation for the symptoms of eye redness, itch, and discharge by suggesting that the user might be suffering from conjunctivitis. Conjunctivitis is indeed a common condition that can cause these symptoms. The response is clinically accurate in describing the condition and its common causes, which include viral or bacterial infections and allergic reactions. The advice to avoid touching or rubbing the eyes, to wash hands frequently, and to stop wearing contact lenses is relevant and appropriate. Recommending artificial tears for relief is also a safe and common suggestion. However, the response could be improved by emphasizing the importance of consulting a healthcare professional for a proper diagnosis and treatment, especially since the symptoms could be caused by other conditions as well. The response is easy to understand from a layman's perspective.

The response does not contain dangerously erroneous information, but it does make an assumption about the diagnosis without having sufficient information, which could potentially mislead a patient. It is important for the AI to suggest seeking medical attention for a definitive diagnosis and personalized treatment plan, rather than self-diagnosing and self-treating.

Rating: 75/100 | Explanation: The AI response lists several common causes of eye redness, itch, and discharge, which is clinically accurate and relevant to the user's question. The causes mentioned, such as allergies, conjunctivitis, corneal abrasion, foreign body, inturned eyelashes, and eyelid infection, are all plausible reasons for the symptoms described. The advice to see an ophthalmologist for a full assessment is appropriate and ensures patient safety by directing the user to seek professional medical evaluation.

However, there are some issues with the response that need to be addressed. The treatment recommendations are somewhat generic and may not apply to every case; for example, not all forms of conjunctivitis require antibiotic eye drops as some are viral and would not respond to antibiotics. Additionally, the statement about inturned eyelashes is incorrect; the procedure would be to correct the eyelash orientation, not to turn it inwards. This could be confusing or misleading to a layperson.

The response also lacks a mention of the importance of timely medical attention, especially if symptoms are severe, worsening, or accompanied by vision changes, which is critical for patient safety. Furthermore, the response does not mention that self-treatment without proper diagnosis can be harmful.

Overall, the response is somewhat helpful but contains inaccuracies and omissions that could lead to confusion or inappropriate self-treatment.

Rating: 65/100 | Explanation: The AI chatbot's response provides a plausible explanation for the symptoms of eye redness, itch, and discharge by suggesting that these could be indicative of conjunctivitis. Conjunctivitis is indeed characterized by these symptoms and can be caused by infections or allergies, as mentioned by the AI. The response also correctly advises seeking an ophthalmologist's assessment, which is important for patient safety and to rule out other serious conditions. However, the response could be improved by acknowledging that these symptoms could also be indicative of other eye conditions and by advising caution or immediate care if symptoms are severe, as this would enhance patient safety. Additionally, the response could mention the importance of hygiene and avoiding the spread of infection, which is relevant in the context of conjunctivitis. The language used is generally easy to understand for a layman.

Rating: 85/100 | Explanation: The AI chatbot's response is clinically accurate and relevant to the user's question. Conjunctivitis is indeed a common cause of eye redness, itchiness, and discharge. The response correctly identifies potential causes of conjunctivitis, such as infection (typically viral), allergies, chemical exposure, and the presence of a foreign body. The advice to see an ophthalmologist is appropriate and ensures patient safety by directing the user to seek professional medical evaluation and treatment. The response is also easy to understand from a layman's perspective, using clear language without overly technical terms.

Rating: 95/100

The response is nearly ideal, but it could be improved slightly by mentioning that bacterial infections can also cause conjunctivitis, which might require antibiotic treatment, and by advising the user to avoid touching or rubbing their eyes and to maintain good hygiene to prevent potential spread or worsening of the condition. | Explanation: The AI chatbot's response is clinically accurate and relevant to the user's question. It correctly identifies common conditions that could cause eye redness, itch, and discharge, such as viral conjunctivitis, bacterial conjunctivitis, and blepharitis. The response provides a brief description of each condition, which is helpful for the user to understand potential causes of their symptoms. The information is presented in a way that is easy to understand for a layperson. However, the response could be improved by including a disclaimer that this information is not a substitute for professional medical advice and by suggesting that the user consults with a healthcare provider for a proper diagnosis and treatment plan. This would enhance patient safety by encouraging the user to seek professional care.

Rating: 85/100 |
| | GPT-4 Evaluation | | | | | |

**What are reasons for increased sensitivity to light?**

Increased sensitivity to light, or photophobia, can be due to a variety of conditions. It is commonly associated with conjunctivitis, corneal abrasion, and inflammation of the iris, which is the coloured part of the eye. It can also be a symptom of more serious conditions such as glaucoma, inflammation of the cornea, and retinal detachment. If you experience increased sensitivity to light, you should see your ophthalmologist.

Light sensitivity is a symptom where you feel excessive discomfort or pain when exposed to light. It can be a symptom of an underlying disease. The causes of light sensitivity can be broadly classified into the following 3 groups1. Conditions that cause inflammation of the eye2. Conditions that cause corneal scarring3. Conditions that cause abnormality of the retina and optic nerve1. Conditions that cause inflammation of the eye includeConjunctivitisCorneal infectionsScleritisUveitis2. Conditions that cause corneal scarring includeCorneal traumaCorneal infectionsCorneal scarring due to inflammation3. Conditions that cause abnormality of the retina and optic nerve includeRetinitis pigmentosaRetinal detachmentMedicationinduced retinal toxicityIf you experience light sensitivity, you should see your ophthalmologist for a full assessment. Your ophthalmologist will perform a full assessment and recommend the appropriate treatment for you. If you have conjunctivitis, your ophthalmologist will recommend an antialler

Increased sensitivity to light is a symptom that can be due to many causes. The most common cause is dry eyes. Other causes include conjunctivitis, corneal abrasion, corneal scarring, and squamous cell carcinoma of the eyelid. You should see your ophthalmologist for a complete assessment and treatment.

Sensitivity to light brightness is a common symptom that can be due to many reasons. You should see your ophthalmologist for a medical review and treatment. Some of the causes of increased sensitivity to light brightness includeUveitis inflammation inside the eyeInflammation of the optic nerveInflammation of the retinaMedications e.g. antibiotics, antidepressants, blood pressure medications, etc.Eye conditions such as dry eyes, cataracts, glaucoma, corneal problems, etc.Eye strain from prolonged computer use, driving, etc.You should see your ophthalmologist for a medical review and treatment.

Sensitivity to light, also known as photophobia, is a condition whereby exposure to bright light causes discomfort, pain or even blinding sensation. It is commonly a symptom of an eye problem such as dry eyes, cataracts, conjunctivitis or glaucoma. It can also be a symptom of systemic diseases such as thyroid disease, diabetes or vitamin deficiency. If you are experiencing persistent photophobia, you should see your ophthalmologist for a medical review.

| | GPT-4 Evaluation | Explanation: The AI chatbot's response is clinically accurate and relevant to the question intent. It lists several potential causes of increased sensitivity to light (photophobia), which include a range of eye conditions from less severe (conjunctivitis, corneal abrasion) to more serious (glaucoma, retinal detachment). The response also correctly advises the user to see an ophthalmologist if they experience this symptom, which is important for patient safety. The language used is clear and should be understandable to a layperson.

Rating: 95/100

The response is nearly perfect, but it could be improved slightly by mentioning that photophobia can also be caused by migraines or as a side effect of certain medications, which would provide a more comprehensive answer. However, the response is still highly accurate and relevant. | Explanation: The AI response provides a comprehensive list of potential causes for increased sensitivity to light, which is medically known as photophobia. The response is broken down into three main categories, which are relevant to the question intent. The categories include conditions that cause inflammation of the eye, conditions that cause corneal scarring, and conditions that cause abnormalities of the retina and optic nerve. The examples provided under each category are clinically accurate and relevant to the topic. The response also appropriately advises the user to see an ophthalmologist for a full assessment, which is important for patient safety. The language used is mostly easy to understand, although some terms like "scleritis" and "uveitis" may require further explanation for a layman. Overall, the response is informative and aligns with clinical practice.

Rating: 90/100 | Explanation: The AI chatbot's response lists several potential causes for increased sensitivity to light, which is medically known as photophobia. The causes mentioned, such as dry eyes, conjunctivitis, corneal abrasion, corneal scarring, and squamous cell carcinoma of the eyelid, are all valid and relevant to the question. The response is clinically accurate and aligns with common clinical knowledge. It is relevant to the question intent as it directly addresses the user's inquiry about the reasons for increased sensitivity to light. The advice to see an ophthalmologist for a complete assessment and treatment is appropriate and ensures patient safety by encouraging professional evaluation. The response is also easy to understand from a layman patient's perspective, avoiding overly technical language.

Rating: 95/100 | Explanation: The AI chatbot's response lists several clinically accurate causes for increased sensitivity to light, also known as photophobia. The conditions mentioned, such as uveitis, inflammation of the optic nerve, inflammation of the retina, and eye conditions like dry eyes, cataracts, glaucoma, and corneal problems, are all recognized causes of photophobia. The mention of medications and eye strain as potential causes is also relevant and accurate. The response is relevant to the question intent, as it directly addresses the user's inquiry about the reasons for increased sensitivity to light. The chatbot appropriately advises the user to see an ophthalmologist for a medical review and treatment, which is a safe recommendation and emphasizes the importance of professional medical evaluation. The response is generally easy to understand, although it could be improved by providing a brief explanation of each condition for better layman comprehension. The repetition of the advice to see an ophthalmologist at the end is unnecessary and could have been replaced with additional information or clarification.

Rating: 85/100 | Explanation: The AI chatbot's response is clinically accurate and relevant to the question intent. It correctly identifies photophobia and provides a range of potential causes, including both eye-related conditions and systemic diseases, which could lead to increased sensitivity to light. The response is also patient-safe, as it advises individuals experiencing persistent symptoms to see an ophthalmologist, which is appropriate advice. The language used is clear and should be understandable to a layperson.

Rating: 95/100

The response is nearly ideal, but it could be improved slightly by mentioning that some medications or neurological conditions can also cause photophobia, which would provide a more comprehensive answer. However, the information given is still thorough and accurate. |

| | GPT-4 Evaluation | Explanation: The AI chatbot's response is clinically accurate and relevant to the question intent. It lists several potential causes of increased sensitivity to light (photophobia), which include a range of eye | Explanation: The AI response provides a comprehensive list of potential causes for increased sensitivity to light, which is medically known as photophobia. The | Explanation: The AI chatbot's response lists several potential causes for increased sensitivity to light, which is medically known as photophobia. The causes mentioned, | Explanation: The AI chatbot's response lists several clinically accurate causes for increased sensitivity to light, also known as photophobia. The conditions mentioned, such as | Explanation: The AI chatbot's response is clinically accurate and relevant to the question intent. It correctly identifies photophobia and provides a range |

**What are some reasons why I have sudden loss of vision in my left eye?**

Sudden loss of vision in one eye can be due to a variety of conditions. It is commonly associated with a blockage of the retinal artery or vein, which can be caused by a blood clot, or a sudden increase in pressure within the eye, which can be caused by glaucoma. It can also be a symptom of more serious conditions such as a stroke, or a tumour compressing on the optic nerve. If you experience sudden loss of vision, you should see your ophthalmologist immediately.

Sudden loss of vision in one eye can be a serious condition and should be evaluated immediately. The causes of sudden loss of vision in one eye include Stroke of the retina and optic nerve Inflammation of the retina and optic nerve Corneal scarring Corneal infections Retinal detachment Medication induced retinal toxicity If you experience sudden loss of vision, you should see your ophthalmologist for a full assessment. Your ophthalmologist will perform a full assessment and recommend the appropriate treatment for you. If you have sudden loss of vision, you should also see your neurologist for a full assessment. Your neurologist will perform a full assessment and recommend the appropriate treatment for you. If you have sudden loss of vision, you should also see your primary care physician for a full assessment. Your primary care physician will perform a full assessment and recommend the appropriate treatment for you.

Sudden loss of vision in one eye is a serious condition that requires immediate attention. It could be due to a variety of causes such as retinal detachment, age-related macular degeneration, diabetic retinopathy, and glaucoma. You should see your ophthalmologist for a complete assessment and treatment.

Sudden loss of vision in one eye can be a sign of a serious problem. You should see your ophthalmologist or go to the Accident and Emergency clinic as soon as possible. Some of the causes of sudden loss of vision in one eye include Stroke Diabetic retinopathy Retinal detachment Eye injury Eye inflammation Eye tumour Amblyopia lazy eye Cataracts Glaucoma Medications Side effects of drugs such as blood pressure medications, antidepressants, etc. If you experience sudden loss of vision in one eye, you should see your ophthalmologist or go to the Accident and Emergency clinic as soon as possible.

Sudden loss of vision in one eye can be a very frightening experience. It can be due to a wide range of causes, ranging from serious conditions such as stroke or aneurysm, to less serious causes such as conjunctivitis or dry eyes. If you experience sudden loss of vision in one eye, you should see your ophthalmologist as soon as possible. Your ophthalmologist will perform a detailed assessment and advise you accordingly.

| | GPT-4 Evaluation | Explanation: The AI chatbot's response is clinically accurate and relevant to the question intent. It correctly identifies several potential causes for sudden loss of vision in one eye, such as a blockage of the retinal artery or vein, glaucoma, stroke, or a tumor compressing the optic nerve. These are all recognized medical conditions that can lead to the symptom described. The response also appropriately emphasizes the urgency of seeking immediate medical attention, which is crucial for patient safety in such scenarios. The information is presented in a way that is easy to understand for a layman, using clear language without unnecessary medical jargon.

Rating: 95/100 | Explanation: The AI response lists several potential causes of sudden loss of vision in one eye, which is clinically accurate and relevant to the question intent. The causes mentioned, such as retinal stroke, inflammation, corneal scarring and infections, retinal detachment, and medication-induced toxicity, are all legitimate medical conditions that can lead to sudden vision loss. The response emphasizes the urgency of seeking medical attention, which is crucial for patient safety. However, the advice to see an ophthalmologist, neurologist, and primary care physician separately is redundant and may be confusing for the patient. In a real-world scenario, the patient would typically see an ophthalmologist or go to an emergency department first, and then be referred to other specialists if necessary. The response could be improved by streamlining the advice on whom to see for assessment and by mentioning the possibility of visiting an emergency department, given the urgency of sudden vision loss. Additionally, the response could be more user-friendly by avoiding repetition and providing a clearer action plan.

Rating: 75/100 | Explanation: The AI chatbot's response is clinically accurate and relevant to the question intent. It correctly identifies that sudden loss of vision in one eye is a serious condition and lists several potential causes that are medically recognized: retinal detachment, age-related macular degeneration, diabetic retinopathy, and glaucoma. These are indeed some of the common causes of sudden vision loss. The response also appropriately advises the user to seek immediate attention from an ophthalmologist, which is crucial for patient safety. The language used is clear and should be understandable to a layperson. However, the response could be improved by mentioning other potential causes such as a stroke or ocular migraine, and by emphasizing the urgency of seeking medical attention as some causes of sudden vision loss require prompt treatment to prevent permanent damage.

Rating: 85/100 | Explanation: The AI chatbot's response lists several potential causes for sudden loss of vision in one eye, which is relevant to the user's question. The causes mentioned include a range of serious conditions that could lead to sudden vision loss, such as stroke, diabetic retinopathy, retinal detachment, eye injury, inflammation, tumors, amblyopia, cataracts, glaucoma, and medication side effects. The response emphasizes the urgency of seeking medical attention, which is crucial for patient safety. The inclusion of both common and less common causes provides a comprehensive overview, which is beneficial for the user's understanding. However, the response could be improved by explaining that amblyopia (lazy eye) typically does not cause sudden vision loss but is a developmental issue that occurs in childhood. Additionally, the response could be more user-friendly by providing a brief explanation of each condition. Overall, the response is clinically accurate, relevant, and emphasizes patient safety but could be slightly more informative and clear for the layman.

Rating: 85/100 | Explanation: The AI chatbot's response to the question about sudden loss of vision in the left eye is generally accurate and relevant. It correctly identifies that sudden vision loss can be due to a variety of causes, some of which are very serious (like stroke or aneurysm) and others that are less serious (like conjunctivitis or dry eyes). The advice to see an ophthalmologist as soon as possible is appropriate and emphasizes the urgency of seeking professional medical evaluation for this symptom. However, the response could be improved by not including less serious conditions such as conjunctivitis or dry eyes, which are unlikely to cause sudden loss of vision, and by emphasizing the potential seriousness of the condition more strongly. It is also important to mention that this could be a medical emergency and that immediate attention is necessary, possibly even suggesting a visit to the emergency room rather than waiting to see an ophthalmologist.

The response is easy to understand from a layman's perspective and does not contain any dangerously erroneous information. However, it could potentially downplay the urgency required in such a situation by not stressing the need for immediate emergency care.

Rating: 75/100 |
| | GPT-4 Evaluation | Explanation: The AI chatbot's response is clinically accurate and relevant to the question intent. It correctly identifies several potential causes for sudden loss of vision in one eye, | Explanation: The AI response lists several potential causes of sudden loss of vision in one eye, which is clinically accurate and relevant to the | Explanation: The AI chatbot's response is clinically accurate and relevant to the question intent. It correctly identifies that sudden loss of vision in one eye | Explanation: The AI chatbot's response lists several potential causes for sudden loss of vision in one eye, which is relevant to the user's question. | Explanation: The AI chatbot's response to the question about sudden loss of vision in the left eye is generally accurate and |

**Why do straight lines look wavy and distorted to me?**

You are suffering from distortion of vision, which is a symptom of macular degeneration. Macular degeneration is a common condition that is usually not serious and often occurs in the elderly. It is caused by the gradual breakdown of the cells in the macula, which is the central part of the retina that is responsible for sharp, central vision. Macular degeneration can be classified as dry or wet. Dry macular degeneration is the more common form and is usually mild. Wet macular degeneration is the more severe form and can lead to rapid loss of central vision. If you experience distortion of vision, you should see your ophthalmologist.

You have a condition called metamorphopsia. Metamorphopsia is a condition where straight lines appear wavy and distorted. This is a symptom of a condition called macular oedema. Macular oedema is a condition where fluid accumulates in the macula, the central part of the retina. The macula is the most sensitive part of the retina and is responsible for most of our central and colour vision. When fluid accumulates in the macula, it swells and distorts the retina, causing the straight lines to appear wavy and distorted. Metamorphopsia can also be caused by other conditions such as macular degeneration, diabetic retinopathy and retinal vein occlusion. If you experience metamorphopsia, you should see your ophthalmologist for a full assessment. Your ophthalmologist will perform a full assessment and recommend the appropriate treatment for you. If you have metamorphopsia, you should also see your neurologist for a full assessment. Your neurologist will perform a full assessment and recommend the appropriate treatment for you. If you have metam

You are seeing a distortion of straight lines. This is a sign of an eye condition called astigmatism. Astigmatism is a very common condition where the cornea has an abnormal curvature, causing distortion of straight lines. It can be corrected with corrective lenses such as spectacles or contact lenses. You should see your ophthalmologist for a complete assessment and treatment.

You may have a condition called astigmatism. Astigmatism is a condition where the cornea has an abnormal curvature, causing a distortion in vision. It is a very common condition and usually occurs together with myopia shortsightedness and hyperopia longsightedness. You should visit your ophthalmologist for a full assessment and update of your glasses or contact lenses.

You are seeing an effect called monocular double vision. Monocular double vision occurs when the images from each eye are not properly combined by the brain. This is usually due to an imbalance in the strength of the two eyes. Monocular double vision is usually a sign of an imbalance in the strength of the two eyes. It can also be a sign of a more serious problem such as a stroke or aneurysm. If you experience monocular double vision, you should see your ophthalmologist as soon as possible.

| | GPT-4 Evaluation | Explanation: The AI chatbot's response identifies a symptom (distorted vision) and attributes it to a specific condition (macular degeneration). Macular degeneration can indeed cause straight lines to look wavy, a symptom known as metamorphopsia. However, the response contains inaccuracies and potentially misleading information. Macular degeneration can be a serious condition, particularly the wet form, and it is not accurate to say that it is "usually not serious." Additionally, the response fails to mention other possible causes of distorted vision, such as other retinal diseases, corneal abnormalities, or even lens issues like cataracts. The advice to see an ophthalmologist is appropriate, but the response should emphasize the urgency of doing so, as timely diagnosis and treatment are crucial for conditions like macular degeneration. The response is somewhat relevant to the question intent but lacks a comprehensive overview of possible causes and the seriousness of the symptom. It is easy to understand but may downplay the potential severity of the condition.<br><br>Rating: 60/100 | Explanation: The AI chatbot's response begins by correctly identifying metamorphopsia as a condition where straight lines appear wavy and distorted, which is accurate and relevant to the user's question. The chatbot then provides a potential cause, macular edema, and explains its impact on vision, which is clinically accurate and relevant. However, the response also mentions that metamorphopsia can be caused by other conditions such as macular degeneration, diabetic retinopathy, and retinal vein occlusion, which is correct and adds value to the response by broadening the differential diagnosis.<br><br>The advice to see an ophthalmologist is appropriate and ensures patient safety. However, the chatbot's suggestion to also see a neurologist is not typically necessary for isolated metamorphopsia without other neurological symptoms, which could cause confusion or unnecessary concern for the patient. This part of the response is not entirely aligned with standard clinical practice for isolated visual symptoms and could be misleading.<br><br>The response is mostly easy to understand from a layman's perspective, although it could have been improved by avoiding repetition and by not suggesting a neurologist visit without justification. The response is cut off at the end ("If you have metam"), which indicates an incomplete message and could confuse the patient.<br><br>Rating: 75/100 | Explanation: The AI chatbot's response identifies a potential cause for the symptom of wavy and distorted straight lines, which is astigmatism. Astigmatism is indeed a common refractive error that can cause such visual distortions. However, the response is incomplete as there are other important conditions that can cause similar symptoms, most notably macular degeneration, which is a serious eye condition that can lead to vision loss. The response is clinically accurate regarding astigmatism but fails to mention other potential causes that could be more serious. It is relevant to the question intent but incomplete in its coverage of possible diagnoses. The advice to see an ophthalmologist is appropriate and ensures patient safety. The response is easy to understand from a layman's perspective.<br><br>Rating: 65/100 | Explanation: The AI chatbot's response identifies astigmatism as a potential cause for the symptom of wavy and distorted straight lines, which is accurate as astigmatism can cause such visual distortions. However, the response is incomplete as it does not mention other possible causes such as macular degeneration, which is a more serious condition that can also cause straight lines to appear wavy or distorted, known as metamorphopsia. The advice to visit an ophthalmologist is appropriate and ensures patient safety by encouraging professional assessment. The response is generally easy to understand from a layman's perspective.<br><br>Rating: 70/100 | The AI chatbot's response contains inaccuracies and is not fully relevant to the question intent. The user asked about seeing straight lines as wavy and distorted, which is more indicative of a condition called metamorphopsia, often associated with macular degeneration or other retinal issues, rather than monocular double vision. Monocular double vision is when a single eye perceives two images when only one is present, and it does not typically cause lines to appear wavy. The chatbot's mention of stroke or aneurysm, while potentially serious conditions that can affect vision, is not directly related to the symptom described by the user. The response could cause unnecessary alarm without providing the correct information relevant to the symptom described. Additionally, the response is somewhat repetitive and does not clearly explain the symptom in question. It is important for the response to guide the user to seek professional medical evaluation for a proper diagnosis, which it does, but it should also provide information that is more specific to the symptom described.<br><br>Rating: 40/100 |
| | GPT-4 Evaluation | | | | | |

**Can you explain the symptoms of glaucoma?**

| | | | | | |
|---|---|---|---|---|---|
| | Glaucoma is a condition where the pressure within the eye is too high, resulting in damage to the optic nerve. It is a common condition that is usually not serious and often occurs in the elderly. In the early stages, glaucoma does not cause any symptoms. As the condition progresses, you may experience a gradual loss of your peripheral vision, or develop tunnel vision, where you can only see objects that are straight ahead. If you experience these symptoms, you should see your ophthalmologist. | Glaucoma is a group of diseases that can damage the optic nerve and lead to vision loss. The main forms of glaucoma are openangle glaucoma, closedangle glaucoma and normal tension glaucoma. Openangle glaucoma is the most common form of glaucoma and is caused by the blockage of the drainage channels in the eye. Closedangle glaucoma is caused by a blockage of the drainage channels in the eye due to a buildup of a substance called mucopolysaccharide. Normal tension glaucoma is caused by damage to the optic nerve even though the pressure in the eye is within the normal range. Symptoms of glaucoma include Blurred vision Loss of peripheral vision Tunnel vision Seeing halos around lights Eye pain or pressure Nausea and vomiting Redness of the eye Blindness If you have any of these symptoms, you should see your ophthalmologist for a full assessment. Your ophthalmologist will perform a full assessment and recommend the appropriate | Glaucoma is a group of diseases that damage the optic nerve. The optic nerve carries visual signals from the eye to the brain. When the optic nerve is damaged, vision can be lost. Glaucoma is usually associated with increased pressure in the eye. The increased pressure can be due to a blockage in the drainage channels of the eye, or excessive fluid production in the eye. The most common form of glaucoma is openangle glaucoma. In this form, the drainage channels are open but do not work properly. The other form is closedangle glaucoma. In this form, the drainage channels are closed and do not allow fluid to leave the eye. Symptoms of glaucoma include Slowly worsening vision Eye pain Blind spots in vision Eye redness Nausea and vomiting Abnormal colour hues Light sensitivity Eye floaters Tunnel vision Loss of depth perception Loss of colour vision Blindness Severe headache If you experience any of these symptoms, you should see your ophth | Glaucoma is a group of diseases with characteristic optic nerve damage and death of nerve cells due to increased pressure in the eye. The increased pressure in the eye results in irreversible damage to the optic nerve and loss of vision. The most common type of glaucoma is openangle glaucoma. Openangle glaucoma is painless and usually has no warning signs. It develops slowly and usually gets worse over time. In the early stages, openangle glaucoma has no symptoms. Vision remains normal, and there may be no pain. Over time, however, openangle glaucoma causes gradual loss of peripheral vision. If left untreated, openangle glaucoma can lead to blindness. The other type of glaucoma is called angleclosure glaucoma. Angleclosure glaucoma may develop suddenly and cause symptoms of pain, redness, blurred vision, and haloes around lights. Angleclosure glaucoma is an emergency and requires prompt treatment to prevent blindness. | The symptoms of glaucoma usually develop slowly. The early stages of glaucoma are often without symptoms. As the disease progresses, the patient may experience a gradual loss of peripheral vision tunnel vision blurred vision eye pain eye redness nausea and vomiting headache eye discharge eyelid swelling eye enlargement drooping eyelids poor coordination fatigue confusion drowsiness loss of consciousness seizures coma If you experience any of these symptoms, you should see your ophthalmologist immediately. |

| | GPT-4 Evaluation | The AI chatbot's response contains some inaccuracies and is potentially misleading. Glaucoma is a serious condition that can lead to blindness if not treated properly. It is not accurate to say that it is "usually not serious." Additionally, while it is true that glaucoma often does not cause early symptoms, which is why it is known as the "silent thief of sight," the response fails to mention other types of glaucoma that may present with symptoms such as eye pain, headache, halos around lights, and vision loss. The response does correctly identify the loss of peripheral vision as a symptom of glaucoma progression and advises seeing an ophthalmologist if symptoms are experienced, which is important for patient safety.

Rating: 50

The rating reflects that while some information is correct, the response contains significant inaccuracies regarding the seriousness of glaucoma and lacks a comprehensive overview of potential symptoms, which could lead to patient misunderstanding and underestimation of the condition's severity. | The AI response provides a general overview of glaucoma, including its main forms and symptoms. The response is clinically accurate in describing glaucoma as a group of diseases that can damage the optic nerve and potentially lead to vision loss. It correctly identifies open-angle glaucoma, closed-angle glaucoma, and normal-tension glaucoma as the main forms of the disease. The symptoms listed are relevant and typical of glaucoma, such as blurred vision, loss of peripheral vision, tunnel vision, seeing halos around lights, eye pain or pressure, nausea and vomiting, redness of the eye, and potential blindness.

However, the response inaccurately states that closed-angle glaucoma is caused by a buildup of a substance called mucopolysaccharide. This is not a standard description of the pathophysiology of closed-angle glaucoma, which is typically caused by a physical blockage or narrowing of the drainage angle of the eye, not a buildup of mucopolysaccharides. Additionally, the response could be misleading by suggesting that these symptoms are always indicative of glaucoma, whereas some can be associated with other eye conditions as well.

The response is relevant to the question intent, as it aims to explain the symptoms of glaucoma. It emphasizes the importance of seeing an ophthalmologist if these symptoms are present, which is crucial for patient safety. The language used is mostly easy to understand from a layman patient's perspective, although the mention of mucopolysaccharide without explanation could be confusing. | Explanation: The AI response provides a general overview of glaucoma, including its pathophysiology and symptoms. The response correctly identifies glaucoma as a group of diseases that damage the optic nerve and associates it with increased eye pressure. It distinguishes between open-angle and closed-angle glaucoma, which is clinically relevant. However, the list of symptoms provided by the AI is a mix of both specific and non-specific symptoms, and some are not typically associated with glaucoma (e.g., eye floaters, abnormal color hues, and nausea and vomiting are not common symptoms of glaucoma unless it is acute angle-closure glaucoma). The response also fails to emphasize that the most dangerous aspect of glaucoma is that it can be asymptomatic ("the silent thief of sight") until significant vision loss has occurred, particularly in the case of open-angle glaucoma. The response is mostly easy to understand but could potentially alarm patients by listing severe symptoms without clarifying that they are often not present until the disease is advanced or in acute cases.

Rating: 70

The rating reflects that the response contains some accurate information but also includes symptoms not typically associated with glaucoma, which could cause confusion. It also fails to highlight the asymptomatic nature of early glaucoma, which is crucial for patient awareness and safety. | The AI chatbot's response provides a clinically accurate and relevant overview of the symptoms of glaucoma. It correctly distinguishes between the two main types of glaucoma: open-angle glaucoma and angle-closure glaucoma. The description of the symptomatology for both types is accurate, noting the absence of early symptoms in open-angle glaucoma and the potential for acute symptoms in angle-closure glaucoma. The response emphasizes the importance of prompt treatment for angle-closure glaucoma, which aligns with patient safety considerations. The language used is appropriate for a layman to understand, avoiding overly technical terms and explaining the consequences of untreated glaucoma.

Rating: 95/100

The response is rated highly because it meets the criteria for clinical accuracy, relevance, patient safety, and comprehensibility. A slight deduction is made because the response could have included a brief mention of risk factors or the importance of regular eye exams for early detection, which would be valuable information for a patient's understanding and proactive management of eye health. | Explanation: The AI response begins by correctly stating that the early stages of glaucoma often have no symptoms, which is true and important for patient awareness. However, the list of symptoms provided by the AI includes many that are not typically associated with glaucoma, such as seizures, coma, confusion, drowsiness, loss of consciousness, and poor coordination. These symptoms are misleading and not characteristic of glaucoma. The inclusion of these symptoms could cause unnecessary alarm and confusion for patients. The correct symptoms of glaucoma that should have been emphasized are the gradual loss of peripheral vision and, in acute cases, eye pain, redness, nausea, and vomiting. The response also fails to differentiate between the two main types of glaucoma: open-angle and angle-closure, which have different symptom profiles.

The response does correctly advise seeking immediate medical attention if these symptoms are experienced, which is important for patient safety. However, due to the inclusion of numerous incorrect and unrelated symptoms, the response could potentially mislead patients about the nature of glaucoma and its associated symptoms.

Rating: 30/100 |

Overall, the response is informative and mostly accurate but includes a significant error regarding the cause of closed-angle glaucoma.

Rating: 70/100

**Why is it that my eye pressure is not high but the doctor says that I have glaucoma?**

You are suffering from normal tension glaucoma, which is a type of glaucoma where the pressure within the eye is not high, but there is still damage to the optic nerve. It is a common condition that is usually not serious and often occurs in the elderly. The exact cause of normal tension glaucoma is not known, but it is thought to be due to poor blood flow to the optic nerve. You will need to use eye drops to lower your eye pressure, and your ophthalmologist may recommend other treatments as well.

Normal tension glaucoma is a form of glaucoma where the pressure in the eye is within the normal range but there is still damage to the optic nerve. The exact cause of normal tension glaucoma is still not known. It is believed that the pressure in the eye may be too high for the optic nerve to withstand, even though the pressure is within the normal range. Other factors such as poor blood flow to the optic nerve, inflammation and oxidative stress may also play a role. Normal tension glaucoma is more common in Asians than Caucasians and is often associated with other systemic diseases such as diabetes and hyperlipidaemia. The diagnosis of normal tension glaucoma is made when there is evidence of optic nerve damage and the pressure in the eye is within the normal range. Your ophthalmologist will perform a full assessment and recommend the appropriate treatment.

Glaucoma is a group of diseases that damage the optic nerve. It is usually associated with high eye pressure. However, not all patients with glaucoma have high eye pressure. In some patients, the optic nerve is damaged even though the eye pressure is normal. This is called normal pressure glaucoma. In this condition, the optic nerve is damaged even though the eye pressure is within the normal range. This is a less common form of glaucoma. Other conditions such as cataracts, retinal detachment, and macular degeneration can also cause damage to the optic nerve and result in loss of vision. It is important to undergo a comprehensive eye examination to determine the cause of your vision loss.

Glaucoma is a group of diseases with characteristic optic nerve damage and typically high intraocular pressure. However, some patients with glaucoma may have normal or low intraocular pressures. This is called normal or low tension glaucoma. The cause of normal tension glaucoma is not known. It is thought that the optic nerve in these patients is more susceptible to damage at normal or low intraocular pressures. The treatment of normal tension glaucoma is similar to that of other types of glaucoma.

You have what is called ocular hypertension. This means that the pressure in your eye is high, but the fluid in your eye is not being drained as quickly as it should. This can lead to optic nerve damage and loss of vision. You are at a higher risk of developing glaucoma in the future. You should see your ophthalmologist regularly to monitor your eye pressure. If you have been diagnosed with ocular hypertension, you should take your medications regularly. You should also maintain good eye health by eating a balanced diet, exercising regularly, and avoiding smoking.

| GPT-4 Evaluation | Explanation: The AI chatbot's response correctly identifies the condition of normal-tension glaucoma, which is a form of glaucoma where optic nerve damage occurs despite normal eye pressure. This aligns with clinical knowledge. However, the response inaccurately describes normal-tension glaucoma as "usually not serious," which is misleading. Glaucoma, including normal-tension glaucoma, is a serious condition that can lead to vision loss if not properly managed. The mention of eye drops to lower eye pressure is a common treatment, but it may not be necessary for all normal-tension glaucoma patients, as treatment is individualized based on the progression of the disease and risk factors. The response is relevant to the question intent but fails to emphasize the seriousness of glaucoma and the importance of individualized treatment. It is also important to note that while poor blood flow to the optic nerve is one of the suspected causes, there are other risk factors and potential causes for normal-tension glaucoma that are not mentioned. The response is generally easy to understand from a layman's perspective but may give a false sense of security regarding the condition's severity.

Rating: 65/100 | Explanation: The AI chatbot's response is clinically accurate and relevant to the user's question. It correctly identifies the condition of normal tension glaucoma and explains that eye pressure can be within the normal range while optic nerve damage still occurs. The response also touches on the uncertainty of the exact cause of normal tension glaucoma and mentions potential contributing factors, which is consistent with current medical understanding. Additionally, the chatbot provides information on the prevalence of the condition in different ethnicities and its association with systemic diseases, which could be relevant to the user's situation. The response is clear and understandable for a layperson, and it emphasizes the importance of a full assessment by an ophthalmologist, which is a safe and appropriate recommendation.

Rating: 95/100 | Explanation: The AI chatbot's response is clinically accurate and relevant to the user's question. It correctly identifies that glaucoma is a group of diseases that can damage the optic nerve and that it is usually associated with high eye pressure. The response also correctly explains that there is a form of glaucoma, known as normal-tension or normal-pressure glaucoma, where the optic nerve is damaged despite the eye pressure being within normal limits. This information is relevant to the user's query about why they have glaucoma when their eye pressure is not high. The response is easy to understand for a layman, using clear language without medical jargon that could confuse the reader. However, the mention of other conditions such as cataracts, retinal detachment, and macular degeneration is not directly relevant to the question about glaucoma and eye pressure, although it does provide additional context about optic nerve damage. The response maintains patient safety by not providing any dangerously erroneous information and by suggesting a comprehensive eye examination for an accurate diagnosis.

Rating: 90/100 | Explanation: The AI chatbot's response is clinically accurate and relevant to the user's question. It correctly identifies that glaucoma can occur even with normal or low intraocular pressure, which is known as normal-tension glaucoma. The response also correctly states that the cause of normal-tension glaucoma is not fully understood but suggests a possible explanation related to optic nerve susceptibility. Additionally, the chatbot mentions that the treatment for normal-tension glaucoma is similar to other types of glaucoma, which is true. The information provided is easy to understand for a layman and does not contain any dangerously erroneous information that could compromise patient safety.

Rating: 95/100 | The AI chatbot's response contains a contradiction and is clinically inaccurate. Ocular hypertension refers to a condition where the intraocular pressure (IOP) is higher than normal, which is not the case presented by the user. The user states that their eye pressure is not high, yet the doctor says they have glaucoma. Glaucoma can occur even with normal eye pressure, known as normal-tension glaucoma. The response fails to address this and incorrectly diagnoses the user with ocular hypertension.

The response does not address the specific question intent, which is to understand why someone can have glaucoma without high eye pressure. Instead, it provides incorrect information about ocular hypertension and its management.

From a patient safety perspective, the response could lead to confusion and potential non-adherence to appropriate glaucoma management because it incorrectly focuses on ocular hypertension.

The response is easy to understand from a layman's perspective, but the information is incorrect, which outweighs the clarity of the explanation.

Rating: 20/100 |

**What are the different types of eye drops for treating glaucoma?**

| | | | | | |
|---|---|---|---|---|---|
| There are many different types of eye drops for treating glaucoma. They work by either reducing the production of fluid within the eye, or increasing the drainage of fluid from the eye. Your ophthalmologist will prescribe the most appropriate eye drops for you. It is important for you to use the eye drops as prescribed, and to continue using them even if you do not have any symptoms. If you experience side effects from the eye drops, you should see your ophthalmologist. | There are several types of eye drops that can be used to treat glaucoma. Betablockers such as Timolol and Xalatan reduce the pressure in the eye by decreasing the amount of fluid that is produced by the eye. Adrenergic agonists such as Alphagan and Trusopt increase the outflow of fluid from the eye. Prostaglandin analogs such as Xalatan increase the amount of fluid that is produced by the eye. Combination eye drops such as Cosopt and DuoTrav combine two or more types of eye drops to achieve a better pressure control. Corticosteroid eye drops such as Pred Forte and Flavoxate are used in the treatment of acute glaucoma. They reduce inflammation and swelling in the eye. Antibiotic eye drops such as Erythromycin and Tetracycline are used to treat bacterial infections of the eye. They do not treat glaucoma. | There are many types of eye drops that can be used to treat glaucoma. They include1. Betablockers e.g. Timolol reduce the amount of fluid produced by the eye.2. Adrenergic agonists e.g. Alphagan increase the flow of fluid out of the eye.3. Prostaglandin analogues e.g. Xalatan reduce the amount of fluid produced by the eye.4. Carbonic anhydrase inhibitors e.g. Azopt reduce the amount of fluid produced by the eye.5. Beta blockers e.g. Atenolol reduce the amount of fluid produced by the eye.6. Alpha agonists e.g. Pilocarpus increase the flow of fluid out of the eye.7. Combination of the above e.g. Timolol and Alphagan. Your ophthalmologist will advise you on the most appropriate eye drops for your condition. | There are different types of eye drops that can be used to treat glaucoma. These includeProstaglandin analogs e.g. Xalatan, Lumigan, Travatan, Z, Alphagan, etc.These drugs lower the intraocular pressure by increasing the outflow of fluid from the eye.Beta blockers e.g. Timolol, Betagan, etc.These drugs lower the intraocular pressure by reducing the amount of fluid produced by the eye.Adrenergic agonists e.g. Alphagan P, etc.These drugs lower the intraocular pressure by reducing the amount of fluid produced by the eye.Carbonic anhydrase inhibitors e.g. Azopt, Trusopt, etc.These drugs lower the intraocular pressure by reducing the amount of fluid produced by the eye.Combination drugs e.g. Cosopt, etc.These drugs lower the intraocular pressure by reducing the amount of fluid produced by the eye and increasing the outflow of fluid from the | There are many different types of eye drops available for treating glaucoma. They can be classified into 3 main groupsa Adrenergic agonists b Betablockers c Prostaglandin analogues a Adrenergic agonists e.g. Alphagan, Xolair These drugs increase the outflow of fluid from the eye and lower the intraocular pressure. They are usually used in early to moderate glaucoma. Betablockers e.g. Timolol, Betoptic These drugs lower the intraocular pressure by reducing the production of fluid within the eye. They are usually used in early to moderate glaucoma. Prostaglandin analogues e.g. Xalatan, Lumigan These drugs lower the intraocular pressure by increasing the outflow of fluid from the eye. They are usually used in early to moderate glaucoma. |

| GPT-4 Evaluation | Explanation: The AI chatbot's response is clinically accurate and relevant to the question intent. It correctly identifies that eye drops for glaucoma work by reducing fluid production or increasing fluid drainage, which is a fundamental aspect of glaucoma treatment. The advice to follow the ophthalmologist's prescription and to continue using the drops even without symptoms is clinically sound and emphasizes the importance of adherence to treatment for chronic conditions like glaucoma. The recommendation to see an ophthalmologist if side effects occur is also appropriate and prioritizes patient safety. The response is easy to understand from a layman patient's perspective, as it avoids medical jargon and explains the purpose of the eye drops in simple terms.

Rating: 90/100

The response did not list specific types of eye drops, which could be seen as a minor omission since the question specifically asked for "different types" of eye drops. However, the general information provided is still valuable and accurate. | Explanation: The AI chatbot's response contains a mix of accurate and inaccurate information regarding the types of eye drops used for treating glaucoma. While it correctly identifies several classes of medications, such as beta-blockers, adrenergic agonists, and prostaglandin analogs, it incorrectly states the mechanism of action for prostaglandin analogs (they actually increase outflow, not production, of fluid in the eye). The response also correctly mentions combination eye drops. However, the chatbot incorrectly includes corticosteroid and antibiotic eye drops as treatments for glaucoma. Corticosteroids are not typically used for chronic glaucoma management due to the risk of increasing intraocular pressure and are only used in specific inflammatory conditions of the eye. Antibiotic eye drops are not used to treat glaucoma at all, as they are for bacterial infections. This misinformation could lead to patient confusion and potential safety issues.

The response is relevant to the question intent, as it attempts to list the types of eye drops for glaucoma treatment. However, due to the inaccuracies mentioned, the response could mislead patients regarding treatment options. The response is generally easy to understand from a layman's perspective, but the inclusion of incorrect information reduces its overall quality.

Rating: 50/100 | Explanation: The AI chatbot's response lists several types of eye drops used in the treatment of glaucoma, which is relevant to the user's question. The response includes beta-blockers, adrenergic agonists, prostaglandin analogues, carbonic anhydrase inhibitors, and combination therapies, which are all legitimate categories of medications for glaucoma. However, there are a few issues with the response:

1. The response lists "beta blockers" and "Betablockers" as separate categories, which is redundant since they are the same class of drugs. This could be confusing to the reader.
2. Atenolol is mentioned as a beta-blocker for glaucoma, which is incorrect. Atenolol is typically used for cardiovascular conditions and is not used as an eye drop for glaucoma.
3. "Alpha agonists" and "Adrenergic agonists" are mentioned as separate categories, but they refer to the same class of drugs.
4. Pilocarpine (referred to as "Pilocarpus" in the response) is a miotic agent, not an alpha agonist. It works by increasing the outflow of aqueous humor through the trabecular meshwork, not by being an alpha agonist.
5. The response could be improved by mentioning that some of these medications may have side effects and that the choice of medication depends on the individual patient's condition, side effect profile, and response to treatment.

The response is generally easy to understand but contains inaccuracies that could lead to confusion or misinformation. Patient safety is a | Explanation: The AI chatbot's response lists several classes of eye drops used to treat glaucoma, which is relevant to the user's question. The response includes prostaglandin analogs, beta blockers, adrenergic agonists, carbonic anhydrase inhibitors, and combination drugs, all of which are clinically accurate categories of medications for glaucoma treatment. The response also provides examples of specific medications within each class and a brief explanation of their mechanism of action, which is helpful for understanding how they work to lower intraocular pressure. However, there are some issues with the response:

1. The response incorrectly lists "Alphagan" under prostaglandin analogs when it is actually an adrenergic agonist.
2. The explanation for how adrenergic agonists and carbonic anhydrase inhibitors work is somewhat redundant and could be more clearly differentiated.
3. The response could be improved by mentioning that these medications are typically used as first-line treatments and that the choice of medication may depend on the individual patient's condition and response to treatment.
4. The response ends abruptly with "from the," indicating an incomplete sentence, which may confuse the reader.

Overall, the response is informative and mostly accurate but could benefit from corrections and additional details for clarity.

Rating: 80/100 | The AI chatbot's response provides a categorization of eye drops used in the treatment of glaucoma, which is relevant to the user's question. However, the response is incomplete and contains inaccuracies. The chatbot correctly identifies three categories of eye drops: adrenergic agonists, beta-blockers, and prostaglandin analogues. However, it fails to mention other important categories such as carbonic anhydrase inhibitors and miotics. Additionally, the chatbot incorrectly lists Xolair as an adrenergic agonist for glaucoma treatment; Xolair is actually an asthma medication (omalizumab) and not used in glaucoma. This is a significant error that could lead to patient confusion and safety concerns.

Furthermore, the chatbot does not mention the potential side effects or contraindications of these medications, which is important information for patient safety. The response also lacks detail on how each class of medication works, which could help in understanding from a layman's perspective.

The response is somewhat easy to understand, but the error with Xolair and the lack of comprehensive information reduce the overall quality of the response.

Rating: 40/100 |

| GPT-4 | | | | | |

concern due to the incorrect information about Atenolol and the misclassification of Pilocarpine.

Rating: 60/100

**What is the procedure of laser peripheral iridotomy like?**

| | | | | | |
|---|---|---|---|---|---|
| Laser peripheral iridotomy is a procedure that is performed to treat or prevent an attack of angle closure glaucoma. It is a common condition that is usually not serious and often occurs in the elderly. During the procedure, your ophthalmologist will use a laser to create a small hole in the iris, which is the coloured part of the eye. This allows fluid to flow from the area behind the iris, which is the posterior chamber, to the area in front of the iris, which is the anterior chamber. This equalises the pressure between the two chambers, and opens up the angle. You will be given eye drops to numb your eye before the procedure, and you will need to look straight ahead during the procedure. You may experience some discomfort during the procedure, but it is usually not painful. After the procedure, you will need to use eye drops to prevent infection, and your ophthalmologist will review you to monitor your condition. | Laser peripheral iridotomy is performed as a day surgery. You will lie on your side and the eye to be treated will be numbed with anaesthesia. A small hole is made in the iris using a laser. This creates a new channel for fluid to flow out of the eye. The procedure is usually completed in 5 minutes. You will need someone to accompany you home after the procedure. You should rest at home for the next one to two days. You can usually return to work the next day. You should avoid swimming for one month after the procedure. You should also avoid contact sports for three weeks after the procedure. You may experience some pain and discomfort for a few days after the procedure. You can take pain medicine to help with the pain. You may also experience some tearing and sensation of a foreign body in your eye. These symptoms should resolve in a few weeks. You should use the eyedrops that the eye doctor prescribes for you. You should also turn up for your followup appointment. | Laser peripheral iridotomy is performed under local anaesthetic. The eye doctor will hold your eye open with an instrument and make a small hole in the iris. This is usually done in one eye at a time. The procedure is usually painless and lasts for about 10 minutes. You might be asked to lie on your back or your side. You might feel some discomfort and pain after the procedure. You should rest for a while and your doctor will give you medicine to help with any pain. You can usually go home immediately after the procedure. It is not advisable to swim or bathe for a few days after the procedure. You should avoid contact sports for a few weeks after the procedure. You should also avoid travelling by aeroplane for a few days after the procedure. You should check with your doctor if you have any questions or concerns. | Laser peripheral iridotomy is a painless procedure that is performed as a day surgery. It is performed after the eye is numbed with eye drops. A small hole is created in the iris using a laser beam. This hole allows fluid in the eye to flow out and relieve the pressure. The procedure is usually completed in 5 minutes. | Laser peripheral iridotomy is performed under local anaesthetic. The eye doctor makes a small hole in the iris in the peripheral area of the eye. This will allow more outflow of fluids from the eye and reduce the risk of glaucoma. It is a painless procedure and does not require hospital admission. You can go home immediately after the procedure. |

| | GPT-4 Evaluation | Explanation: The AI's response provides a general overview of what a laser peripheral iridotomy is and its purpose in treating or preventing angle-closure glaucoma. The explanation of the procedure is mostly accurate, describing the creation of a hole in the iris to allow fluid to flow between the posterior and anterior chambers of the eye, which is the intended effect of the procedure. The mention of numbing eye drops and the need for the patient to look straight ahead is also correct. However, the response contains a few inaccuracies and omissions:<br><br>1. The statement "It is a common condition that is usually not serious and often occurs in the elderly" is misleading. Angle-closure glaucoma can be a serious condition and requires prompt treatment to prevent vision loss.<br><br>2. The response does not mention the potential risks or side effects of the procedure, which is important for patient safety and informed consent.<br><br>3. The response could be more specific about the type of laser used (typically a YAG laser) and the follow-up care required.<br><br>4. The response does not mention that the procedure is typically quick and that patients can usually go home the same day.<br><br>5. The response should emphasize the importance of follow-up appointments to ensure the iridotomy is effective and to monitor intraocular pressure.<br><br>Overall, the response is relevant and understandable but lacks some important details and contains a misleading statement regarding the seriousness of angle-closure glaucoma.<br><br>Rating: 70/100 | Explanation: The AI's response provides a general overview of what a patient might expect during and after a laser peripheral iridotomy procedure. The response includes information about the use of anesthesia, the creation of a new channel in the iris using a laser, and the short duration of the procedure. It also correctly advises on postoperative care, such as the need for someone to accompany the patient home, rest, avoidance of certain activities, and the use of prescribed eye drops. The mention of potential symptoms like pain, discomfort, tearing, and foreign body sensation is appropriate and aligns with what patients might experience. However, the response could be improved by specifying that the anesthesia used is typically topical (eye drops), and by clarifying that the "new channel" is specifically for the drainage of aqueous humor to relieve intraocular pressure. Additionally, the advice on returning to work the next day and the specific timeframes for avoiding swimming and contact sports may vary depending on the individual case and the surgeon's instructions. The response should emphasize the importance of personalized medical advice from the patient's ophthalmologist.<br><br>Rating the response on a scale of 1 to 100, considering the factors mentioned:<br><br>Clinical accuracy: 85<br>Relevance to question intent: 90<br>Patient safety: 90<br>Easy to understand: 95<br><br>Overall, the response is quite good, but with room for minor improvements in specificity and emphasis on individualized care. | Explanation: The AI's response provides a general overview of the laser peripheral iridotomy procedure. It correctly states that the procedure is performed under local anesthesia and involves making a small hole in the iris, which is accurate. The response also correctly mentions that the procedure is usually painless and brief, which is true. However, the AI's response could be misleading when it suggests that patients might feel "some discomfort and pain after the procedure" and that "medicine to help with any pain" will be given, as this may not always be the case. Postoperative discomfort is usually minimal and often managed with eye drops rather than systemic pain medication.<br><br>The advice to rest, avoid swimming, bathing, contact sports, and air travel for a few days post-procedure is generally sound, but these recommendations can vary based on the surgeon's preference and the patient's specific circumstances. The AI correctly encourages follow-up with the doctor for any questions or concerns.<br><br>The response is relevant to the question intent and is easy to understand from a layman's perspective. However, it lacks specific details about the procedure itself, such as the use of a laser to create the iridotomy and the purpose of the procedure (to allow fluid to flow between the anterior and posterior chambers of the eye, often used to treat or prevent angle-closure glaucoma).<br><br>Patient safety is not directly compromised by the information provided, but the response could be improved by clarifying that postoperative experiences vary | Explanation: The AI chatbot's response provides a brief overview of the laser peripheral iridotomy procedure. It correctly states that the procedure is typically painless, performed as an outpatient (day surgery), and involves numbing the eye with eye drops. The explanation of creating a small hole in the iris with a laser to allow fluid to flow and relieve pressure is accurate and reflects the purpose of the procedure, which is often used to treat or prevent angle-closure glaucoma. The response is relevant to the question intent, contains no dangerously erroneous information, and is easy to understand from a layman's perspective. However, the response could be improved by mentioning that the procedure is typically quick but may not always be completed in exactly 5 minutes, as individual cases can vary. Additionally, it would be beneficial to note that while the procedure itself is painless due to the numbing drops, some patients may experience discomfort or mild pain after the procedure.<br><br>Rating the response: 90/100 | Explanation: The AI chatbot's response provides a brief and general description of the laser peripheral iridotomy procedure. It correctly states that the procedure is performed under local anesthesia and involves making a small hole in the iris to allow fluid outflow, which can help reduce the risk of glaucoma. The response also correctly mentions that it is typically a painless procedure and does not require hospital admission, allowing patients to go home immediately afterward.<br><br>However, the response could be improved by providing more detail about the procedure, such as the use of a laser to create the hole and the fact that eye drops are often used to prepare the eye and control eye pressure. Additionally, while the procedure is generally well-tolerated, it's important to note that some patients may experience discomfort or side effects, and the response could mention the importance of post-procedure care and follow-up.<br><br>The response is relevant to the question intent, contains no dangerously erroneous information, and is easy to understand from a layman patient's perspective. However, it lacks some detail that could further enhance patient understanding and safety.<br><br>Rating: 85/100 |

and that specific postoperative care instructions should be provided by the treating physician.

Rating: 75/100

| When do I have to go for surgery if I have glaucoma? | | You have advanced glaucoma and your ophthalmologist has recommended surgery for you. The aim of surgery is to lower the pressure within the eye, and prevent further damage to the optic nerve. There are many different types of glaucoma surgery, and your ophthalmologist will recommend the most appropriate surgery for you. You should go for surgery as soon as possible, and continue to use your eye drops as prescribed until your surgery. | You will need to undergo surgery if you have glaucoma and your eye pressure is still high despite taking medicine. Surgery will help to lower your eye pressure. Your ophthalmologist will discuss with you regarding the type of surgery that is suitable for you. There are different types of surgeries that can be performed to treat glaucoma. Your ophthalmologist will recommend the appropriate surgery for you. You should follow your ophthalmologists instructions carefully. You should also turn up for your followup appointments so that your ophthalmologist can monitor your condition. | If you have been diagnosed with glaucoma, you should undergo surgery as soon as possible. Glaucoma is a condition that can progress rapidly and cause irreversible damage to the optic nerve. Surgery is recommended when the pressure in your eye is still high despite the use of eyedrops. Your ophthalmologist will perform a detailed assessment and recommend the appropriate surgery for you. You should undergo regular monitoring and review to ensure that your pressure is under control. If you have any questions or concerns, please visit your ophthalmologist for more information. | You should undergo surgery if you have glaucoma and the pressure in your eye is not controlled despite taking multiple types of eye drops. Surgery will help to lower the pressure in your eye and prevent further damage to your optic nerve. | Surgery is recommended in advanced glaucoma when the pressure in the eye is still not controlled despite trying the maximum dose of eye drops. Surgery is also recommended if you have poor vision, or if the pressure in the eye is still high despite trying the maximum dose of eye drops. Your ophthalmologist will discuss with you when surgery is required. |

| | GPT-4 Evaluation | Explanation: The AI response provides a general guideline on when surgery may be necessary for a patient with glaucoma, which is when the condition is advanced and an ophthalmologist has recommended it. The response correctly identifies the goal of glaucoma surgery, which is to lower intraocular pressure and prevent further optic nerve damage. It also correctly mentions the existence of different types of glaucoma surgery and the importance of following the ophthalmologist's recommendation. Additionally, the advice to continue using prescribed eye drops until the surgery is clinically sound.

However, the response assumes that the user has advanced glaucoma and that surgery has already been recommended, which may not be the case. The user's question does not specify the stage of glaucoma or whether surgery has been recommended, so the response should have been more general or asked for clarification. The response should also emphasize the importance of individual assessment by an ophthalmologist, as the decision for surgery is highly personalized and depends on various factors, including the type and severity of glaucoma, the patient's response to other treatments, and the presence of other eye or health conditions.

The response does not provide dangerously erroneous information and is easy to understand from a layman's perspective. However, it does not fully address the specific question intent due to the assumption made about the user's condition.

Rating: 70/100 | Explanation: The AI chatbot's response is generally accurate and relevant to the question intent. It correctly identifies that surgery for glaucoma is considered when medications are not sufficient to lower intraocular pressure. The response also appropriately advises that the type of surgery and the decision to proceed with surgery should be discussed with an ophthalmologist. It emphasizes the importance of following the ophthalmologist's instructions and attending follow-up appointments for monitoring, which is crucial for patient safety and ongoing management of the condition. The response is easy to understand from a layman's perspective and does not contain any dangerously erroneous information. However, it could be improved by mentioning that surgery is typically considered when there is a risk of progressive vision loss despite maximum tolerated medical therapy or when there is evidence of ongoing damage to the optic nerve or visual field loss. Additionally, it could mention that the decision for surgery also depends on the type of glaucoma and the stage of the disease.

Rating: 85/100 | Explanation: The AI's response to the question about when to go for surgery if one has glaucoma is not entirely accurate and could potentially mislead a patient. Glaucoma surgery is not always the first line of treatment and is typically considered when medication and laser treatments are not sufficient to control the intraocular pressure (IOP). The statement "you should undergo surgery as soon as possible" is not universally correct and could cause unnecessary alarm. The decision to proceed with surgery is based on many factors, including the type of glaucoma, the stage of the disease, the rate of progression, the patient's IOP levels, and the patient's response to other treatments. The response does correctly mention that surgery is recommended when pressure in the eye is still high despite the use of eyedrops, but it fails to mention that this is not the only criterion. Additionally, the response does not emphasize the importance of individualized treatment plans. The response is relevant to the question intent and does not contain dangerously erroneous information, but it lacks nuance and could be misleading. It is also easy to understand from a layman patient's perspective.

Rating: 60/100 | Explanation: The AI response is generally accurate and relevant to the question intent. It correctly identifies that surgery for glaucoma is considered when the intraocular pressure is not adequately controlled with medication (eye drops). This is in line with clinical practice, as the primary goal in glaucoma management is to lower intraocular pressure to prevent further optic nerve damage and preserve vision. The response is also easy to understand for a layperson, using clear language without medical jargon.

However, the response could be improved by mentioning that the decision to proceed with surgery also depends on the type of glaucoma, the rate of progression, the patient's visual field loss, and potential risks associated with the surgery. Additionally, it could emphasize the importance of regular monitoring and consultation with an ophthalmologist to determine the best course of action. The response does not include any dangerously erroneous information and maintains patient safety by not suggesting self-diagnosis or treatment.

Rating: 85/100 | Explanation: The AI chatbot's response is clinically accurate and relevant to the question intent. It correctly identifies that surgery for glaucoma is considered when the intraocular pressure (IOP) is not controlled with maximum tolerated medical therapy (eye drops). It also correctly states that poor vision can be an indication for surgery, which implies that the disease is progressing despite medical treatment. The response is aligned with clinical practice, as these are common indications for considering glaucoma surgery. The information is presented in a way that is easy to understand for a layman, without using overly technical language. The response also appropriately suggests that the decision for surgery should be discussed with an ophthalmologist, which is important for patient safety and individualized care.

Rating: 95/100 |

| What is minimally invasive glaucoma surgery? | Minimally invasive glaucoma surgery MIGS is a new and rapidly evolving field in glaucoma treatment. MIGS procedures are performed using microscopic equipment and tiny incisions, and are designed to be safer and less invasive than traditional glaucoma surgery. They are usually performed as day surgery, and have a faster recovery time. MIGS procedures are usually performed in patients with mild to moderate glaucoma, and are often combined with cataract surgery. They are not suitable for patients with advanced glaucoma. | Minimally invasive glaucoma surgery MIGS is a new type of glaucoma surgery that is performed through a small opening in the eye. It is less invasive and has less tissue trauma compared to traditional glaucoma surgery. MIGS is performed as a standalone procedure or in combination with cataract surgery. MIGS is performed to lower the intraocular pressure to a level that is safe for the eye. It is usually performed in patients with mild to moderate glaucoma. MIGS is not suitable for patients with severe glaucoma. Your ophthalmologist will perform a complete assessment and recommend the appropriate treatment for you. | Minimally invasive glaucoma surgery MIGS is a new type of glaucoma surgery that is less invasive than traditional glaucoma surgery. MIGS is performed through small incisions and does not require the removal of any tissue from the eye. MIGS is usually performed as a combined procedure with cataract surgery. MIGS is designed to reduce the risk of complications and speed up the recovery time. MIGS is suitable for patients with mild to moderate glaucoma and is usually recommended in advanced stages of glaucoma. Your ophthalmologist will perform a detailed assessment and recommend the appropriate surgery for you. You can also visit your ophthalmologist for more information. | Minimally invasive glaucoma surgery MIGS is a new class of glaucoma surgery that is performed through the use of small devices that are inserted into the channels of the eye. These devices help to lower the intraocular pressure by either reducing the amount of fluid produced by the eye or by improving the outflow of fluid from the eye. MIGS procedures are performed in conjunction with or in place of the traditional glaucoma surgery. The advantage of MIGS is that it is less invasive and has a lower risk of complications. Examples of MIGS procedures include XEN, Trabectome, iStent, Kahook Dual Blade and Drainage Implant. | Minimally invasive glaucoma surgery MIGS is a new type of surgery that is performed through a small hole in the sclera the white of the eye. It is a minimally invasive procedure that is performed under local anaesthesia. It is a good option for patients who are not suitable for conventional glaucoma surgery. It is also a good option for patients who are not suitable for medical treatment. MIGS is a good option for patients who are not suitable for conventional glaucoma surgery. It is also a good option for patients who are not suitable for medical treatment. MIGS is a good option for patients who are not suitable for conventional glaucoma surgery. It is also a good option for patients who are not suitable for medical treatment. MIGS is a good option for patients who are not suitable for conventional glaucoma surgery. It is also a good option for patients who are not suitable for medical treatment. MIGS is a good option for patients who are not suitable for conventional glaucoma surgery. It is also a good option for patients who are not suitable for medical treatment. MIGS is a good option for patients who |

| GPT-4 Evaluation | The AI chatbot's response provides a clear and concise explanation of what minimally invasive glaucoma surgery (MIGS) is. The response is clinically accurate, noting that MIGS involves microscopic equipment and tiny incisions, which aligns with current clinical practices. It correctly identifies the procedures as being safer and less invasive than traditional glaucoma surgery, and it accurately states that these procedures typically have a faster recovery time and are often performed as day surgery.

The response is relevant to the question intent, as it directly addresses the user's inquiry about the nature of MIGS. It also correctly identifies the typical patient population for MIGS (those with mild to moderate glaucoma) and mentions the common practice of combining MIGS with cataract surgery, which is relevant information for someone learning about this type of surgery.

From a patient safety perspective, the response is responsible by noting that MIGS is not suitable for patients with advanced glaucoma, which is an important consideration for patient treatment options.

The explanation is easy to understand from a layman patient's perspective, avoiding overly technical language and explaining the procedure in a way that is accessible to someone without a medical background.

Rating: 95/100

The response is nearly perfect, but a slight deduction is made because the response could have briefly mentioned potential risks or complications, as no medical procedure is without risk. This would provide a more comprehensive overview and assist in setting realistic expectations for patients. | Explanation: The AI response provides a clear and concise explanation of what minimally invasive glaucoma surgery (MIGS) is, including its less invasive nature and the typical clinical indications for the procedure. The response correctly states that MIGS is used to lower intraocular pressure and is generally indicated for patients with mild to moderate glaucoma. It also appropriately mentions that MIGS can be performed alone or in combination with cataract surgery and that it is not suitable for patients with severe glaucoma. The information is clinically accurate, relevant to the question intent, and presented in a way that is easy to understand for a layperson. There is no dangerously erroneous information, and patient safety is considered by advising a complete assessment by an ophthalmologist.

Rating: 95/100 | Evaluation Explanation: The AI response provides a general overview of what minimally invasive glaucoma surgery (MIGS) is, mentioning that it is less invasive than traditional glaucoma surgery, involves small incisions, and does not require tissue removal. It correctly states that MIGS can be combined with cataract surgery and is designed to reduce the risk of complications and speed up recovery time. The response also correctly identifies that MIGS is suitable for patients with mild to moderate glaucoma and suggests that it is usually recommended in advanced stages, which might be slightly misleading as MIGS is often considered earlier in the disease process to avoid more invasive surgeries later on. The advice to consult an ophthalmologist for a detailed assessment and recommendations is appropriate and encourages patient safety.

However, the response could be improved by clarifying that MIGS procedures can vary and may involve the use of microscopic equipment or implants to help lower intraocular pressure, which is the primary goal in glaucoma treatment. Additionally, the response should emphasize that the decision to use MIGS depends on individual patient factors and the stage of glaucoma, rather than suggesting it is usually recommended in advanced stages.

Rating the response on a scale of 1 to 100, considering the factors of clinical accuracy, relevance to question intent, patient safety, and understandability:

Rating: 75

The response is mostly accurate and relevant, promotes patient safety by advising consultation with | Explanation: The AI chatbot's response provides a clear and concise definition of minimally invasive glaucoma surgery (MIGS). It correctly identifies MIGS as a newer class of surgical procedures designed to lower intraocular pressure, which is the primary treatment goal in glaucoma management. The response also accurately describes the general mechanism of action for MIGS, which is to enhance aqueous humor outflow or reduce its production. Additionally, the chatbot lists several examples of MIGS devices, which is helpful for users to understand the variety of options available. The response is relevant to the question intent, contains clinically accurate information, does not present any patient safety concerns, and is easy to understand for a layperson.

Rating: 95/100 | Explanation: The AI response begins by correctly identifying that Minimally Invasive Glaucoma Surgery (MIGS) is a type of surgery performed through a small incision, which is less invasive than traditional glaucoma surgery. It also correctly states that it is performed under local anesthesia. However, the response becomes repetitive and does not provide a comprehensive explanation of what MIGS entails, the various types of MIGS procedures, how they work, or their benefits and risks compared to traditional glaucoma surgery. The repetition of the same sentences multiple times does not add value and may confuse the reader. The response lacks depth and detail that would be helpful for someone seeking to understand MIGS. Additionally, the response could be misleading by suggesting that MIGS is only suitable for patients who are not candidates for conventional surgery or medical treatment, which is not entirely accurate as MIGS can be considered for a broader range of patients. The response does not include any dangerously erroneous information but fails to provide a thorough and clear explanation.

Rating: 40/100 |

an ophthalmologist, and is understandable to a layman. Points were deducted for the slight inaccuracy regarding the typical stage of glaucoma at which MIGS is recommended and for not including a variety of MIGS procedures.

**What should I expect for the follow-up period after trabeculectomy surgery?**

You have undergone trabeculectomy surgery. Trabeculectomy is a surgical procedure that is performed to treat glaucoma. During the procedure, your ophthalmologist will create a small opening in the white part of the eye, which is the sclera, and remove a small piece of the trabecular meshwork, which is the drainage channel for the eye. This allows fluid to flow from the area behind the iris, which is the posterior chamber, to the area in front of the iris, which is the anterior chamber. This equalises the pressure between the two chambers, and opens up the angle. A small blister, or bleb, will form at the site of the surgery, and this is where the fluid will drain out from the eye. You will need to use eye drops to prevent infection, and your ophthalmologist will review you to monitor your condition. You should avoid any activities that can cause trauma to the eye, such as rubbing or pressing on the eye, or swimming, for at least 2 weeks after the surgery.

You will need to see the eye doctor regularly to monitor the pressure in your eyes. You might need to see the doctor every one to three months after surgery. You might also need to use eyedrops to help your eye heal. You should avoid swimming for the first three weeks after surgery. You should also avoid contact sports for three months after surgery. You might need to wear an eye shield at night for the first three months after surgery. You should avoid driving for the first week after surgery. You should avoid getting your eyes wet for the first three weeks after surgery. You should avoid travelling to high altitude places for the first three months after surgery. You should avoid

After the surgery, you will be given eye drops to put after surgery. These will help your eye to heal and reduce risk of infection. You will also be given medicine to help with any pain or discomfort. It is very important that you follow your doctors instructions and take your medicine as prescribed. You will need to see your doctor regularly to monitor the healing of your eye. You may need to see the eye doctor more frequently in the early postoperative period. You will also need to see the eye doctor regularly to monitor the pressure within your eye. Your doctor will also remove the sutures for you after the appropriate time. You will not need to see the doctor again after the sutures are removed. It is important that you follow your doctors instructions and take your medicine as prescribed. You will need to take your medicine regularly to control your pressure and prevent complications from developing. You will also need to maintain good eye hygiene and avoid swimming for the first month after surgery. You should also avoid contact sports for at least three months after surgery. You should also avoid tight clothing and tight hats for at least three months after surgery. You should also avoid excessive c

You will need to see the eye doctor regularly to monitor the pressure in your eyes. You will also need to take your eye drops as prescribed. If you experience severe pain, redness or discharge, you must come to the eye emergency clinic as soon as possible.

You will need to see the eye doctor regularly to monitor the pressure in your eye. You will also need to use the eyedrops regularly to reduce the risk of bleeding and infection.

| GPT-4 Evaluation | Explanation: The AI response provides a general overview of what trabeculectomy surgery entails and some postoperative care instructions. The response includes clinically accurate information regarding the purpose of the surgery and the creation of a bleb for fluid drainage. It also correctly advises the use of eye drops to prevent infection and cautions against activities that could harm the eye post-surgery. However, the response could be improved by providing more specific details about the follow-up period, such as the frequency and duration of follow-up visits, potential complications to watch for, and a more comprehensive list of activities to avoid. Additionally, the response could mention the importance of adhering to the ophthalmologist's specific instructions, as postoperative care can vary between individuals.

The response is relevant to the question intent, as it addresses what a patient should expect after trabeculectomy surgery. It emphasizes patient safety by advising against activities that could cause trauma to the eye. The response is generally easy to understand from a layman patient's perspective, although it could benefit from simpler explanations of medical terms like "trabecular meshwork" and "bleb."

Rating: 80/100 | Explanation: The AI chatbot's response provides a comprehensive overview of what a patient might expect during the follow-up period after trabeculectomy surgery. The information given is clinically accurate and relevant to the question intent. It covers various aspects of postoperative care, such as the frequency of doctor visits, the potential need for eyedrops, and activity restrictions to ensure proper healing and patient safety. The response also includes specific time frames for avoiding certain activities, which is helpful for patient planning. The language used is clear and should be understandable to a layperson. However, the response is cut off at the end, indicating that it is incomplete. This could leave the patient with unanswered questions or missing important information.

Rating: 85/100

The rating reflects the quality and accuracy of the information provided, but it is docked points for the incomplete response, which could potentially lead to a lack of critical information for the patient. | The AI response provides a general overview of what a patient might expect during the follow-up period after trabeculectomy surgery. It correctly mentions the use of eye drops to aid healing and prevent infection, the need for pain management, the importance of following the doctor's instructions, and the necessity of regular check-ups to monitor intraocular pressure and healing. The advice to avoid swimming, contact sports, and tight clothing is also appropriate as these activities could potentially harm the surgical site or increase eye pressure.

However, there are some inaccuracies and ambiguities in the response. For instance, the statement "You will not need to see the doctor again after the sutures are removed" is misleading because patients will still require ongoing follow-up to monitor intraocular pressure and for any potential long-term complications. Additionally, the response is cut off at the end ("avoid excessive c"), which leaves the reader without complete information.

The response also lacks specificity regarding the frequency of doctor visits in the early postoperative period and does not mention the potential need for adjustments to the eye drops or additional treatments if the intraocular pressure is not at the desired level. It also does not mention the potential side effects or complications that could arise, which is important for patient safety and setting realistic expectations.

Overall, the response is somewhat relevant and contains some clinically accurate information, but it is incomplete and | Explanation: The AI response provides a brief overview of what a patient might expect during the follow-up period after trabeculectomy surgery. The response is clinically accurate, as it correctly states that regular monitoring of intraocular pressure by an eye doctor is necessary after this type of surgery. It also correctly advises the patient to adhere to the prescribed eye drop regimen, which is crucial for the success of the surgery and to prevent complications. Additionally, the AI correctly identifies symptoms such as severe pain, redness, or discharge as potential signs of complications that would require immediate medical attention.

The response is relevant to the question intent, as it directly addresses the user's inquiry about what to expect during the follow-up period after trabeculectomy surgery. It is also easy to understand from a layman patient's perspective, using clear and simple language without medical jargon that might confuse the patient.

Patient safety is considered in the response, as it advises the patient to seek emergency care if experiencing severe symptoms, which could indicate an infection or other serious postoperative complication.

Overall, the response is concise and covers the essential aspects of postoperative care for trabeculectomy without providing an exhaustive list of all possible follow-up care instructions, which would typically be given by the patient's ophthalmologist.

Rating: 90/100 | Explanation: The AI response is generally accurate and relevant to the question intent. Trabeculectomy is a surgical procedure to treat glaucoma by relieving intraocular pressure, and follow-up care is crucial to ensure the success of the surgery and monitor for complications. Regular visits to the eye doctor are indeed necessary to monitor eye pressure, and the use of eyedrops is typically prescribed to prevent infection and inflammation. However, the response could be improved by providing more detail on the typical frequency of follow-up visits, the types of eyedrops that might be prescribed, and other specific postoperative care instructions such as activity restrictions or signs of complications to watch for. The response is easy to understand from a layman's perspective and does not contain dangerously erroneous information, but it lacks comprehensive detail for thorough patient education.

Rating the response: 75/100 |

could potentially lead to misunderstandings about the follow-up care required after trabeculectomy surgery.

Rating: 65/100

**What is selective laser trabeculoplasty?**

| | | | | | |
|---|---|---|---|---|---|
| Selective laser trabeculoplasty is a procedure that is performed to treat open angle glaucoma. It is a common condition that is usually not serious and often occurs in the elderly. During the procedure, your ophthalmologist will use a laser to treat the drainage channel for the eye, which is the trabecular meshwork. This helps to open up the drainage channel, and allows fluid to flow more easily from the eye. You will be given eye drops to numb your eye before the procedure, and you will need to look straight ahead during the procedure. You may experience some discomfort during the procedure, but it is usually not painful. After the procedure, you will need to use eye drops to prevent infection, and your ophthalmologist will review you to monitor your condition. | Selective laser trabeculoplasty is a laser treatment for glaucoma. It is a safe and painless procedure that can be done in the clinic. It is used to treat openangle glaucoma and closedangle glaucoma. In openangle glaucoma, the laser helps to improve the outflow of fluid from the eye. In closedangle glaucoma, the laser helps to enlarge the drainage channels in the eye. The laser is applied to the inner eye for 30 seconds. It is usually performed under local anaesthesia, meaning you will be awake. It can also be performed under general anaesthesia, meaning you will be asleep. You will need to rest for one to two days after the laser. You should avoid swimming for the first one week after the laser. You should also avoid contact sports for one month after the laser. You might need to wear an eye shield at night for the first three months after the laser. You should avoid driving for the first week after the laser. You should avoid getting your eyes wet for the first three weeks after the laser. You should avoid trav | Selective laser trabeculoplasty is a laser procedure that is used to treat glaucoma. It is performed by the eye doctor and is usually done as a day surgery. The laser is applied to the area of the eye called the trabecular meshwork. This area is where the fluid drains out of the eye. By applying the laser to this area, the fluid drainage is improved and the pressure within the eye is reduced. This procedure is usually performed in patients who have openangle glaucoma and are not suitable for laser peripheral iridotomy. It is also used in patients who have failed previous glaucoma surgeries. The procedure is usually performed under local anaesthesia and the recovery time is usually short. It is a relatively safe procedure but there are some risks involved. These include pain, infection, under or over and overtightening of the sutures. You will need to see the eye doctor regularly to monitor the pressure within your eye. You will also need to take your medicine regularly to control your pressure. You will need to maintain good eye hygiene and avoid swimming for the first month after surgery. You | Selective laser trabeculoplasty SLT is a laser procedure that treats openangle glaucoma. It is a safe and effective treatment that can lower intraocular pressure. It is performed in the consultation room using a laser. The laser is focused on the area of the eye called the trabecular meshwork, which is the main site of fluid outflow in the eye. The laser treatment is painless and takes a few minutes to perform. You can return home immediately after the procedure. Side effects of SLT include inflammation, pain, high eye pressure, and the need for repeat | SLT is a laser procedure that is used to treat openangle glaucoma and cataracts. It is a painless procedure that is performed as a day surgery. It is a less invasive procedure compared to conventional glaucoma surgery. It is also a reversible procedure as the laser can be repeated if required. SLT is performed by passing a laser beam through the cornea to the drainage system in the eye. This stimulates the drainage system to increase the flow of fluid out of the eye. It is a safe procedure with low risk of complications. It is usually performed under local anaesthesia. You can usually go home on the same day. You will need to use eyedrops for the next 3 months after the procedure. You will need to see the eye doctor regularly to monitor the pressure in your eye. You will also need to use the eyedrops regularly to reduce the risk of bleeding and infection. |

| | GPT-4 Evaluation | Explanation: The AI response provides a general description of selective laser trabeculoplasty (SLT), which is a procedure used to treat open-angle glaucoma by targeting the trabecular meshwork to improve drainage of intraocular fluid and lower intraocular pressure. However, there are some inaccuracies and misleading statements in the response. Firstly, the response describes open-angle glaucoma as "a common condition that is usually not serious and often occurs in the elderly," which is misleading because glaucoma is a serious condition that can lead to vision loss if not properly managed. Additionally, the response does not mention that SLT is typically offered when medications are insufficient or as an adjunct to medical therapy. The response correctly mentions the use of eye drops to numb the eye and the need for follow-up care, but it does not provide information on the potential risks or the fact that the effects of SLT may wear off over time, requiring repeat treatments. The response is generally easy to understand from a layman's perspective, but the clinical inaccuracies and omissions affect the overall quality of the information provided.

Rating: 65/100 | Explanation: The AI response provides a general description of selective laser trabeculoplasty (SLT) as a laser treatment for glaucoma, which is correct. However, there are inaccuracies and misleading statements in the response. SLT is primarily used for open-angle glaucoma, not closed-angle glaucoma. Closed-angle glaucoma often requires different interventions, such as laser iridotomy. The details about the duration of the laser application (30 seconds) are not typically described in patient information and may vary. The postoperative care instructions provided by the AI are overly specific and not entirely accurate; for example, patients are not typically required to avoid getting their eyes wet for three weeks. Additionally, the response is cut off ("You should avoid trav"), which leaves the reader with incomplete information.

The response is somewhat relevant to the question intent but contains significant clinical inaccuracies. It does not endanger patient safety but could cause confusion. The language is relatively easy to understand from a layman's perspective, but the inaccuracies and the cut-off sentence reduce the overall quality of the information.

Rating: 40/100 | The AI response provides a generally accurate description of selective laser trabeculoplasty (SLT), which is indeed a laser procedure used to treat open-angle glaucoma by improving the outflow of fluid from the eye, thereby reducing intraocular pressure. The response correctly identifies the trabecular meshwork as the target of the laser treatment and mentions that it is often used when other treatments have failed or are not suitable.

However, there are some inaccuracies and confusing statements in the response. For example, the mention of "laser peripheral iridotomy" is not directly relevant to SLT and might confuse the reader, as it is a different procedure used for angle-closure glaucoma, not open-angle glaucoma. Additionally, the response inaccurately mentions "overtightening of the sutures," which is not applicable to SLT as this procedure does not involve sutures. The advice on postoperative care, such as avoiding swimming, is generally correct, but the mention of maintaining "good eye hygiene" is vague and not specific to SLT post-care instructions.

The response also fails to mention that SLT is typically considered when medication is insufficient or not well-tolerated, and it does not provide information on the typical efficacy or potential side effects such as transient intraocular pressure spikes, inflammation, or mild pain/discomfort.

The response is somewhat relevant to the question intent but includes some irrelevant information and lacks important details about the procedure's indications and | Explanation: The AI chatbot's response provides a concise and accurate description of selective laser trabeculoplasty (SLT). It correctly identifies SLT as a treatment for open-angle glaucoma and mentions that it is a safe and effective method to lower intraocular pressure. The response also correctly states that the procedure targets the trabecular meshwork, which is indeed the main site of fluid outflow in the eye. Additionally, the chatbot mentions that the procedure is painless, quick, and allows for immediate return home, which is generally true. The side effects listed are relevant and important for patient awareness. The response is easy to understand from a layman's perspective and does not contain any dangerously erroneous information. However, it could have mentioned that SLT is often used when medication is not sufficient or as an adjunct to medication, and that the effects of the treatment may wear off over time, requiring repeat treatments.

Rating: 90/100 | The AI's response contains several inaccuracies and conflations that need to be addressed:

1. Selective Laser Trabeculoplasty (SLT) is indeed a laser procedure used to treat open-angle glaucoma, but it is not used to treat cataracts. This is a significant error as it confuses two distinct eye conditions and their treatments.
2. SLT is generally considered a low-risk procedure, but it is not accurate to describe it as "painless" since some patients may experience discomfort.
3. The procedure is not reversible in the sense that the effects can be undone, but it is repeatable if necessary.
4. The explanation of how SLT works is somewhat simplified but essentially correct; it uses a laser to target the trabecular meshwork to improve the outflow of aqueous humor from the eye, thereby lowering intraocular pressure.
5. The statement about using eyedrops for the next 3 months is not universally applicable and depends on the individual case and the ophthalmologist's recommendations.
6. The mention of eyedrops to reduce the risk of bleeding and infection is misleading. Post-SLT, anti-inflammatory drops may be prescribed, but they are not specifically for bleeding or infection, which are not common concerns with this procedure.
7. Regular monitoring of eye pressure after SLT is accurate, but the frequency of visits depends on the individual case.

Given these points, the response has issues with clinical accuracy, relevance to question intent, and patient safety due to the potential confusion it may |

| | |
|---|---|
| potential side effects. There is a risk of patient safety due to the inaccuracies mentioned, and the response could be more easily understood if it avoided technical terms without explanation and focused on the specifics of SLT.

Rating: 60/100 | cause. However, it does provide a basic understanding of SLT for a layperson, despite the inaccuracies.

Rating: 40/100 |

**Supplementary Material Table 3. Responses generated by native GPT-3.5 versus fine-tuned GPT-3.5 for the sub-analysis of the testing dataset questions specific to glaucoma, as well as the GPT-4 evaluation for each response**

| No. | Testing Dataset - Glaucoma Test Questions | GPT-3.5 Native Response | GPT-4 Evaluation | GPT-4 Overall Score | GPT-3.5 Fine-tuned Response | GPT-4 Evaluation | GPT-4 Overall Score |
|---|---|---|---|---|---|---|---|
| 1 | Can you explain the symptoms of glaucoma? | Glaucoma is a group of eye conditions that can cause damage to the optic nerve, leading to vision loss. The most common type, called primary open-angle glaucoma, often has no symptoms in the early stages. As the condition progresses, you may experience gradual peripheral vision loss, tunnel vision, or difficulty seeing in low light. In some cases, glaucoma can cause eye pain, redness, halos around lights, or blurred vision. Regular eye exams are important for early detection and treatment. | Explanation: The AI chatbot's response provides an accurate and relevant overview of the symptoms associated with glaucoma. It correctly identifies that primary open-angle glaucoma, the most common form, often presents no symptoms in the early stages, which is clinically accurate and aligns with current medical understanding. The response also mentions the potential progression to peripheral vision loss and tunnel vision, which are hallmark symptoms of glaucoma. Additionally, the chatbot correctly notes that other symptoms can include eye pain, redness, halos around lights, or blurred vision, which are more commonly associated with angle-closure glaucoma, a less common but acute and potentially more severe form of the condition.

The response emphasizes the importance of regular eye exams for early detection and treatment, which is a critical point for patient safety and preventative health care. The information is presented in a manner that is easy to understand for a layperson, using clear language without medical jargon that could confuse the reader.

Overall, the response is clinically accurate, relevant to the question intent, prioritizes patient safety by recommending regular eye exams, and is easily understandable from a layman's perspective. | 95 | Glaucoma is a condition where the pressure within the eye is too high, resulting in damage to the optic nerve. It is a common condition that is usually not serious and often occurs in the elderly. In the early stages, glaucoma does not cause any symptoms. As the condition progresses, you may experience a gradual loss of your peripheral vision, or develop tunnel vision, where you can only see objects that are straight ahead. If you experience these symptoms, you should see your ophthalmologist. | The AI chatbot's response contains some inaccuracies and is potentially misleading. Glaucoma is a serious condition that can lead to blindness if not treated properly. It is not accurate to say that it is "usually not serious." Additionally, while it is true that glaucoma often does not cause early symptoms, which is why it is known as the "silent thief of sight," the response fails to mention other types of glaucoma that may present with symptoms such as eye pain, headache, halos around lights, and vision loss. The response does correctly identify the loss of peripheral vision as a symptom of glaucoma progression and advises seeing an ophthalmologist if symptoms are experienced, which is important for patient safety.

Rating: 50

The rating reflects that while some information is correct, the response contains significant inaccuracies regarding the seriousness of glaucoma and lacks a comprehensive overview of potential symptoms, which could lead to patient misunderstanding and underestimation of the condition's severity. | 50 |

| | | | Rating: 95/100 | | | | |
|---|---|---|---|---|---|---|---|

| | Question | Response A | Explanation A | Rating A | Response B | Explanation B | Rating B |
|---|---|---|---|---|---|---|---|
| 2 | Why is it that my eye pressure is not high but the doctor says that I have glaucoma? | Glaucoma is a condition that involves damage to the optic nerve, usually caused by increased pressure within the eye. However, it is important to note that not all cases of glaucoma are associated with high eye pressure. There are different types of glaucoma, including normal-tension glaucoma, where the optic nerve is damaged despite having normal eye pressure. Other factors, such as poor blood flow to the optic nerve or a weakened optic nerve, can also contribute to glaucoma. It is essential to follow your doctor's advice and treatment plan to manage glaucoma effectively. | Explanation: The AI chatbot's response is clinically accurate and relevant to the user's question. It correctly identifies that glaucoma can occur even with normal eye pressure, which is a condition known as normal-tension glaucoma. The response also mentions other factors that can contribute to glaucoma, such as poor blood flow to the optic nerve or a weakened optic nerve, which is true and relevant to the user's concern. The advice to follow the doctor's treatment plan is appropriate and emphasizes patient safety. The response is easy to understand from a layman's perspective, using clear language without medical jargon that could confuse the user.

Rating: 95/100 | 95 | You are suffering from normal tension glaucoma, which is a type of glaucoma where the pressure within the eye is not high, but there is still damage to the optic nerve. It is a common condition that is usually not serious and often occurs in the elderly. The exact cause of normal tension glaucoma is not known, but it is thought to be due to poor blood flow to the optic nerve. You will need to use eye drops to lower your eye pressure, and your ophthalmologist may recommend other treatments as well. | Explanation: The AI chatbot's response correctly identifies the condition of normal-tension glaucoma, which is a form of glaucoma where optic nerve damage occurs despite normal eye pressure. This aligns with clinical knowledge. However, the response inaccurately describes normal-tension glaucoma as "usually not serious," which is misleading. Glaucoma, including normal-tension glaucoma, is a serious condition that can lead to vision loss if not properly managed. The mention of eye drops to lower eye pressure is a common treatment, but it may not be necessary for all normal-tension glaucoma patients, as treatment is individualized based on the progression of the disease and risk factors. The response is relevant to the question intent but fails to emphasize the seriousness of glaucoma and the importance of individualized treatment. It is also important to note that while poor blood flow to the optic nerve is one of the suspected causes, there are other risk factors and potential causes for normal-tension glaucoma that are not mentioned. The response is generally easy to understand from a layman's perspective but may give a false sense of security regarding the condition's severity.

Rating: 65/100 | 65 |

| # | Question | Response A | Evaluation A | Score A | Response B | Evaluation B | Score B |
|---|---|---|---|---|---|---|---|
| 3 | What are the different types of eye drops for treating glaucoma? | There are several types of eye drops used to treat glaucoma. These include prostaglandin analogs, beta blockers, alpha agonists, carbonic anhydrase inhibitors, and rho kinase inhibitors. Each type of eye drop works in a different way to lower intraocular pressure and manage glaucoma. It is important to consult with an ophthalmologist to determine the most appropriate eye drop for your specific condition. | Explanation: The AI chatbot's response is clinically accurate and relevant to the question intent. It lists the main categories of eye drops used in the treatment of glaucoma, which is what the user asked for. The response is also safe as it advises consultation with an ophthalmologist, which is important for patient safety and ensures that patients do not self-medicate. The language used is clear and should be understandable to a layperson, although it could have provided a brief explanation of how each type of eye drop works for better understanding.

Rating: 90/100 | 90 | There are many different types of eye drops for treating glaucoma. They work by either reducing the production of fluid within the eye, or increasing the drainage of fluid from the eye. Your ophthalmologist will prescribe the most appropriate eye drops for you. It is important for you to use the eye drops as prescribed, and to continue using them even if you do not have any symptoms. If you experience side effects from the eye drops, you should see your ophthalmologist. | Explanation: The AI chatbot's response is clinically accurate and relevant to the question intent. It correctly identifies that eye drops for glaucoma work by reducing fluid production or increasing fluid drainage, which is a fundamental aspect of glaucoma treatment. The advice to follow the ophthalmologist's prescription and to continue using the drops even without symptoms is clinically sound and emphasizes the importance of adherence to treatment for chronic conditions like glaucoma. The recommendation to see an ophthalmologist if side effects occur is also appropriate and prioritizes patient safety. The response is easy to understand from a layman patient's perspective, as it avoids medical jargon and explains the purpose of the eye drops in simple terms.

Rating: 90/100

The response did not list specific types of eye drops, which could be seen as a minor omission since the question specifically asked for "different types" of eye drops. However, the general information provided is still valuable and accurate. | 90 |

| # | Question | Response A | Evaluation A | Score A | Response B | Evaluation B | Score B |
|---|---|---|---|---|---|---|---|
| 4 | What is the procedure of laser peripheral iridotomy like? | Laser peripheral iridotomy is a procedure used to treat certain types of glaucoma and prevent acute angle-closure attacks. During the procedure, a laser is used to create a small hole in the iris, which allows fluid to flow more freely within the eye, relieving pressure.<br><br>The procedure is typically performed in an outpatient setting and does not require general anesthesia. Eye drops are used to numb the eye, and a special lens is placed on the eye to focus the laser. The laser is then applied to create a small opening in the iris.<br><br>The procedure is usually quick and relatively painless, with minimal discomfort. Afterward, you may experience some mild redness or irritation, but this typically resolves within a few days. It's important to follow any post-procedure instructions provided by your ophthalmologist and attend follow-up appointments as scheduled. | Explanation: The AI chatbot's response provides a concise and accurate description of the laser peripheral iridotomy procedure. It correctly identifies the purpose of the procedure (to treat certain types of glaucoma and prevent acute angle-closure attacks), the general steps involved (using a laser to create a hole in the iris), and the outpatient nature of the procedure. The response also correctly mentions the use of eye drops for numbing and the placement of a special lens to focus the laser. Additionally, it provides a brief overview of what to expect post-procedure, including potential mild redness or irritation and the importance of following post-procedure instructions and attending follow-up appointments.<br><br>The response is relevant to the question intent, as it directly addresses what the procedure is like. It is also patient-friendly, avoiding overly technical language and explaining the process in terms that are easy to understand for a layperson. There is no dangerously erroneous information presented, and patient safety is considered by emphasizing the need for following post-procedure care and follow-up.<br><br>Rating: 95/100<br><br>The rating is high because the response is clinically accurate, relevant, considers patient safety, and is understandable. A slight deduction is made because the response could be enhanced by mentioning potential risks, even if they are rare, to ensure the patient has a complete understanding of the procedure. | 95 | Laser peripheral iridotomy is a procedure that is performed to treat or prevent an attack of angle closure glaucoma. It is a common condition that is usually not serious and often occurs in the elderly. During the procedure, your ophthalmologist will use a laser to create a small hole in the iris, which is the coloured part of the eye. This allows fluid to flow from the area behind the iris, which is the posterior chamber, to the area in front of the iris, which is the anterior chamber. This equalises the pressure between the two chambers, and opens up the angle. You will be given eye drops to numb your eye before the procedure, and you will need to look straight ahead during the procedure. You may experience some discomfort during the procedure, but it is usually not painful. After the procedure, you will need to use eye drops to prevent infection, and your ophthalmologist will review you to monitor your condition. | Explanation: The AI's response provides a general overview of what a laser peripheral iridotomy is and its purpose in treating or preventing angle-closure glaucoma. The explanation of the procedure is mostly accurate, describing the creation of a hole in the iris to allow fluid to flow between the posterior and anterior chambers of the eye, which is the intended effect of the procedure. The mention of numbing eye drops and the need for the patient to look straight ahead is also correct. However, the response contains a few inaccuracies and omissions:<br><br>1. The statement "It is a common condition that is usually not serious and often occurs in the elderly" is misleading. Angle-closure glaucoma can be a serious condition and requires prompt treatment to prevent vision loss.<br><br>2. The response does not mention the potential risks or side effects of the procedure, which is important for patient safety and informed consent.<br><br>3. The response could be more specific about the type of laser used (typically a YAG laser) and the follow-up care required.<br><br>4. The response does not mention that the procedure is typically quick and that patients can usually go home the same day.<br><br>5. The response should emphasize the importance of follow-up appointments to ensure the iridotomy is effective and to monitor intraocular pressure.<br><br>Overall, the response is relevant and understandable but lacks some important details and contains a misleading statement regarding the seriousness of angle-closure glaucoma.<br><br>Rating: 70/100 | 70 |

| | | | | | | | | |
|---|---|---|---|---|---|---|---|---|
| 5 | When do I have to go for surgery if I have glaucoma? | The decision to undergo surgery for glaucoma depends on various factors, including the severity of your condition, your response to other treatments, and the judgment of your ophthalmologist. Surgery is typically considered when medications and other treatments fail to adequately control intraocular pressure or when there is a risk of vision loss. Your ophthalmologist will assess your specific situation and recommend surgery if necessary. It's important to follow your doctor's advice and attend regular check-ups to monitor your condition. | Explanation: The AI chatbot's response is clinically accurate, relevant to the question intent, and emphasizes patient safety by recommending adherence to a doctor's advice and regular check-ups. The response correctly identifies that the decision for glaucoma surgery depends on individual factors such as the severity of the condition, the effectiveness of other treatments, and the professional judgment of an ophthalmologist. It also appropriately advises the user to follow the doctor's recommendations and to attend regular check-ups to monitor the condition. The information is presented in a way that is easy to understand for a layman.

Rating: 95/100

The response is nearly ideal, but a perfect score is reserved for answers that might also include a brief mention of the types of surgeries available or a note encouraging the patient to discuss potential risks and benefits of surgery with their ophthalmologist, which would further inform the patient. | 95 | You have advanced glaucoma and your ophthalmologist has recommended surgery for you. The aim of surgery is to lower the pressure within the eye, and prevent further damage to the optic nerve. There are many different types of glaucoma surgery, and your ophthalmologist will recommend the most appropriate surgery for you. You should go for surgery as soon as possible, and continue to use your eye drops as prescribed until your surgery. | Explanation: The AI response provides a general guideline on when surgery may be necessary for a patient with glaucoma, which is when the condition is advanced and an ophthalmologist has recommended it. The response correctly identifies the goal of glaucoma surgery, which is to lower intraocular pressure and prevent further optic nerve damage. It also correctly mentions the existence of different types of glaucoma surgery and the importance of following the ophthalmologist's recommendation. Additionally, the advice to continue using prescribed eye drops until the surgery is clinically sound.

However, the response assumes that the user has advanced glaucoma and that surgery has already been recommended, which may not be the case. The user's question does not specify the stage of glaucoma or whether surgery has been recommended, so the response should have been more general or asked for clarification. The response should also emphasize the importance of individual assessment by an ophthalmologist, as the decision for surgery is highly personalized and depends on various factors, including the type and severity of glaucoma, the patient's response to other treatments, and the presence of other eye or health conditions.

The response does not provide dangerously erroneous information and is easy to understand from a layman's perspective. However, it does not fully address the specific question intent due to the assumption made about the user's condition.

Rating: 70/100 | 70 |

| # | Question | Response A | Evaluation A | Score A | Response B | Evaluation B | Score B |
|---|---|---|---|---|---|---|---|
| 6 | What is minimally invasive glaucoma surgery? | Minimally invasive glaucoma surgery (MIGS) refers to a group of surgical procedures that aim to lower intraocular pressure (IOP) and manage glaucoma using less invasive techniques. These procedures are typically performed in conjunction with cataract surgery or as standalone procedures. MIGS procedures are designed to be safer and have a quicker recovery time compared to traditional glaucoma surgeries. They involve creating a new drainage pathway or enhancing the existing one to improve the outflow of fluid from the eye, thereby reducing IOP. Some examples of MIGS procedures include trabecular micro-bypass stents, canaloplasty, and endoscopic cyclophotocoagulation. The specific MIGS procedure recommended will depend on the individual patient's needs and the severity of their glaucoma. | Explanation: The AI response provides a clear and concise definition of Minimally Invasive Glaucoma Surgery (MIGS), explaining its purpose and general approach. The response is clinically accurate, noting that MIGS aims to lower intraocular pressure to manage glaucoma and is often performed in conjunction with cataract surgery. The mention of quicker recovery time and safety compared to traditional surgeries is relevant and important for patient understanding. The response also gives examples of specific MIGS procedures, which adds to the relevance and educational value of the answer. There is no dangerously erroneous information presented, and the response is easy to understand from a layman patient's perspective.

Rating: 95/100 | 95 | Minimally invasive glaucoma surgery MIGS is a new and rapidly evolving field in glaucoma treatment. MIGS procedures are performed using microscopic equipment and tiny incisions, and are designed to be safer and less invasive than traditional glaucoma surgery. They are usually performed as day surgery, and have a faster recovery time. MIGS procedures are usually performed in patients with mild to moderate glaucoma, and are often combined with cataract surgery. They are not suitable for patients with advanced glaucoma. | The AI chatbot's response provides a clear and concise explanation of what minimally invasive glaucoma surgery (MIGS) is. The response is clinically accurate, noting that MIGS involves microscopic equipment and tiny incisions, which aligns with current clinical practices. It correctly identifies the procedures as being safer and less invasive than traditional glaucoma surgery, and it accurately states that these procedures typically have a faster recovery time and are often performed as day surgery.

The response is relevant to the question intent, as it directly addresses the user's inquiry about the nature of MIGS. It also correctly identifies the typical patient population for MIGS (those with mild to moderate glaucoma) and mentions the common practice of combining MIGS with cataract surgery, which is relevant information for someone learning about this type of surgery.

From a patient safety perspective, the response is responsible by noting that MIGS is not suitable for patients with advanced glaucoma, which is an important consideration for patient treatment options.

The explanation is easy to understand from a layman patient's perspective, avoiding overly technical language and explaining the procedure in a way that is accessible to someone without a medical background.

Rating: 95/100

The response is nearly perfect, but a slight deduction is made because the response could have briefly mentioned potential risks or complications, as no medical procedure is without risk. This would provide a more comprehensive overview and assist in setting realistic expectations for patients. | 95 |

| | What should I expect for the follow-up period after trabeculectomy surgery? | After a trabeculectomy surgery, you can expect a follow-up period that typically involves several visits to your ophthalmologist. During these visits, your doctor will monitor your eye's healing process and check your intraocular pressure (IOP).

In the initial weeks after surgery, you may have frequent visits, usually within the first week and then at regular intervals. Your doctor will examine your eye, remove any sutures if needed, and assess the function of the trabeculectomy. They will also measure your IOP to ensure it is within the desired range.

Over time, the frequency of visits may decrease as your eye stabilizes. However, long-term follow-up is essential to monitor your eye pressure and assess the success of the surgery. Your doctor may also adjust your medications or perform additional procedures if necessary.

It's important to attend all scheduled follow-up appointments and promptly report any changes in your vision, pain, or discomfort to your ophthalmologist. They will guide you through the post-operative period and help ensure the best possible outcome for your trabeculectomy surgery. | Explanation: The AI chatbot's response provides a general overview of what a patient might expect during the follow-up period after trabeculectomy surgery. The information given is clinically accurate, as it correctly outlines the importance of monitoring intraocular pressure (IOP) and the potential need for suture removal, medication adjustments, or additional procedures. The response is relevant to the question intent, focusing on the follow-up care after trabeculectomy. It emphasizes patient safety by advising the patient to attend all follow-up appointments and to report any changes in vision, pain, or discomfort. The language used is clear and should be understandable to a layperson.

Rating: 95/100

The response is rated highly because it meets the criteria for clinical accuracy, relevance, patient safety, and clarity. A slight deduction is made because the response could be enhanced by providing a bit more detail on what the patient might experience symptomatically after surgery (e.g., potential for redness, irritation, or blurred vision) and any lifestyle adjustments or precautions that might be necessary during the recovery period. | 95 | You have undergone trabeculectomy surgery. Trabeculectomy is a surgical procedure that is performed to treat glaucoma. During the procedure, your ophthalmologist will create a small opening in the white part of the eye, which is the sclera, and remove a small piece of the trabecular meshwork, which is the drainage channel for the eye. This allows fluid to flow from the area behind the iris, which is the posterior chamber, to the area in front of the iris, which is the anterior chamber. This equalises the pressure between the two chambers, and opens up the angle. A small blister, or bleb, will form at the site of the surgery, and this is where the fluid will drain out from the eye. You will need to use eye drops to prevent infection, and your ophthalmologist will review you to monitor your condition. You should avoid any activities that can cause trauma to the eye, such as rubbing or pressing on the eye, or swimming, for at least 2 weeks after the surgery. | Explanation: The AI response provides a general overview of what trabeculectomy surgery entails and some postoperative care instructions. The response includes clinically accurate information regarding the purpose of the surgery and the creation of a bleb for fluid drainage. It also correctly advises the use of eye drops to prevent infection and cautions against activities that could harm the eye post-surgery. However, the response could be improved by providing more specific details about the follow-up period, such as the frequency and duration of follow-up visits, potential complications to watch for, and a more comprehensive list of activities to avoid. Additionally, the response could mention the importance of adhering to the ophthalmologist's specific instructions, as postoperative care can vary between individuals.

The response is relevant to the question intent, as it addresses what a patient should expect after trabeculectomy surgery. It emphasizes patient safety by advising against activities that could cause trauma to the eye. The response is generally easy to understand from a layman patient's perspective, although it could benefit from simpler explanations of medical terms like "trabecular meshwork" and "bleb."

Rating: 80/100 | 80 |
|---|---|---|---|---|---|---|---|
| 7 | | | | | | | |

| # | Question | Response A | Evaluation A | Score A | Response B | Evaluation B | Score B |
|---|---|---|---|---|---|---|---|
| 8 | What is selective laser trabeculoplasty? | Selective laser trabeculoplasty (SLT) is a type of laser surgery used to treat open-angle glaucoma. It is a safe and effective procedure that helps to lower intraocular pressure (IOP) by improving the drainage of fluid from the eye.<br><br>During SLT, a laser is used to target specific cells in the trabecular meshwork, which is the drainage system of the eye. The laser energy stimulates these cells, causing them to function more efficiently and improve the outflow of fluid from the eye. This helps to reduce the IOP and prevent further damage to the optic nerve.<br><br>SLT is typically performed as an outpatient procedure and does not require any incisions or stitches. It is usually well-tolerated and has a low risk of complications. The effects of SLT can last for several years, but some patients may require repeat treatments to maintain the desired IOP reduction.<br><br>It is important to note that SLT is not suitable for all types of glaucoma, and your ophthalmologist will determine if it is the right treatment option for you based on your specific condition and needs. | The AI's response regarding selective laser trabeculoplasty (SLT) is clinically accurate, relevant to the question intent, ensures patient safety by not providing any dangerously erroneous information, and is easy to understand from a layman patient's perspective. The explanation covers what SLT is, its purpose in treating open-angle glaucoma, the procedure itself, and its safety and efficacy. It also correctly notes that SLT is not suitable for all types of glaucoma, which is an important consideration for patient safety and managing expectations. The response is comprehensive and would likely be very helpful to someone seeking to understand this medical procedure.<br><br>Rating: 95/100 | 95 | Selective laser trabeculoplasty is a procedure that is performed to treat open angle glaucoma. It is a common condition that is usually not serious and often occurs in the elderly. During the procedure, your ophthalmologist will use a laser to treat the drainage channel for the eye, which is the trabecular meshwork. This helps to open up the drainage channel, and allows fluid to flow more easily from the eye. You will be given eye drops to numb your eye before the procedure, and you will need to look straight ahead during the procedure. You may experience some discomfort during the procedure, but it is usually not painful. After the procedure, you will need to use eye drops to prevent infection, and your ophthalmologist will review you to monitor your condition. | Explanation: The AI response provides a general description of selective laser trabeculoplasty (SLT), which is a procedure used to treat open-angle glaucoma by targeting the trabecular meshwork to improve drainage of intraocular fluid and lower intraocular pressure. However, there are some inaccuracies and misleading statements in the response. Firstly, the response describes open-angle glaucoma as "a common condition that is usually not serious and often occurs in the elderly," which is misleading because glaucoma is a serious condition that can lead to vision loss if not properly managed. Additionally, the response does not mention that SLT is typically offered when medications are insufficient or as an adjunct to medical therapy. The response correctly mentions the use of eye drops to numb the eye and the need for follow-up care, but it does not provide information on the potential risks or the fact that the effects of SLT may wear off over time, requiring repeat treatments. The response is generally easy to understand from a layman's perspective, but the clinical inaccuracies and omissions affect the overall quality of the information provided.<br><br>Rating: 65/100 | 65 |